\PassOptionsToPackage{table}{xcolor}
\documentclass{SCIS2021}
\usepackage[T1]{fontenc}
\usepackage{lmodern}
\usepackage[english]{babel}
\usepackage{xcolor}
\usepackage{placeins,array,ragged2e}
\usepackage[algo2e,ruled,vlined,linesnumbered]{algorithm2e}
\SetAlgoNlRelativeSize{-1}
\usepackage{multirow,threeparttable,dcolumn,makecell,bbm,enumitem}
\usepackage{microtype,verbatim,xfrac,etoolbox}
\newcolumntype{C}[1]{>{\centering\arraybackslash}m{#1}}

\graphicspath{{figures/}}
\makeatletter
\patchcmd{\@maketitle}
  {\footnote{*{\thinspace}Corresponding author (email:~\@authoremail)}}
  {\ifx\@authoremail\@empty\else\footnote{*{\thinspace}Corresponding author (email:~\@authoremail)}\fi}
  {}{\PackageError{fortune-scis}{Corresponding-author patch failed}{Check the target class.}}
\makeatother

\begin{document}
\ArticleType{RESEARCH PAPER}
\Year{}\Month{}\Vol{}\No{}\DOI{}\ArtNo{}
\ReceiveDate{}\ReviseDate{}\AcceptDate{}\OnlineDate{}
\title[Flying over The Uncertain Nature (FORTUNE)]{Flying over The Uncertain Nature (FORTUNE): Intelligent and Humanistic 3D Path Planning for Low-Altitude Collaboration}{Flying over The Uncertain Nature (FORTUNE): Intelligent and Humanistic 3D Path Planning for Low-Altitude Collaboration}

\author[1]{Wenbo Zhu}{}
\author[1]{Wenhan Jia}{}
\author[1]{Minghui Liwang}{minghuiliwang@tongji.edu.cn}
\author[1]{Yiguang Hong}{}
\author[2]{Xianbin Wang}{}



\AuthorMark{Zhu W.}
\AuthorCitation{Zhu W., et al.}
\address[1]{Department of Control Science and Engineering, Shanghai Institute of Intelligent Science and Technology,\\Tongji University, Shanghai, China}
\address[2]{Department of Electrical and Computer Engineering, Western University, Ontario, Canada}

\abstract{The proliferation of low-altitude intelligent agents is driving an increasing demand for timely and socially responsible collaborative sensing in distributed and dynamic urban environments. Nevertheless, jointly handling heterogeneous spatio-temporal sensing demands, real-time environmental uncertainty, and human-centered operational constraints within a unified multi-UAV optimization framework remains highly challenging. In this paper, we investigate three-dimensional (3D) multi-uncrewed aerial vehicle (UAV) path planning and joint task assignment for low-altitude collaboration under heterogeneous spatio-temporal uncertainty of ground points of interest (PoIs). Unlike existing studies that assume static and fully known PoIs, we formulate a unified model that simultaneously captures persistent (Type-$\mathsf{I}$), temporally predictable (Type-$\mathsf{II}$), and emergent unpredictable (Type-$\mathsf{III}$) demands. Furthermore, following the principle of \textit{low-altitude for good}, we explicitly model altitude-dependent societal and environmental costs, including noise exposure and public safety risks, thereby jointly optimizing sensing effectiveness and socially compliant UAV operations. To solve the resulting large-scale mixed-integer nonlinear optimization problem, we propose \textit{FORTUNE}, a hierarchical offline-online cooperative framework. Offline, a Transformer-based predictor forecasts the activation windows of Type-$\mathsf{II}$ PoIs, while an enhanced sparrow search algorithm ($\rm ES^2A$), empowered by priority-aware decoding and danger-aware evolution, proactively generates globally coordinated flight plans. Online, a lightweight trajectory refinement module efficiently accommodates newly emerging Type-$\mathsf{III}$ PoIs while preserving global mission coherence. Extensive experiments on real-world traffic datasets and synthetic scenarios demonstrate that FORTUNE consistently outperforms state-of-the-art benchmarks, validating its effectiveness, scalability, and practical applicability to dynamic low-altitude systems.}


\maketitle

\section{Introduction}\label{sec:Intro}
Driven by the flexibility and affordability of low-altitude platforms, rapid advances in sensing/communication/computing have accelerated the deployment of uncrewed aerial vehicles (UAV)-enabled applications in urban monitoring, environmental surveillance, and intelligent transportation~\cite{UAVbackground1,UAVbackground2,backupone}. As a representative paradigm, low-altitude collaborative sensing leverages multiple UAVs to serve timely and large-scale perception for missions such as crowd flow observation and abnormal event detection (e.g., sudden gatherings, traffic accidents, and emergency incidents)~\cite{UAVbackground3,background00,background000,background0000}. Meanwhile, the commercial expansion of the UAV industry, projected to surpass \$48.5 billion by 2029~\cite{UAVbackground4,background0}, further highlights its transformative potential in supporting next-generation intelligent systems. 

Nevertheless, practical multi-UAV collaboration faces fundamental challenges: ground points of interest (PoIs) are inherently distributed and time-varying, with hotspots forming, shifting, and dissipating over short time scales (e.g., crowds relocating across venues). Most existing studies oversimplify this reality by assuming PoIs are known a priori and static, or by merely adopting age-of-information (AoI) as a surrogate for data timeliness~\cite{UAVbackground1,UAVbackground5}, thereby limiting their applicability in real-world deployments. Meanwhile, the increasing density of low-altitude UAV operations also raises societal and operational concerns, including public anxiety over potential crashes and noise disturbance. 

\subsection{Motivations in Form of Q\&A}
The above observations suggest that the challenge of low-altitude collaborative sensing is not merely algorithmic, but structural: it concerns \textit{how dynamic and distributed task patterns should be abstracted}, \textit{how anticipation and real-time adaptation should be reconciled}, and \textit{how operational effectiveness should be balanced with humanistic impact within a unified optimization framework}. Addressing these issues requires moving beyond incremental refinements of existing models toward a more principled formulation. Accordingly, answering the following research questions (RQs) forms our fundamental motivation.

\noindent
$\bullet$~\textit{RQ1: How can multi-UAV systems fundamentally characterize and handle heterogeneous, time-evolving sensing demands, instead of relying on the oversimplified assumption of static and known PoIs?} Existing studies typically assume that sensing targets are fixed and known a priori, which substantially simplifies trajectory planning but departs from real-world low-altitude scenarios where sensing demands exhibit heterogeneous spatio-temporal dynamics. To overcome this limitation, we adopt a structured view of task dynamics that distinguishes different levels of predictability and persistence in ground sensing demands, i.e., PoIs are classified into three types. This abstraction enables proactive planning over stable and partially predictable components while explicitly acknowledging intrinsic uncertainty, thereby bridging the gap between idealized static models and realistic dynamic environments.

\noindent
$\bullet$~\textit{RQ2: How can environmental and societal impact be rigorously embedded into 3D multi-UAV path optimization, so that sensing effectiveness and public acceptability are jointly considered rather than treated as secondary constraints?} Low-altitude UAV operations inherently introduce trade-offs between sensing quality and societal impact, particularly due to altitude-dependent effects. For example, flying at low altitudes may inevitably generate noise, induce public anxiety over potential crashes, and raise privacy concerns regarding terrestrial environments, thereby disturbing both humans and wildlife on the ground. As this paper advocates the concept of \emph{“low-altitude for good (LA4Good)”}, it is essential to explicitly account for socially-compliant and environmental implications of low-altitude UAV deployments. Rather than treating these factors as external constraints, we integrate them directly into our optimization goal. By modeling a certain penalty, the resulting framework achieves a principled balance between task effectiveness and socially responsible operation.

\noindent
$\bullet$~\textit{RQ3: How can proactive planning and reactive adaptation be unified in multi-UAV 3D path design, such that predictable task patterns are efficiently exploited without sacrificing responsiveness to unexpected events?} A fundamental tension in dynamic collaborative sensing lies between long-term efficiency and short-term responsiveness. Purely reactive strategies lead to myopic decisions, whereas fully preplanned strategies fail under unforeseen events. We address this tension by developing a planning paradigm that integrates anticipative path generation with globally consistent online refinement. By preserving awareness of the overall mission state during adjustments, the framework enables rapid adaptation to emergent events while maintaining system-level coordination and efficiency.

\begin{figure}[!b]
	\centering
	\setlength{\abovecaptionskip}{-0.0 mm}
	\includegraphics[width=\linewidth]{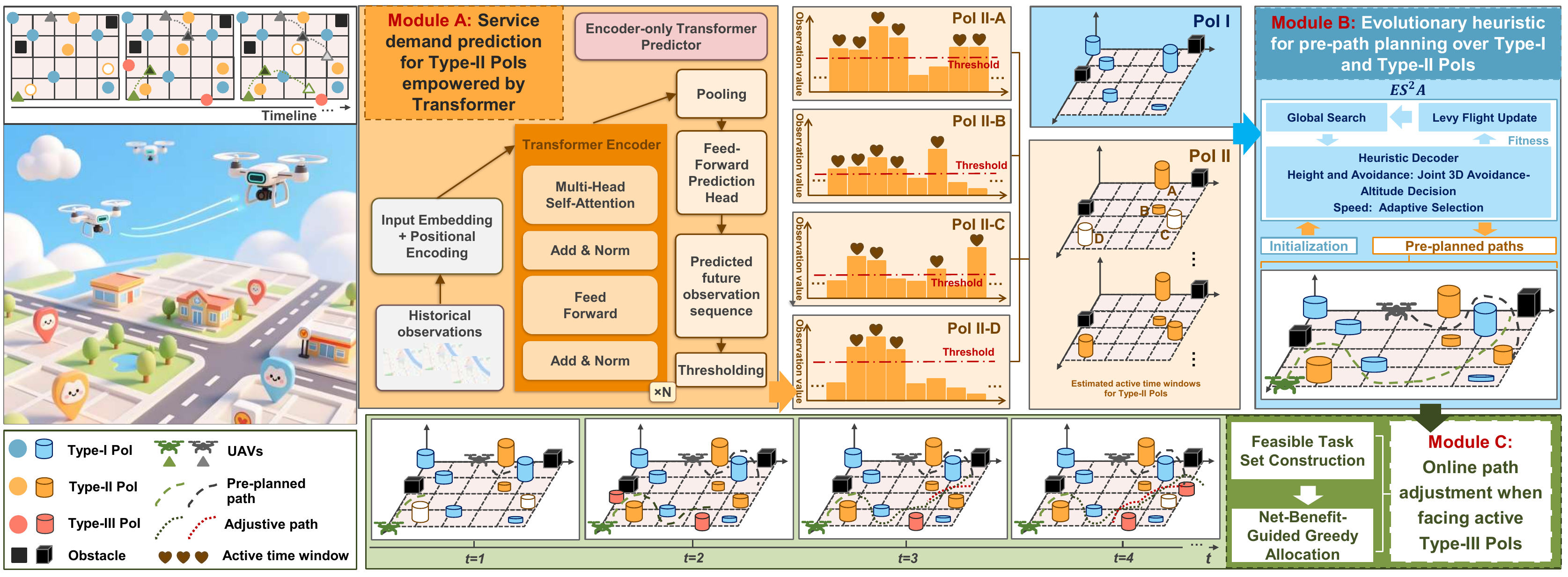}
	\caption{Schematic of FORTUNE with three modules integrated. }
	\label{fig:system}
\end{figure}

\subsection{Literature Review}
We review existing related literature from various perspectives, further highlighting our key differences (also see Table \ref{literaturesummary}, find details in \ref{literature} due to space limitation).

\noindent
$\bullet$~\textit{View 1. Static/known vs. uncertain PoIs.}
Most UAV-assisted data collection studies assume that PoI locations, quantities, and active periods are fixed or known, covering hotspot-based trajectory planning \cite{2026TCCNRLstatic,2026TCOMRLknown}, data freshness optimization \cite{2026TMCfixedregions,2025TMCAoI,2025TONAoIMARL}, and collaborative sensing with predefined targets \cite{2025TVTtwoStagestatic,2025TASEDPoI}. Although heterogeneous AoI requirements were considered in \cite{UAVbackground5}, its setting differs from this paper. Although our previous study \cite{ourtsusc} made an early attempt to integrate demand prediction with evolutionary optimization for multi-UAV crowdsensing in a 2D setting, it considered only a simplified form of task dynamics. In contrast, this work develops a unified 3D framework for PoIs with heterogeneous uncertainty levels, jointly covering persistent, temporally predictable, and unexpectedly emerging demands, where the locations and activation windows of emergent PoIs are unknown in advance.

\noindent
$\bullet$~\textit{View 2. UAV path optimization.}
Existing approaches mainly include RL-based, heuristic, and hybrid methods. RL has been extensively applied to trajectory planning and task allocation \cite{2025TMCAoI,2025TONAoIMARL,2025TVTtwoStagestatic,2026TCCNRLstatic,2026TCOMRLknown,2024TVTDeng,2023TMCRL,2025TITSstaticPoI}, but often suffers from low sample efficiency, unstable training, and limited scalability. Heuristic methods improve computational efficiency \cite{2024IOT,2025ASCheuristic,2024TII}, yet commonly assume static PoIs or deterministic task windows. Hybrid methods decompose complex optimization into multiple stages \cite{2025TCOMthreeStage,2026TCCNheuristicRL,2026TCCNmultistage}, but rarely address uncertain tasks. Our framework therefore integrates PoI activation-window prediction, path pre-planning, and greedy online adjustment.

\noindent
$\bullet$~\textit{View 3. Obstacle avoidance vs. socially compliant navigation.}
Conventional UAV planning mainly considers physical obstacles, energy consumption, and mission performance \cite{2025TITSstaticPoI,2025TVTheuristic,2026TCCNheuristicRL}, while overlooking human discomfort caused by close-range, low-altitude flights and rotor noise. Although proxemics, perceived safety, and social norms have been studied in socially aware robot navigation \cite{2025SANG,2025ACMsurvey}, they remain underexplored for UAVs. We incorporate a human-aware social cost to jointly optimize sensing performance and social acceptability under the LA4Good paradigm.

\begin{table}[t]
\centering
\scriptsize
\setlength{\tabcolsep}{3pt}
\renewcommand{\arraystretch}{1.2}
\caption{Summary of related literature}
\label{literaturesummary}
\begin{tabular}{
C{2.4cm}
C{1.2cm}
C{1.2cm}
C{1.1cm}
C{1.3cm}
C{2.1cm}
C{1.8cm}
C{1.8cm}
}

\toprule

& \multicolumn{7}{c}{\textbf{Attributes}} \\

\cmidrule(lr){2-8}

\multirow{2}{*}{\textbf{Related Study}}
& \multicolumn{2}{c}{\textbf{Path}}
& \multicolumn{2}{c}{\textbf{PoI/UE}}
& \multirow{2}{*}{\makecell{\textbf{Social}\\\textbf{Factor}}}
& \multirow{2}{*}{\textbf{AoI}}
& \multirow{2}{*}{\textbf{Solution}} \\

\cmidrule(lr){2-3}
\cmidrule(lr){4-5}

& 2D & 3D
& \makecell{Static\\Known}
& Dynamic
& & & \\

\midrule

\cite{2026TCCNRLstatic}
& \checkmark &  & \checkmark &  &  &  & RL \\

\cite{2026TMCfixedregions}
& \checkmark &  & \checkmark &  &  & \checkmark & Contract \\

\cite{2026TCOMRLknown}, \cite{2025TITSstaticPoI}
&  & \checkmark & \checkmark &  &  &  & RL \\

\cite{2025TMCAoI}
& \checkmark &  & \checkmark &  &  & \checkmark & MACL \\

\cite{2025TONAoIMARL}
& \checkmark &  & \checkmark &  &  & \checkmark & RL \\

 \cite{UAVbackground5},\cite{2025TASEDPoI}
& \checkmark &  &  & \checkmark &  &  & RL \\

\cite{2026TCCNheuristicRL}
&  & \checkmark &  &  &  & \checkmark & Hybrid \\

\cite{2024TVTDeng}
& \checkmark &  & \checkmark &  &  &  & RL \\

\cite{2024IOT}, \cite{2025ASCheuristic}
& \checkmark &  & \checkmark &  &  &  & Heuristic \\

\cite{2024TII}
&  & \checkmark & \checkmark &  &  &  & Hybrid \\

\cite{2025TCOMthreeStage}
& \checkmark &  & \checkmark &  &  & \checkmark & Hybrid \\

\cite{ourtsusc}
& \checkmark &  & & \checkmark  &  &  Time window & Hybrid \\

\rowcolor[HTML]{CBCEFB}
\textbf{FORTUNE}
&  & \checkmark &  & \checkmark & \checkmark & Time window &
\makecell{Hybrid} \\

\bottomrule

\end{tabular}
\end{table}

\subsection{Spotlight and Contribution}
Recall the above efforts and challenges, we propose \textit{FORTUNE}, abbreviated from \textit{“Flying OveR The Uncertain NaturE”}, an intelligent and humanistic engine for efficient multi-UAV collaboration, over dynamic and heterogeneous PoIs. Main contributions are summarized below.

\noindent
$\bullet$~\textit{We are interested in a novel 3D multi-UAV path planning and task scheduling problem for low-altitude collaborative sensing characterized by static obstacles and dynamic as well as uncertain PoIs.} Specifically, we model PoIs into three heterogeneous types (Type-$\mathsf{I}$, Type-$\mathsf{II}$ and Type-$\mathsf{III}$) to capture diverse spatio-temporal characteristics, thereby moving beyond conventional static and homogeneous assumptions. Furthermore, in line with the LA4Good principle, we explicitly incorporate altitude-induced societal and environmental impacts, such as noise disturbance and perceived safety risk, into the objective function as penalty terms. Together, we formulate a practically meaningful yet mathematically challenging optimization problem that reflects the intrinsic trade-offs between sensing efficiency and socially responsible operation.

\noindent
$\bullet$~\textit{To tackle the optimization with NP hardness, we propose FORTUNE, which establishes a hierarchical offline-online cooperative framework that decomposes it into one regression problem and two path optimization problems, integrating intelligent prediction (Module A), pre-planning (Module B), and online adjustment (Module C).} In the offline stage, a Transformer-based module predicts the active time windows of Type-$\mathsf{II}$ PoIs, enabling anticipatory trajectory pre-planning for Type-$\mathsf{I}$ and -$\mathsf{II}$ PoIs via a dynamic evolutionary optimization scheme under practical constraints such as collision avoidance. In the online stage, when Type-$\mathsf{III}$ PoIs emerge, a time-efficient greedy adjustment module refines UAV trajectories in real time to enhance the overall objective. Collectively, these modules establish a prediction-aware and adaptively refined solution framework that balances proactive efficiency with real-time responsiveness under dynamic and uncertain sensing environments.

\noindent
$\bullet$~\textit{We conduct comprehensive experiments on both real-world and numerical datasets, to verify our performance across diverse parameters. }Experiments demonstrate consistent gains over representative benchmarks.

\section{Core Modeling}\label{sec: modeling and problem}
To describe spatial-temporal dynamism, the whole time horizon is discretized by set $\mathbb{T}=\{1,…,t,…,|\mathbb{T}|\}$, while $\Delta{t}$ denotes the duration of each timeslot.

\subsection{Modeling of PoIs}\label{sec:poi modeling}
In real-world networks, PoIs can either be known, or dynamic and uncertain, on factors such as geographical location, significance, and data volume required for collection. In our modeling, different types of PoIs are considered to capture diverse data needs. For notational simplicity, we use $i$ as the index of PoIs across all types.

\noindent $\bullet$ \textit{Type-$\mathsf{I}$ PoI.} Type-$\mathsf{I}$ PoIs are defined as spatially fixed sensing targets with persistent validity over the entire mission horizon. Their sensing demands are temporally invariant and remain active regardless of external dynamics or contextual variations. Typical instances include permanent infrastructure inspection points, long-term environmental monitoring sites, and predefined land-use survey locations. The deterministic nature of their spatial distribution and sensing requirements enables efficient scheduling and proactive path optimization for UAV operations. In our modeling, let $\mathbb{P}^{(\mathsf{I})}=\{p_1^{(\mathsf{I})},…,p_i^{(\mathsf{I})},…,p_{|\mathbb{P}|}^{(\mathsf{I})}\}$ be the set of type-$\mathsf{I}$ PoIs, where each $p_i^{(\mathsf{I})}\in\mathbb{P}^{(\mathsf{I})}$ is assumed to carry a long-lasting task opening for data collection across the whole time domain (namely, this PoI type always has data collection needs). In particular, each $p_i^{(\mathsf{I})}$ can be represented by a tuple $\left(\bm{l}_i^{(\mathsf{I})}, d_i^{(\mathsf{I})}, r_i^{(\mathsf{I})}\right)$, with $\bm{l}_i^{(\mathsf{I})}=\left(x_i^{(\mathsf{I})},y_i^{(\mathsf{I})},h_i^{(\mathsf{I})}\right)$ denoting the 3D-space coordinate of $p_i^{(\mathsf{I})}$; while $d_i^{(\mathsf{I})}$ and $r_i^{(\mathsf{I})}$ represent the data volume required by the task of $p_i^{(\mathsf{I})}$, as well as the overall reward that the UAVs who have contributed to $p_i^{(\mathsf{I})}$ can get, after completing its task.

\noindent $\bullet$ \textit{Type-$\mathsf{II}$ PoI.} Type-$\mathsf{II}$ PoIs are spatially fixed sensing targets whose demands exhibit predictable temporal dynamics. Their sensing requirements are activated only during specific time windows that recur according to underlying human behavioral patterns or environmental processes. Typical examples include rush-hour traffic monitoring at major intersections and air-quality sensing near heavily traveled roads during commuting periods. The temporal predictability of these demands allows UAVs to perform proactive mission planning through time-aware task assignment and path optimization. Accordingly, we involve a set of type-$\mathsf{II}$ PoIs denoted by $\mathbb{P}^{(\mathsf{II})}=\left\{p_1^{(\mathsf{II})}, \ldots, p_i^{(\mathsf{II})}, \ldots, p_{\lvert \mathbb{P}^{(\mathsf{II})} \rvert}^{(\mathsf{II})}
\right\}$, where each $p_i^{(\mathsf{II})} \in \mathbb{P}^{(\mathsf{II})}$ is represented by a tuple
$\left( \boldsymbol{l}_i^{(\mathsf{II})}, \tau_i^{(\mathsf{II})} \right)$.
Specifically, $\bm{l}_i^{(\mathsf{II})}=\left( x_i^{(\mathsf{II})}, y_i^{(\mathsf{II})}, h_i^{(\mathsf{II})} \right)$ denotes the 3D spatial coordinate of $p_i^{(\mathsf{II})}$. Recall the features of type-$\mathsf{II}$ PoIs, they may appear multiple times throughout the entire time horizon, and each appearance may last for several timeslots, just like stars shining on the map (e.g., a traffic intersection may have data collection requirement during rush hour, while remaining idle for the rest of the day). For ease of understanding, when a PoI has data demands, we say that it is active during the corresponding timeslot. Regarding the above, $\tau_i^{(\mathsf{II})}
=\left\{\tilde{\tau}_i^{(a)} \,\middle|\, 0 \le a \le {A}_i\right\}$ is utilized to record the information of $p_i^{(\mathsf{II})}$ across the whole time horizon, where $\tilde{\tau}_i^{(a)}$ denotes an event corresponding to the $a$-th appearance of PoI $p_i^{(\mathsf{II})}$, and
${A}_i$ indicates the total number of occurrences of $p_i^{(\mathsf{II})}$ on the considered map over the entire time horizon. To further capture the duration of $p_i^{(\mathsf{II})}$ being active, we define each event $\tilde{\tau}_i^{(a)} \in \tau_i^{(\mathsf{II})}$ as $\tilde{\tau}_i^{(a)}=\left\{
\left[ \mathbbm{t}_i^{(+,a)}, \mathbbm{t}_i^{(-,a)} \right], d_i^{(\mathsf{II},a)},
r_i^{(\mathsf{II},a)}\right\}$, where $\left[ \mathbbm{t}_i^{(+,a)}, \mathbbm{t}_i^{(-,a)} \right]$ forms an active time window, with
$\mathbbm{t}_i^{(+,a)} \in \mathbb{T}$ and $\mathbbm{t}_i^{(-,a)} \in \mathbb{T}$ describing the starting and closing timeslots of the $a$-th appearance
$\left( \mathbbm{t}_i^{(+,a)} \le \mathbbm{t}_i^{(-,a)} \right)$, and satisfying
$\mathbbm{t}_i^{(-,a)} < \mathbbm{t}_i^{(+,a+1)}$.
Besides, assuming that a task arises along with each active event, we use
$d_i^{(\mathsf{II},a)}$ and $r_i^{(\mathsf{II},a)}$ to indicate the corresponding data requirement and reward.

To model the emergence of Type-$\mathsf{II}$ PoIs, we define an \emph{observation value} $o_i^{(\mathsf{II})}$ that measures the potential information gain associated with a sensing target. The observation value acts as a trigger signal for PoI activation: a sensing demand is generated once $o_i^{(\mathsf{II})}$ exceeds a predefined threshold. For instance, in intelligent transportation systems, traffic flow intensity can serve as the observation value of an intersection. High traffic volumes may induce sensing demands for traffic monitoring, incident detection, and emission assessment, thereby motivating UAV-assisted data collection. Such activation events can be autonomously identified through real-time observations collected and processed by edge intelligence infrastructures, including roadside units (RSUs) and edge servers. Accordingly, when $o_i^{(\mathsf{II})}$ exceeds a pre-determined threshold $o^{(*)}$, we consider $p_i^{(\mathsf{II})}$ to be activated, whereas it becomes inactive once
$o_i^{(\mathsf{II})}$ falls below $o^{(*)}$.
Apparently, this enables capturing the active time windows of different PoIs.

\noindent $\bullet$ \textit{Type-$\mathsf{III}$ PoI.} Type-$\mathsf{III}$ PoIs are characterized by both spatial and temporal uncertainty. These PoIs emerge unexpectedly in response to sudden events, with no prior knowledge of their location, timing, or sensing requirements. Despite their unpredictability, they often represent critical or high-priority scenarios that demand immediate attention and rapid deployment. Examples involve unexpected traffic accidents on highways, crowd congestion incidents in public spaces, or emergency situations in large outdoor gatherings. Due to their dynamic and urgent nature, these PoIs require highly adaptive, leveraging real-time task updates and on-demand UAV redirection to ensure timely and effective responses. Accordingly, at the beginning of each timeslot $t$, we let the set of type-$\mathsf{III}$ PoIs be $\mathbb{P}^{(\mathsf{III},t)}=\{p_1^{(\mathsf{III},t)},…,p_i^{(\mathsf{III},t)},…\}$, with each PoI associated with a task. More importantly, the number of elements in $\mathbb{P}^{(\mathsf{III},t)}$, namely, $|\mathbb{P}^{(\mathsf{III},t)}|$, can not be known in advance (before $t$), due to the unpredictability and randomness of this type\footnote{A reasonable assumption refers to the appearance timeslot of a Type-$\mathsf{III}$ PoI is uncertain; however, once it emerges, its duration is supposed to be known. Therefore, upon its occurrence, we can capture its active time window.}. Apparently, for each type-$\mathsf{III}$ PoI with starting time ${t}$, we record it as $p_i^{(\mathsf{III},t)} \in \mathbb{P}^{(\mathsf{III},t)}$
and describe it by a tuple$\left( \bm{l}_i^{(\mathsf{III},t)}, \bm{\tau}_i^{(\mathsf{III},t)} \right)$, where $\bm{l}_i^{(\mathsf{III},t)}=\left(x_i^{(\mathsf{III},t)}, y_i^{(\mathsf{III},t)}, h_i^{(\mathsf{III},t)}\right)$ denotes the 3D spatial coordinate of $p_i^{(\mathsf{III},t)}$. Moreover, $\bm{\tau}_i^{(\mathsf{III},t)}=\left\{\left[t, \mathbbm{t}_i^{(-,t)} \right],d_i^{(\mathsf{III},t)},r_i^{(\mathsf{III},t)}\right\}$ collects the information of its active time window $\left[t, \mathbbm{t}_i^{(-,t)} \right]$, the corresponding data requirement
$d_i^{(\mathsf{III},t)}$, and reward $r_i^{(\mathsf{III},t)}$ that can be paid to UAVs upon successful task completion.

\subsection{Modeling of UAVs}\label{sec:uav modeling}
Our model involves UAVs collected by $\mathbb{U}=\left\{u_1, \ldots, u_m, \ldots, u_{\lvert \mathbb{U} \rvert}\right\}$, each equipped with advanced sensors and onboard processors to assist data collection. All UAVs depart from their designated starting locations $\mathbb{I}^{(\mathsf{start})}=\left\{\mathbbm{i}_1^{(\mathsf{start})}, \ldots, \mathbbm{i}_m^{(\mathsf{start})}, \ldots, \mathbbm{i}_{\lvert\mathbb{U} \rvert}^{(\mathsf{start})}\right\}$ and travel to their corresponding destinations $\mathbb{I}^{(\mathsf{end})}=\left\{\mathbbm{i}_1^{(\mathsf{end})}, \ldots, \mathbbm{i}_m^{(\mathsf{end})}, \ldots,\mathbbm{i}_{\lvert \mathbb{U} \rvert}^{(\mathsf{end})}\right\}$.
In particular, each UAV $u_m \in \mathbb{U}$ can be described by a tuple $\left\{
\boldsymbol{l}_m^{(\mathsf{uav},+)}, \boldsymbol{l}_m^{(\mathsf{uav},-)}, d_m^{(\mathsf{uav},{t})}, e_m^{(\mathsf{hov})}, e_m^{(\mathsf{move},v)}, e_m^{(+)}\right\}$, where $\boldsymbol{l}_m^{(\mathsf{uav},+)}=\left(x_m^{(\mathsf{uav},+)},y_m^{(\mathsf{uav},+)},h_m^{(\mathsf{uav},+)}\right)$ and $\boldsymbol{l}_m^{(\mathsf{uav},-)}=\left(x_m^{(\mathsf{uav},-)},y_m^{(\mathsf{uav},-)},h_m^{(\mathsf{uav},-)}\right)$
denote the three-dimensional coordinates of the take-off location and destination of UAV $u_m$, respectively. To ensure the security and privacy of data collection (e.g., preventing sensitive PoIs from close-range reconnaissance by UAVs), each UAV is required to maintain a flight altitude no lower than the minimum safety altitude specified for the corresponding PoIs, denoted by $h_m^{(\mathsf{uav},\min)}$ when performing sensing tasks. Moreover, $d_m^{(\mathsf{uav},t)}$ represents the data sensing capability of UAV $u_m$ during each timeslot (e.g., measured in Mbits$/\Delta t$). Moreover, during $t$, UAV $u_m$ can take one of two actions, namely hovering and flying. Accordingly, $e_m^{(\mathsf{hov})}$ denotes the unit energy consumption for hovering, while $e_m^{(\mathsf{mov},\tilde{v})}$ represents the unit energy consumption when flying at velocity\footnote{To simplify the problem, we neglect the energy consumption caused by acceleration and consider only the energy associated with maintaining different constant flight speeds. This approximation reduces the modeling complexity while still capturing the key impact of speed on energy usage.} $\tilde{v}$. Due to the typically limited endurance of UAVs, we assume that each UAV is initially equipped with an energy budget $e_m^{(+)}$ at take-off\footnote{Since our goal is to investigate dynamic task scheduling and trajectory coordination rather than wireless resource management, the communication process is abstracted from the optimization. Specifically, once a PoI is sensed, the generated data are assumed to be streamed to the nearest AP during subsequent flight whenever connectivity is available. Such an abstraction is consistent with cellular-connected UAV networks, where communication and mobility are naturally pipelined, allowing data uploading to overlap with flight instead of introducing additional mission delay~\cite{communSupport}. As a result, the mission completion time in our model is dominated by the physical sensing process and UAV mobility, while communication latency is assumed to be non-blocking.}.

Also, each UAV should complete its planned flight from the designated starting location to the corresponding destination before depleting its available energy. We further define $q_m^{(\mathsf{uav},t)}=f^{(\mathsf{qua})}\left(\rho_m^{(\mathsf{uav})},h_m^{(\mathsf{uav},t)}
\right)$ to denote the data quality achieved by UAV $u_m$ at timeslot $t$, where $f^{(\mathsf{qua})}(\cdot)$ is a quality evaluation function depending on two key factors.
Specifically, \textit{(i)} $\rho_m^{(\mathsf{uav})}$ represents the inherent sensing capability of UAV $u_m$ (e.g., being equipped with high-definition wide-angle cameras or high-accuracy air pollution sensing modules), which enables higher-quality data acquisition; and \textit{(ii)} $h_m^{(\mathsf{uav},t)}$ denotes the flight altitude of UAV $u_m$ at timeslot $t$, for which an important trade-off arises from the following two aspects.

\noindent$\bullet$ \textit{How $\rho_m^{(\mathsf{uav})}$ impacts $q_m^{(\mathsf{uav},t)}$:} Hardware platforms with superior performance typically enable the collection of higher-quality data, for instance, producing clearer video of the assigned PoI.

\noindent$\bullet$ \textit{How $h_m^{(\mathsf{uav},t)}$ impacts $q_m^{(\mathsf{uav},t)}$:} Subject to the minimum safety altitude constraint, a lower flight altitude generally enables a UAV to acquire higher-quality multi-modal sensing data, including images with finer spatial resolution and audio signals with improved clarity. Nevertheless, operating at lower altitudes may significantly increase the risk of collision with surrounding obstacles\footnote{Obviously, note that UAV is not allowed to arbitrarily minimize its altitude; it should satisfy the PoI-specific minimum altitude constraint, and only above this threshold can it adjust its altitude selection.}. Conversely, detouring to avoid such obstacles often results in longer travel distances and, consequently, substantially higher energy consumption. More critically, as this paper advocates the concept of \emph{``Low-Altitude for Good'' (LA4Good)}, it is essential to account for socially compliant and environmental implications of low-altitude UAV operations.
Specifically, flying at low altitudes may inevitably generate noise, induce panic over potential UAV crashes, and raise privacy concerns regarding terrestrial environments, thereby disturbing humans and wildlife on the ground. Such disturbances can, in turn, diminish the rewards or utilities associated with certain PoIs, which will be formally modeled and discussed in Sec.~\ref{sec: social factors}.

To describe the location of UAV $u_m$ at the beginning of each timeslot, we use 
$\bm{l}_m^{(\mathsf{uav},t)} = \left( x_m^{(\mathsf{uav},t)},\, y_m^{(\mathsf{uav},t)},\, h_m^{(\mathsf{uav},t)} \right)$
to denote the coordinate of $u_m$ at timeslot $t$, and $v_m^{(\mathsf{xy},t)}$, $v_m^{(\mathsf{h},t)}$ show the horizontal and vertical moving velocities of $u_m$, respectively. Apparently, the acceleration of each UAV should not exceed a threshold due to hardware limitations, for example, $\left| v_m^{(\mathsf{xy},t+1)} - v_m^{(\mathsf{xy},t)} \right| \leq v^{(\mathsf{xy},\mathsf{max})},
\quad
\left| v_m^{(\mathsf{h},t+1)} - v_m^{(\mathsf{h},t)} \right| \leq v^{(\mathsf{h},\mathsf{max})},$
where $v^{(\mathsf{xy},\mathsf{max})}$ and $v^{(\mathsf{h},\mathsf{max})}$ are the maximum horizontal and vertical accelerations. Therefore, the flying distance within $\Delta t$ of each UAV can also be constrained such that
$
\left\| \bm{l}_m^{(\mathsf{uav},t+1)} - \bm{l}_m^{(\mathsf{uav},t)} \right\| \leq l^{(\mathsf{max})},
$
where $l^{(\mathsf{max})}$ is the maximum flying distance in a certain period of time. Accordingly, the trajectory of UAV $u_m$ can be defined by
$
\mathbb{L}_m = \left\{ \bm{l}_m^{(\mathsf{uav},1)},\, \bm{l}_m^{(\mathsf{uav},2)},\, \ldots,\, \bm{l}_m^{(\mathsf{uav},t)},\, \ldots,\, \bm{l}_m^{(\mathsf{uav},|T|)} \right\}.
$

\subsection{Analysis on Task Completion and Reward}\label{sec:task comp and reward}
\noindent
\textit{Way to evaluate the completion of a task.} To better describe the assignment between UAVs and different types of PoIs across various timeslots, we use binary indicator $\alpha$ to show whether a UAV is assigned to the task of a corresponding PoI. The answer is yes when $\alpha=1$; while $\alpha=0$, otherwise. In addition, another binary variable $\beta$ is introduced to indicate whether the UAV's flight path covers the location of the corresponding PoI, ensuring the spatial feasibility of the assigned task. Specifically, the UAV task can be executed only when the PoI lies within the UAV’s flight path and the PoI has been assigned to this UAV (i.e., $\alpha\beta=1$). Due to that diverse PoI types may have different active time windows\footnote{To facilitate analysis, to determine whether the UAV has arrived at the location of the PoI (i.e., whether the PoI lies on the UAV's flight trajectory), we assume that the PoI and the UAV share the same horizontal coordinate at time $t$. However, in real-world scenarios, certain deviations inevitably exist. This simplification is adopted for the purpose of mathematical modeling in this paper.}, they have different modeling regarding $\alpha$ and $\beta$, and thus the description of task completion can also be different.

\noindent
$\bullet$ \textit{For type-$\mathsf{I}$ PoIs.} We first define $\alpha_{i,m}^{(\mathsf{I},t)} = 1$ to indicate that the task of type-$\mathsf{I}$ PoI $p_i^{(\mathsf{I})}$ is assigned to UAV $u_m$ at timeslot $t$. Correspondingly, $\beta_{i,m}^{(\mathsf{I},t)}$ can be given by
\begin{equation}
\beta_{i,m}^{(\mathsf{I},t)} =
\begin{cases}
1, \text{if } \alpha_{i,m}^{(\mathsf{I},t)} = 1,\ x_m^{(\mathsf{uav},t)} = x_i^{(\mathsf{I})}, \\~~~~~~~y_m^{(\mathsf{uav},t)} = y_i^{(\mathsf{I})},\ h_m^{(\mathsf{uav},t)} > h_i^{(\mathsf{I})},\\
0, \text{otherwise},
\end{cases}
\label{eq:beta type I}
\end{equation}
which indicates that the trajectory of $u_m$ should also be appropriate for this task to be accomplished. Since the task regarding each PoI should be completed (i.e., the collected data must reach the required data volume of the task) in order to earn the corresponding reward, for type-$\mathsf{I}$ PoI, we use $\hat{d}_i^{(\mathsf{I})}$ to represent the overall received data volume of $p_i^{(\mathsf{I})}$ over the entire time domain, and it can be calculated by:
\begin{equation}
\hat{d}_i^{(\mathsf{I})} = \sum_{t=1}^{|\mathbb{T}|} \sum_{m=1}^{|\mathbb{U}|} \beta_{i,m}^{(\mathsf{I},t)} \, d_m^{(\mathsf{uav},t)}.
\label{eq:data type I}
\end{equation}
Accordingly, we denote $c_i^{(\mathsf{I})}$ as the indicator of whether the task of $p_i^{(\mathsf{I})}$ has been completed, given by:
\begin{equation}
c_i^{(\mathsf{I})} =
\begin{cases}
1, & \hat{d}_i^{(\mathsf{I})} \geq d_i^{(\mathsf{I})}, \\
0, & \text{otherwise}.
\end{cases}
\label{eq:completion type I}
\end{equation}
Apparently, $c_i^{(\mathsf{I})} = 1$ means that the task has been completed.

\noindent
$\bullet$ \textit{For type-$\mathsf{II}$ PoIs.} Then, the modeling of $\alpha_{i,m}^{(\mathsf{II},a,t)}$ and $\beta_{i,m}^{(\mathsf{II},a,t)}$ for type-$\mathsf{II}$ PoIs has a slight difference compared to that of the other types, for which we update their definitions by introducing an integer $a$, with $1 \leq a \leq A_i$. Thus, we have $\alpha_{i,m}^{(\mathsf{II},a,t)} = 1$ indicating that the task associated with the $a$th appearance of type-$\mathsf{II}$ PoI $p_i^{(\mathsf{II})}$ is assigned to UAV $u_m$ at timeslot $t$; while $\alpha_{i,m}^{(\mathsf{II},a,t)} = 0$, otherwise. Accordingly, $\beta_{i,m}^{(\mathsf{II},a,t)}$ can be given by
\begin{equation}
\beta_{i,m}^{(\mathsf{II},a,t)} =
\begin{cases}
1, & \text{if } \alpha_{i,m}^{(\mathsf{II},a,t)} = 1,\ x_m^{(\mathsf{uav},t)} = x_i^{(\mathsf{II})},\ y_m^{(\mathsf{uav},t)} = y_i^{(\mathsf{II})}, \\
  & \quad h_m^{(\mathsf{uav},t)} > h_i^{(\mathsf{II})},\ t \in \left[ \mathbbm{t}_i^{(+,a)},\, \mathbbm{t}_i^{(-,a)} \right], \\
0, & \text{otherwise},
\end{cases}
\label{eq:beta type II}
\end{equation}
meaning that the UAV should arrive at the mapped PoI within its active time window. Similar to previous discussions, for the $a$-th task (the $a$-th appearance) of $p_i^{(\mathsf{II})}$, we have
\begin{equation}
\hat{d}_i^{(\mathsf{II},a)} = \sum_{t=1}^{|\mathbb{T}|} \sum_{m=1}^{|\mathbb{U}|} \beta_{i,m}^{(\mathsf{II},a,t)} \, d_m^{(\mathsf{uav},t)},
\label{eq:data type II}
\end{equation}
and
\begin{equation}
c_i^{(\mathsf{II},a)} =
\begin{cases}
1, & \hat{d}_i^{(\mathsf{II},a)} \geq d_i^{(\mathsf{II},a)}, \\
0, & \text{otherwise}.
\end{cases}
\label{eq:completion type II}
\end{equation}

\noindent
$\bullet$ \textit{For type-$\mathsf{III}$ PoIs.} For $\alpha_{i,m}^{(\mathsf{III},t,t^{\prime})}$ and $\beta_{i,m}^{(\mathsf{III},t,t^{\prime})}$ of  type-$\mathsf{III}$ PoIs, we update their definitions by reinterpreting $t$ as the start time (i.e., activation time) of a type-$\mathsf{III}$ PoI, and introducing $t^{\prime}$ as the timeslot index with $t \leq t^{\prime}$. Thus, $\alpha_{i,m}^{(\mathsf{III},t,t^{\prime})} = 1$ indicates that UAV $u_m$ is scheduled to perform the task of $p_i^{(\mathsf{III},t)}$ during timeslot $t^{\prime}$. Based on this, $\beta_{i,m}^{(\mathsf{III},t,t^{\prime})}$ is calculated by
\begin{equation}
\beta_{i,m}^{(\mathsf{III},t,t^{\prime})} =
\begin{cases}
1, & \begin{aligned}[t]
&\text{if } \alpha_{i,m}^{(\mathsf{III},t,t^{\prime})} = 1, x_m^{(\mathsf{uav},t^{\prime})} = x_i^{(\mathsf{III},t)}, y_m^{(\mathsf{uav},t^{\prime})} = y_i^{(\mathsf{III},t)}, \\
&\quad h_m^{(\mathsf{uav},t^{\prime})} > h_i^{(\mathsf{III},t)},\ t^{\prime} \in \left[ t,\, \mathbbm{t}_i^{(-,t)} \right],
\end{aligned} \\[3pt]
0, & \text{otherwise},
\end{cases}
\label{eq:beta type III}
\end{equation}
In other words, $\beta_{i,m}^{(\mathsf{III},t,t^{\prime})} = 1$ shows that the UAV can arrive at the location of a type-$\mathsf{III}$ PoI that emerged at time $t$ and is still active at $t^{\prime}$, when $\alpha_{i,m}^{(\mathsf{III},t,t^{\prime})} = 1$. Besides, for the task being active at the beginning of timeslot $t$, we have
\begin{equation}
\hat{d}_i^{(\mathsf{III},t)} = \sum_{t^{\prime}=1}^{|\mathbb{T}|} \sum_{m=1}^{|\mathbb{U}|} \beta_{i,m}^{(\mathsf{III},t,t^{\prime})} \, d_m^{(\mathsf{uav},t^{\prime})},
\label{eq:data type III}
\end{equation}
and
\begin{equation}
c_i^{(\mathsf{III},t)} =
\begin{cases}
1, & \hat{d}_i^{(\mathsf{III},t)} \geq d_i^{(\mathsf{III},t)}, \\
0, & \text{otherwise}.
\end{cases}
\label{eq:completion type III}
\end{equation}
\noindent
$\bullet$ \textit{Way to calculate the reward offered by different types of PoIs.} Note that for different PoI types, their rewards, i.e., $r_i^{(\mathsf{I})}$, $r_i^{(\mathsf{II},a)}$, and $r_i^{(\mathsf{III},t)}$, actually depend on two key factors: \textit{(i)} the reward of task completion (RoTC) in time, e.g., reflected by $c_i^{(\mathsf{I})}$, $c_i^{(\mathsf{II},a)}$, $c_i^{(\mathsf{III},t)}$; and \textit{(ii)} the reward of data quality (RoDQ), where higher quality can definitely bring increasing reward. Note that if a task associated with a PoI has not been completed, we have its reward equals to zero. we calculate the reward of different types of PoIs as following: 

\noindent
$\bullet$ \textit{Reward of type-$\mathsf{I}$ PoIs.} Correspondingly, for a type-$\mathsf{I}$ PoI, we have its reward $r_i^{(\mathsf{I})}$ calculated by:
\begin{equation}
r_i^{(\mathsf{I})} =
\underbrace{c_i^{(\mathsf{I})} \, \tilde{r}_i^{(\mathsf{I})}}_{\text{RoTC of Type I PoI}}
+
\underbrace{c_i^{(\mathsf{I})} \, \Delta \tilde{r} \sum_{t=1}^{|\mathbb{T}|} \sum_{m=1}^{|\mathbb{U}|} \beta_{i,m}^{(\mathsf{I},t)} \, q_m^{(\mathsf{uav},t)}}_{\text{RoDQ of Type I PoI}},
\label{eq:reward type I}
\end{equation}
where $\Delta \tilde{r}$ indicates the marginal gain in reward with respect to data quality.

\noindent
$\bullet$ \textit{Reward of type-$\mathsf{II}$ PoIs.} 
Similar to type-$\mathsf{I}$ PoIs, we write the reward of task of type-$\mathsf{II}$ PoI for its $a$-th appearance as:
\begin{equation}
r_i^{(\mathsf{II},a)} =
\underbrace{c_i^{(\mathsf{II},a)} \, \tilde{r}_i^{(\mathsf{II})}}_{\text{RoTC of Type II PoI}}
+
\underbrace{c_i^{(\mathsf{II},a)} \, \Delta \tilde{r} \sum_{t=1}^{|\mathbb{T}|} \sum_{m=1}^{|\mathbb{U}|} \beta_{i,m}^{(\mathsf{II},a,t)} \, q_m^{(\mathsf{uav},t)}}_{\text{RoDQ of Type II PoI}},
\label{eq:reward type II}
\end{equation}

\noindent
$\bullet$ \textit{Reward of type-$\mathsf{III}$ PoIs.} Similarly, the reward of a type-$\mathsf{III}$ PoI emerging at timeslot $t$ is defined by
\begin{equation}
r_i^{(\mathsf{III},t)} =
\underbrace{c_i^{(\mathsf{III},t)} \, \tilde{r}_i^{(\mathsf{III})}}_{\text{RoTC of Type III PoI}}
+
\underbrace{c_i^{(\mathsf{III},t)} \, \Delta \tilde{r} \sum_{t^{\prime}=1}^{|\mathbb{T}|} \sum_{m=1}^{|\mathbb{U}|} \beta_{i,m}^{(\mathsf{III},t,t^{\prime})} \, q_m^{(\mathsf{uav},t^{\prime})}}_{\text{RoDQ of Type III PoI}},
\label{eq:reward_III}
\end{equation}

\subsection{Modeling of Altitude-Dependent Social/Environmental Proxy Brought by UAVs}
\label{sec: social factors}
We provide a novel perspective on LA4Good by explicitly accounting for the potential social and environmental externalities associated with the operation of low-altitude intelligent agents. Existing studies on urban UAV operations and 3D aerial path planning have emphasized that UAV trajectories should be designed not only with respect to conventional objectives, such as energy consumption and travel efficiency, but also by considering community noise exposure, public safety, and other socially relevant factors \cite{RamosRomero2022,Hohmann2023}. Motivated by these considerations, we characterize the altitude-dependent negative effects associated with the UAV flight altitude $h_m^{(\mathsf{uav},t)}$. In particular, UAVs operating at lower altitudes are generally closer to pedestrians, vehicles, buildings, and wildlife habitats, and may therefore result in greater community noise exposure, stronger perceived safety concerns, and more pronounced disturbance to local wildlife, especially birds. These proximity-induced effects may reduce public acceptance of UAV operations and lead to adverse consequences for both the ground environment and the low-altitude ecosystem. Accordingly, we define the ecological penalty function as $f^{(\mathsf{eco},t)}\left(h_m^{(\mathsf{uav},t)}\right)$, which serves as an altitude-dependent proxy for the aggregate social and environmental penalty incurred by UAV $m$ during timeslot $t$. 
We assume that $f^{(\mathsf{eco},t)}\left(h_m^{(\mathsf{uav},t)}\right)$ is a monotonically decreasing function of $h_m^{(\mathsf{uav},t)}$: the lower the UAV flies, the greater the potential impact imposed on the ground environment, including pedestrians and vehicles, as well as on the low-altitude ecosystem, including birds. This modeling assumption is also consistent with noise-aware UAV operation studies, which demonstrate that operational parameters such as flight altitude and separation distance can be adjusted to mitigate community noise impacts \cite{RamosRomero2022}. The specific form of the penalty function may be selected according to the considered deployment environment and application requirements. For example, it can be modeled by a sigmoid function:
\begin{equation}
f^{(\mathsf{eco},t)}= 
\sum_{m=1}^{|\mathbb{U}|} \frac{\varepsilon}{1 + e^{\lambda^{(\times)} \left( h_m^{(\mathsf{uav},t)} - h^{(0)} \right)}},
\label{eq:penalty}
\end{equation}
where $\varepsilon$ is the coefficient of penalty, while $h^{(0)}$ is interpreted as a critical altitude: below it, the UAV has a strong negative effect; above it, the impact becomes progressively smaller.

\subsection{Modeling of UAV Collision Risks (Including Obstacles and Mid-Air Collisions) }
\label{sec: collisions}
While excessively low UAV flight altitudes may cause negative impacts, this paper promotes the broader vision of LA4Good by also addressing two critical dimensions of risk management: \textit{(i) strict control over UAV-obstacle collisions and (ii) soft control over mid-air collisions.} In the following, we offer details regarding these two aspects.

\noindent
$\bullet$ \textit{Strict control over UAV-obstacle collisions.} Given the inevitable presence of obstacles such as tall buildings, especially in urban environments, this paper enforces a strict constraint to prevent UAVs from colliding with them during flight or task execution. Let $\mathbb{B} = \{ b_1, \ldots, b_n, \ldots, b_{|\mathbb{B}|} \}$ be the set of obstacles, where $b_n$ is represented by a tuple $\left( \bm{\pi}_n^{(\mathsf{obs})}, c_n^{(\mathsf{obs})} \right)$. In particular, $\bm{\pi}_n^{(\mathsf{obs})} = \left(x_n^{(\mathsf{obs})},\, y_n^{(\mathsf{obs})},\, h_n^{(\mathsf{obs})}, \mathbbm{r}^{(\mathsf{obs},\mathsf{l})}_n, \mathbbm{r}^{(\mathsf{obs},\mathsf{w})}_n\right)$
denotes a rectangular cuboid space whose central axis is the vertical line through $(x_n^{(\mathsf{obs})},y_n^{(\mathsf{obs})})$ centered at height $h_n^{(\mathsf{obs})}$, and whose horizontal length and width are $2\mathbbm{r}^{(\mathsf{obs},\mathsf{l})}_n$ and $2\mathbbm{r}^{(\mathsf{obs},\mathsf{w})}_n$, respectively. Moreover, $c_n^{(\mathsf{obs})}$ represents the cost of detour (e.g., the length of additional route needs to be taken) when a UAV passes by $b_n$. When an obstacle is encountered along the UAV’s trajectory, we enforce strict control by providing two possible options: \textit{(i)} the UAV adjusts its altitude to overfly the obstacle (and correspondingly consumes energy), or \textit{(ii)} it performs a lateral detour to circumvent the obstacle, which may result in additional travel costs.

\noindent
$\bullet$ \textit{Soft control over mid-air collisions.} The potential for UAV-to-UAV collisions increases as more UAVs operate simultaneously within the same region (e.g., location of a PoI). However, such collisions are not always inevitable. In our modeling, we adopt a soft manner to manage this: instead of directly limiting the number of UAVs, we constrain the probabilistic risk of collision, which increases with UAV density. This allows for flexibility in UAV deployment while maintaining safety-aware operation.

For type-$\mathsf{I}$ PoI $p_i^{(\mathsf{I})} \in \mathbb{P}^{(\mathsf{I})}$, type-$\mathsf{II}$ PoI $p_i^{(\mathsf{II})} \in \mathbb{P}^{(\mathsf{II})}$, and type-$\mathsf{III}$ PoI $p_i^{(\mathsf{III},t)} \in \mathbb{P}^{(\mathsf{III},t)}$, the number of UAVs that may gather at the corresponding location during timeslot $t$ (or $t'$ for type-$\mathsf{III}$ PoI) can be calculated by
$\sum_{m=1}^{|\mathbb{U}|} \beta_{i,m}^{(\mathsf{I},t)}$,
$\sum_{m=1}^{|\mathbb{U}|} \beta_{i,m}^{(\mathsf{II},a,t)}$ (with $1 \leq a \leq A_i$),
and
$\sum_{m=1}^{|\mathbb{U}|} \beta_{i,m}^{(\mathsf{III},t,t^{\prime})}$,
respectively. To ensure a safe low-altitude operating environment, a reference metric is introduced to control UAV density, representing the recommended maximum number of UAVs that can simultaneously occupy a given location, denoted as $s^{(\mathsf{max})}$. Based on the above, the risk of UAV collision at a certain timeslot regarding different PoIs can be modeled using a shifted-and-normalized sigmoid function to keep the risk value within the interval $[0,1]$, given as formulas (\ref{eq:risk_I})-(\ref{eq:risk_III}).

\begin{equation}
\theta_i^{(\mathsf{I},t)} =
\begin{cases}
0, & \text{if } \sum_{m=1}^{|\mathbb{U}|} \beta_{i,m}^{(\mathsf{I},t)} \leq s^{(\mathsf{max})}, \\
\frac{2}{1 + e^{-\lambda^{(\mathsf{num})} \left( \sum_{m=1}^{|\mathbb{U}|} \beta_{i,m}^{(\mathsf{I},t)} - s^{(\mathsf{max})} \right)}} - 1, & \text{if } \sum_{m=1}^{|\mathbb{U}|} \beta_{i,m}^{(\mathsf{I},t)} > s^{(\mathsf{max})},
\end{cases}
\label{eq:risk_I}
\end{equation}


\begin{equation}
\theta_i^{(\mathsf{II},a,t)} =
\begin{cases}
0, & \text{if } \sum_{m=1}^{|\mathbb{U}|} \beta_{i,m}^{(\mathsf{II},a,t)} \leq s^{(\mathsf{max})}, \\
\frac{2}{1 + e^{-\lambda^{(\mathsf{num})} \left( \sum_{m=1}^{|\mathbb{U}|} \beta_{i,m}^{(\mathsf{II},a,t)} - s^{(\mathsf{max})} \right)}} - 1, & \text{if } \sum_{m=1}^{|\mathbb{U}|} \beta_{i,m}^{(\mathsf{II},a,t)} > s^{(\mathsf{max})},
\end{cases}
\label{eq:risk_II}
\end{equation}

\begin{equation}
\theta_i^{(\mathsf{III},t,t^{\prime})} =
\begin{cases}
0, & \text{if } \sum_{m=1}^{|\mathbb{U}|} \beta_{i,m}^{(\mathsf{III},t,t^{\prime})} \leq s^{(\mathsf{max})}, \\
\frac{2}{1 + e^{-\lambda^{(\mathsf{num})} \left( \sum_{m=1}^{|\mathbb{U}|} \beta_{i,m}^{(\mathsf{III},t,t^{\prime})} - s^{(\mathsf{max})} \right)}} - 1, & \text{if } \sum_{m=1}^{|\mathbb{U}|} \beta_{i,m}^{(\mathsf{III},t,t^{\prime})} > s^{(\mathsf{max})}.
\end{cases}
\label{eq:risk_III}
\end{equation}

\section{3D Multi-UAV Path Planning and Task Scheduling over Diverse PoIs with Socially-Compliant Concerns}

\subsection{Problem Description}\label{problem formulation}
Our primary goal is to optimize 3D paths of multiple UAVs coupled with task scheduling over heterogeneous PoIs, in environments with dynamic and uncertain nature. To achieve this, the designed optimization process aims to maximize the net reward derived from UAV services, which strives to achieve a well-balanced trade-off between service quality and the mitigation of potential negative impacts that low-altitude intelligent agents may have on human populations and surrounding ecosystems, while simultaneously incorporating multi-dimensional risk control. Recall our previous discussions, we first write the positive reward that UAVs can obtain for serving diverse PoI types as:
\begin{equation}
\mathbb{R}^{(\mathsf{+})} = 
\sum_{i=1}^{|\mathbb{P}^{(\mathsf{I})}|} r_i^{(\mathsf{I})}
+ \sum_{i=1}^{|\mathbb{P}^{(\mathsf{II})}|} \sum_{a=1}^{A_i} r_i^{(\mathsf{II},a)}
+ \sum_{t=1}^{|\mathbb{T}|} \sum_{i=1}^{|\mathbb{P}^{(\mathsf{III},t)}|} r_i^{(\mathsf{III},t)},
\label{eq:reward_positive}
\end{equation}

Then, the overall negative impacts across the whole time domain can be calculated by:
\begin{equation}
\mathbb{R}^{(\mathsf{-})} = \sum_{t=1}^{|\mathbb{T}|} f^{(\mathsf{eco},t)}\!\left( h_m^{(\mathsf{uav},t)} \right),
\label{eq:reward_negative}
\end{equation}

Accordingly, combining these two, we have the net reward computed by:
\begin{equation}
\mathbb{R}^{(\mathsf{net})} = \mathbb{R}^{(\mathsf{+})} - \mathbb{R}^{(\mathsf{-})}.
\label{eq:reward_net}
\end{equation}
Accordingly, the optimization problem of our interest is defined as the following $\bm{\mathcal{P}}$: 
\begin{equation}
\bm{\mathcal{P}}:~\underset{\bm{\alpha},\, \{\mathbb{L}_m\}_{u_m \in \mathbb{U}}}{\text{argmax}} \quad \mathbb{R}^{(\mathsf{net})}
\end{equation}
\begin{center}
\textit{s.t.}
\end{center}
\vspace{-1.2em}
\begin{flalign}
& \text{(C1):}~\sum_{t=1}^{|\mathbb{T}|} e_m^{(t)} \leq e_m^{(\mathsf{+})},  \forall\, u_m \in \mathbb{U} \notag& \\
& \text{(C2):}~\theta_i^{(\mathsf{I},t)}\leq\theta^{\mathsf{max}},~\theta_i^{(\mathsf{II},a,t)}\leq\theta^{\mathsf{max}},\theta_i^{(\mathsf{III},t,t')}\leq \theta^{\mathsf{max}}  \notag& \\
& \text{(C3):}~\bm{l}^{(\mathsf{uav},+)}=\mathbb{I}^{\mathsf{start}}, \bm{l}^{(\mathsf{uav},-)}=\mathbb{I}^{\mathsf{end}}\notag&\\
& \text{(C4):}~\bm{l}^{(\mathsf{uav},t)} \notin \bm{\pi}^{(\mathsf{obs})}_n, \forall b_n \in \mathbb{B}   \notag& \\
& \text{(C5):}~\left|v_m^{(\mathsf{xy},t+1)}-v_m^{(\mathsf{xy},t)} \right| \leq v^{(\mathsf{xy},\mathsf{max})}, \left|v_m^{(\mathsf{h},t+1)}-v_m^{(\mathsf{h},t)} \right| \leq v^{(\mathsf{h},\mathsf{max})},  \notag&
\end{flalign}
where $\bm{\alpha}$ denotes the profile of all $\alpha_{i,m}^{(\mathsf{I},t)}$, $\alpha_{i,m}^{(\mathsf{II},a,t)}$ and  $\alpha_{i,m}^{(\mathsf{III},t,t^\prime)}$ (note that each UAV can only perform one task during one timeslot), and $\{\mathbb{L}_m\}_{u_m \in \mathbb{U}}$ collects all the paths cross UAVs. Particularly, in $\mathcal{P}$, constraint (C1) limits the energy available to each UAV, while $\sum_{t=1}^{|\mathbb{T}|} e_m^{(t)}$ depens on factors such as $e^{(\mathsf{hov})}_m$, $e^{(\mathsf{mov, \tilde{v}})}$. Constraint (C2) constrains the collision risk among UAVs, while (C3) ensures that each UAV departs from its origin and lands at the designated destination to accomplish the assigned task, and (C4) enforces obstacle avoidance. Moreover, (C5) bounds the horizontal and vertical accelerations within the maximum allowable limits. In particular, problem $\mathcal{P}$ involves tightly coupled decision variables: the binary assignment variable determines the task-UAV matching, while the path variable characterizes the continuous spatial-temporal flight path of each UAV, including geographical coordinates and altitude profiles. Notably, the path is not uniquely determined by the assignment decision, as it should be jointly optimized to enhance task completion quality, mitigate ground impact, and accommodate flight behaviors such as hovering. Consequently, the problem entails the coupling of discrete assignment variables and high-dimensional continuous trajectory variables, leading to a large-scale mixed-integer optimization problem. 

\subsection{Problem Transformation and Solution Design}
\label{transform and solution}
To tackle the resulting complexity, we decompose the problem into two sequential stages, embedded with three sub-problems detailed below: \textit{(i) Offline planning stage (before $t=1$):} During this time, we first estimate the future service demands of type-$\mathsf{II}$ PoIs by solving a regression problem, with a Transformer-based forecasting module (Module A) developed. Based on these predictions, we further formulate an offline optimization problem, employing a carefully designed evolutionary heuristic to pre-plan efficient UAV paths over type-$\mathsf{I}$ and type-$\mathsf{II}$ PoIs, with the aim to maximize the expected net reward (Module B, only associated with these two PoI types). Such a design is motivated by the fact that the information of type-$\mathsf{III}$ PoIs is unavailable until they actually emerge in the operational area. \textit{(ii) Online adaptation stage (from $t=1$):} In this stage, UAVs initially follow the pre-determined paths. Once any type-$\mathsf{III}$ PoI becomes active, we dynamically adjust the current paths by solving an online optimization problem, for which we investigate a greedy-empowered yet time-efficient refinement strategy (Module C), so as to promptly adapt to the newly revealed information and further improve the achieved net reward. This hybrid two-stage framework effectively balances planning optimality and decision-making efficiency, enabling fast responses to unexpected events while preserving high-quality solutions under practical computational constraints. In the following, we delve into details associated with the three modules, forming our FORTUNE.


\subsubsection{Module A. Transformer-based service demand prediction for Type-$\mathsf{II}$ PoIs (Stage 1)}

Type-$\mathsf{II}$ PoIs are spatially fixed sensing targets whose service demands exhibit temporally varying yet predictable activation patterns. Their spatial locations are known in advance, whereas their sensing tasks are activated only during specific time intervals determined by the evolution of the corresponding observation signals. Module A is therefore designed to forecast the future observation sequence of each Type-$\mathsf{II}$ PoI and subsequently convert the predicted sequence into active time windows. The resulting time-window estimates are then provided to Module B for offline task scheduling and UAV path pre-planning. Specifically, the Transformer predictor is trained offline using historical observations. Before UAV take-off, the trained predictor takes the most recent observation sequence of each Type-$\mathsf{II}$ PoI as input and estimates its future demand evolution. For a Type-$\mathsf{II}$ PoI $p_i^{(\mathsf{II})}$, let $t$ denote the first timeslot of the prediction horizon, and let $L^{(\mathsf{his})}$ denote the length of the historical observation window. The corresponding input sequence is expressed as
\begin{equation}
\mathbf{X}_{i}^{(\mathsf{in},t)}
=
\left[
\mathbf{x}_{i}^{(t-L^{(\mathsf{his})})},
\mathbf{x}_{i}^{(t-L^{(\mathsf{his})}+1)},
\ldots,
\mathbf{x}_{i}^{(t-1)}
\right]^{\top}
\in \mathbb{R}^{L^{(\mathsf{his})}\times d^{(\mathsf{model})}},
\end{equation}
where $\mathbf{x}_{i}^{(\tau)}\in\mathbb{R}^{d^{(\mathsf{model})}}$ denotes the embedded representation of the normalized observation feature associated with $p_i^{(\mathsf{II})}$ at timeslot $\tau\in\{t-L^{(\mathsf{his})},\ldots,t-1\}$. For example, in an intelligent transportation scenario, the raw observation feature may correspond to traffic-flow intensity, which is first normalized and projected into a $d^{(\mathsf{model})}$-dimensional embedding space. Here, $d^{(\mathsf{model})}$ denotes the feature embedding dimension of the Transformer.

Since the Transformer does not employ recurrent structures, positional information must be explicitly incorporated to preserve the temporal ordering of the input sequence. Specifically, for the $o$-th temporal position, the sinusoidal positional encoding is defined as
\begin{equation}
\mathbf{E}_{o,2j}
=
\sin\left(
\frac{o}{10000^{2j/d^{(\mathsf{model})}}}
\right),
\mathbf{E}_{o,2j+1}
=
\cos\left(
\frac{o}{10000^{2j/d^{(\mathsf{model})}}}
\right),
\end{equation}
where $o\in\{0,\ldots,L^{(\mathsf{his})}-1\}$ denotes the temporal-position index, and $j$ denotes the embedding-dimension index. The position-aware input representation is then given by
\begin{equation}
\mathbf{Z}_{i}^{(0,t)}
=
\mathbf{X}_{i}^{(\mathsf{in},t)}
+
\mathbf{E},
\end{equation}
where $\mathbf{E}\in\mathbb{R}^{L^{(\mathsf{his})}\times d^{(\mathsf{model})}}$ denotes the positional-encoding matrix. Based on the embedded sequence, the Transformer encoder extracts temporal correlations through multi-head self-attention. For each attention head, the input representation is linearly projected into the query, key, and value matrices, denoted by $\mathbf{Q}$, $\mathbf{K}$, and $\mathbf{V}$, respectively. The scaled dot-product attention is calculated as
\begin{equation}
{\rm A}(\mathbf{Q},\mathbf{K},\mathbf{V})
=
{\rm softmax}
\left(
\frac{\mathbf{Q}\mathbf{K}^{\top}}{\sqrt{d_k}}
\right)
\mathbf{V},
\end{equation}
where $d_k$ denotes the dimension of each key vector. By applying multiple attention heads in parallel, the encoder learns temporal dependencies from different representation subspaces, thereby capturing both short-term fluctuations and long-range correlations in the service-demand evolution. The outputs of the attention heads are concatenated and subsequently processed by position-wise feed-forward networks, residual connections, and layer-normalization operations.

After passing through the stacked Transformer encoder layers, a linear prediction head maps the learned temporal representations to the future observation sequence:
\begin{equation}
\hat{\mathbf{o}}_{i}^{(\mathsf{II},t:t+L^{(\mathsf{pred})}-1)}
=
F_{\Theta}
\left(
\mathbf{X}_{i}^{(\mathsf{in},t)}
\right),
\end{equation}
where $F_{\Theta}(\cdot)$ denotes the Transformer-based predictor parameterized by $\Theta$, $L^{(\mathsf{pred})}$ denotes the prediction horizon, and
$\hat{\mathbf{o}}_{i}^{(\mathsf{II},t:t+L^{(\mathsf{pred})}-1)}
=
[\hat{o}_{i}^{(\mathsf{II},t)},\ldots,
\hat{o}_{i}^{(\mathsf{II},t+L^{(\mathsf{pred})}-1)}]^{\top}$
collects the predicted observation values over the future horizon. The predictor is trained by minimizing the mean squared error between the predicted and ground-truth observation sequences over the training samples.

The predicted observation values are subsequently converted into binary activation states according to the predefined activation threshold $o^{(*)}$. Specifically, Type-$\mathsf{II}$ PoI $p_i^{(\mathsf{II})}$ is predicted to be active at a future timeslot $\tau\in\{t,\ldots,t+L^{(\mathsf{pred})}-1\}$ if
$\hat{o}_{i}^{(\mathsf{II},\tau)}\geq o^{(*)}$; otherwise, it is regarded as inactive. Each maximal sequence of consecutive active timeslots is then identified as a predicted active time window of $p_i^{(\mathsf{II})}$. Finally, the predicted tasks associated with these active windows, together with the Type-$\mathsf{I}$ tasks that remain active throughout the mission horizon, constitute the offline task information supplied to Module B for task assignment and three-dimensional UAV path pre-planning.


\subsubsection{Module B. Offline 3D path pre-planning via $\rm ES^2A$} 

Thanks to Module A predictor, Module B aims to perform offline path pre-planning for Type-$\mathsf{I}$ and Type-$\mathsf{II}$ PoIs by designing three sub-modules (abbreviated as SubM for simplicity). However, directly optimizing task assignments and 3D paths for multiple UAVs is still computationally expensive, since both discrete scheduling decisions and continuous position variables are involved. To this end, we design an \textbf{e}nhanced \textbf{s}parrow \textbf{s}earch \textbf{a}lgorithm ($\rm ES^2A$), where each individual searches over the task priority space rather than directly encoding UAV waypoints.


\noindent
$\bullet$ \textbf{SubM B.1: Priority-driven indirect encoding and Tent-chaotic-mapping-based population initialization.}
A fundamental challenge in multi-UAV path planning is to coordinate task assignment and path  generation across multiple UAVs without introducing an excessively high-dimensional search space. Under a direct coordinate-based representation, the three-dimensional position of each UAV must be explicitly encoded over the considered time horizon. For the timeslot-discretized trajectory model adopted in this paper, such a representation involves a search dimension on the order of $D^{(\mathsf{coord})}
=3|\mathcal{U}||\mathcal{T}|$, even before additional task-assignment, speed-selection, and altitude-decision variables are incorporated. Consequently, the search dimension grows rapidly with both the fleet size and the mission horizon, substantially increasing the difficulty of evolutionary optimization.

To alleviate this issue, we develop a task-priority-driven indirect encoding strategy. Instead of directly encoding the 3D coordinates of individual UAV paths, each candidate solution is represented by a real-valued task-priority vector. The priority vector determines the order in which offline tasks are considered by the hierarchical decoder, whereas the corresponding UAV-task assignments, flight paths, altitudes, and velocities are subsequently generated by the decoder introduced in SubM B.2. Therefore, swarm-level coordination is achieved implicitly through the joint decoding of a global task-priority sequence rather than through the direct optimization of all UAV coordinates. We let $\mathcal{T}^{(\mathsf{off})}=
\left\{
\phi_1,\phi_2,\ldots,\phi_s,\ldots,
\phi_{N^{(\mathsf{task})}}
\right\}$ 
denote the offline task set containing $N^{(\mathsf{task})}$ tasks (tasks associated with all Type-$\mathsf{I}$ PoIs and the predicted active-window instances of Type-$\mathsf{II}$ PoIs obtained from Module A). Each predicted activation window of a Type-$\mathsf{II}$ PoI is treated as an individual offline task because different activation windows may have distinct data requirements, rewards, and temporal constraints. The resulting priority representation has dimension $D^{(\mathsf{pri})}=N^{(\mathsf{task})}$, 
which is independent of the number of discretized trajectory coordinates. When
$N^{(\mathsf{task})}\ll 3|\mathcal{U}||\mathcal{T}|$, this indirect representation substantially reduces the dimension handled by the evolutionary search. Let $g$ denote the generation index. The key concepts used in $\rm ES^2A$ are defined as follows.

\noindent
\textit{Definition 1 (Population):}
At generation $g$, the population of $\rm ES^2A$ is defined as
\begin{equation}
\mathcal{X}^{(g)}
=
\left\{
\mathbf{x}_{1}^{(g)},\mathbf{x}_{2}^{(g)},\ldots,
\mathbf{x}_{o}^{(g)},\ldots,
\mathbf{x}_{N^{(\mathsf{p})}}^{(g)}
\right\},
\end{equation}
where $N^{(\mathsf{p})}$ denotes the population size and
$\mathbf{x}_{o}^{(g)}$ denotes the $o$-th encoded candidate solution at generation $g$.

\noindent
\textit{Definition 2 (Individual):}
An individual $\mathbf{x}_{o}^{(g)}$ is a latent swarm-level candidate solution represented in the task-priority space. It does not directly specify the coordinates of UAV trajectories. Instead, after being processed by the hierarchical decoder in SubM B.2, $\mathbf{x}_{o}^{(g)}$ induces a complete candidate solution comprising task assignments, task-service sequences, three-dimensional UAV paths, flight and collection altitudes, and flight-speed decisions. The fitness of the individual is subsequently evaluated according to the net reward achieved by its decoded solution. Therefore, the evolutionary search is performed over encoded task priorities, whereas feasibility with respect to time windows, energy budgets, obstacle avoidance, altitude limits, and collision risks is handled during decoding.

\noindent
\textit{Definition 3 (Priority vector and priority weight):}
For individual $\mathbf{x}_{o}^{(g)}$, its priority vector is defined as
\begin{equation}
\mathbf{x}_{o}^{(g)}
=
\left[
x_{o,1}^{(g)},x_{o,2}^{(g)},\ldots,
x_{o,s}^{(g)},\ldots,
x_{o,N^{(\mathsf{task})}}^{(g)}
\right]^{\top}
\in
[0,1]^{N^{(\mathsf{task})}},
\end{equation}
where $x_{o,s}^{(g)}$ denotes the priority weight assigned to task $\phi_s$. A larger value of $x_{o,s}^{(g)}$ indicates that task $\phi_s$ should be considered earlier by the decoder. Specifically, the task indices are sorted in descending order of their priority weights to obtain the global priority sequence
\begin{equation}
\label{eq:priority_order}
\boldsymbol{\pi}_{o}^{(g)}
=
\underset{s\in\{1,\ldots,N^{(\mathsf{task})}\}}{\operatorname{argsort}}
{\downarrow}
\left(
x_{o,s}^{(g)}
\right),
\end{equation}
where $\boldsymbol{\pi}_{o}^{(g)}
=
[\pi_{o,1}^{(g)},\ldots,\pi_{o,N^{(\mathsf{task})}}^{(g)}]$
specifies the order in which the tasks are examined during decoding. If equal priority weights occur due to finite numerical precision, they are resolved according to a fixed deterministic tie-breaking rule. It should be emphasized that $\boldsymbol{\pi}_{o}^{(g)}$ represents a task-consideration order rather than a directly executable task schedule. The actual assignment and execution order of each UAV are determined by the decoder according to the current fleet state and the relevant feasibility constraints. In this manner, the original combinatorial scheduling problem is searched through a continuous real-valued priority representation, while the resulting task allocation and UAV paths remain discrete and constraint-aware after decoding.

In a standard $\rm S^2A$, the initial population is commonly generated using independent pseudo-random samples. However, when $N^{(\mathsf{task})}$ is large, a finite number of random samples may provide insufficient coverage of the priority space and may produce highly similar initial task rankings. To improve the dispersion of the initial population, we employ the Tent chaotic map to generate the priority weights in $\mathcal{X}^{(0)}$. Starting from an initial value $z_0\in(0,1)$, the Tent map is defined as
\begin{equation}
z_{k+1}
=
\begin{cases}
2z_k, & 0\leq z_k<0.5,\\
2(1-z_k), & 0.5\leq z_k\leq 1,
\end{cases}
\end{equation}
where $z_k\in[0,1]$ denotes the chaotic variable generated at the $k$-th iteration. A sequence containing
$N^{(\mathsf{p})}N^{(\mathsf{task})}$ chaotic variables is generated and assigned to the initial priority vectors according to
\begin{equation}
x_{o,s}^{(0)}
=
z_{(o-1)N^{(\mathsf{task})}+s},
\quad
1\leq o\leq N^{(\mathsf{p})},
\quad
1\leq s\leq N^{(\mathsf{task})}.
\end{equation}
Under its invariant measure, the standard Tent map exhibits a uniform density over $[0,1]$. Its piecewise-expanding dynamics are therefore employed to promote a broadly dispersed set of initial priority weights and task-ranking patterns. This initialization improves the coverage of the priority search space at the beginning of the evolutionary process and reduces the likelihood that $\rm ES^2A$ prematurely concentrates on a limited set of similar task-priority structures.


\noindent
$\bullet$ \textbf{SubM B.2: Hierarchical heuristic decoding and evaluation.}
The priority sequence obtained in SubM B.1 specifies only the order in which offline tasks are considered and does not directly constitute an executable multi-UAV schedule. In particular, the resulting task assignments and trajectories must jointly satisfy the task active-time-window constraints, UAV energy budgets, three-dimensional motion constraints, obstacle-avoidance requirements, and collision-risk constraints. Therefore, we design a feasibility-preserving hierarchical heuristic decoder that maps each priority-encoded individual to a set of executable UAV task assignments and three-dimensional paths. For each task examined according to the priority sequence, the decoder evaluates the current states of all UAVs and successively determines three coupled decisions: joint obstacle avoidance and flight-altitude selection, data-collection altitude selection, and adaptive discrete speed selection. Whenever heterogeneous quantities, such as sensing quality, ecological impact, energy consumption, and delay, are combined into a scalar score, their normalized dimensionless values are used; for notational simplicity, the corresponding normalization symbols are omitted below.

\noindent
\emph{Strategy 1: Joint obstacle avoidance and flight-altitude decision.}
When UAV $u_m$ flies from PoI $p_i$ to PoI $p_{i'}$, a proper cruising altitude must be selected for the corresponding flight segment\footnote{Here, $p_i$ and $p_{i'}$ generically denote PoIs without distinguishing their types, because Module B performs offline pre-planning only for Type-$\mathsf{I}$ PoIs and the predicted task instances of Type-$\mathsf{II}$ PoIs.}. As discussed in Sec.~\ref{sec: modeling and problem}, the altitude decision involves a trade-off between the ecological penalty associated with low-altitude operation and the additional cost induced by obstacle avoidance and vertical altitude adjustment. Consider a flight segment from $(x_i,y_i,h_i)$ to $(x_{i'},y_{i'},h_{i'})$. We first identify the obstacles whose horizontal projections intersect the straight line connecting the two PoIs and collect them into the set $\mathbbm{b}^{(\mathsf{path})}_{i,i'}$. When $\mathbbm{b}^{(\mathsf{path})}_{i,i'}\neq\emptyset$, let
$h^{(\mathsf{obs})}_{i,i'}$ denote the maximum height among the obstacles in this set. For a candidate cruising altitude $h\in[h^{(\mathsf{min})},h^{(\mathsf{max})}]$, let
$c_{m,i,i'}^{(\mathsf{obs})}(h)$ denote the corresponding obstacle-related cost, including the cost of horizontal detouring and vertical altitude adjustment. The continuous altitude-selection problem can be expressed as
\begin{equation}
\label{heightobj}
h_{m}^{(\mathsf{uav},t)*}
=
\arg\min_{h\in[h^{(\mathsf{min})},h^{(\mathsf{max})}]}
\left(
\gamma_1 f^{(\mathsf{eco})}(h)
+
\gamma_2 c_{m,i,i'}^{(\mathsf{obs})}(h)
\right),
\end{equation}
where $\gamma_1\geq 0$ and $\gamma_2\geq 0$ are weighting coefficients that balance the altitude-dependent ecological penalty and the obstacle-related cost, respectively. Here, $h_{m}^{(\mathsf{uav},t)*}$ represents the cruising altitude selected for the considered segment. The exact ecological penalty accumulated over the corresponding flight timeslots is subsequently evaluated when calculating the decoded solution's net reward.

Directly solving (\ref{heightobj}) for every UAV-task pair may introduce substantial computational overhead because both the ecological penalty and the obstacle-related cost can be nonlinear and route-dependent. We therefore transform the continuous altitude-selection problem into a finite comparison among several representative altitude strategies. Specifically, the following candidate altitudes are considered; for simplicity, the indices $i$ and $i'$ are omitted from the altitude symbols.

\begin{itemize}
    \item \textit{Current-or-target altitude $h^{(\mathsf{ct})}$}: $h^{(\mathsf{ct})}=\max(h_i,h_{i'})$. This candidate avoids introducing an intermediate cruising altitude higher than that required by the two segment endpoints. It therefore provides a baseline strategy with limited additional vertical motion, subject to obstacle feasibility.

    \item \textit{Physical overflight altitude $h^{(\mathsf{limit})}$}: $h^{(\mathsf{limit})}=h^{(\mathsf{obs})}_{i,i'}
    +\delta^{(\mathsf{safe})}$, where $\delta^{(\mathsf{safe})}$ denotes the required vertical safety margin. When $\mathbbm{b}^{(\mathsf{path})}_{i,i'}\neq\emptyset$, this candidate represents the minimum altitude at which the identified obstacles can be directly overflown without horizontal detouring. This candidate is discarded if it exceeds $h^{(\mathsf{max})}$.

    \item \textit{Ecological escape altitude $h^{(\mathsf{eco})}$}: For the continuous ecological penalty function $f^{(\mathsf{eco})}(h)$ in (\ref{eq:penalty}), $h^{(\mathsf{eco})}$ is selected such that $f^{(\mathsf{eco})}\left(h^{(\mathsf{eco})}\right)=\varkappa^{(\mathsf{eco})}$, where $\varkappa^{(\mathsf{eco})}$ is a prescribed low-penalty threshold. This strategy permits additional vertical movement in exchange for reducing the altitude-dependent ecological penalty to a relatively small level.

    \item \textit{Ecological midpoint altitude $h^{(\mathsf{mid})}$}: The altitude $h^{(\mathsf{mid})}$ is selected according to $f^{(\mathsf{eco})}\left(h^{(\mathsf{mid})}\right)=\varkappa^{(\mathsf{mid})}$, where $\varkappa^{(\mathsf{mid})}$ denotes an intermediate penalty threshold. For the sigmoid function adopted in (\ref{eq:penalty}), setting
    $\varkappa^{(\mathsf{mid})}=\varepsilon/2$ yields
    $h^{(\mathsf{mid})}=h^{(0)}$, i.e., the inflection-point altitude. This candidate provides a compromise between excessive vertical climbing and a high ecological penalty.
\end{itemize}

\noindent
After removing duplicated and infeasible values, the candidate-altitude set is given by
\begin{equation}
\mathcal{H}
=
\operatorname{unique}
\left(
\left\{
h^{(\mathsf{ct})},
h^{(\mathsf{limit})},
h^{(\mathsf{eco})},
h^{(\mathsf{mid})}
\right\}
\cap
[h^{(\mathsf{min})},h^{(\mathsf{max})}]
\right),
\end{equation}
where $h^{(\mathsf{limit})}$ is omitted when
$\mathbbm{b}^{(\mathsf{path})}_{i,i'}=\emptyset$. For each $h\in\mathcal{H}$, the decoder examines whether the direct horizontal route is blocked by any obstacle whose height, including the required safety margin, exceeds $h$. If the direct route is blocked, a horizontal Manhattan detour is constructed and its additional distance is denoted by $D^{(\mathsf{mht})}(h)$; otherwise, the UAV follows the straight horizontal route from $p_i$ to $p_{i'}$. The resulting horizontal detour energy is denoted by
$c^{(\mathsf{obs},\mathsf{lvl})}_{m,i,i'}(h)$. Meanwhile, the vertical-motion cost
$c^{(\mathsf{obs},\mathsf{vtl})}_{m,i,i'}(h)$ accounts for the altitude changes from $h_i$ to $h$ and from $h$ to $h_{i'}$. Accordingly, the overall obstacle-related cost is
\begin{equation}
c_{m,i,i'}^{(\mathsf{obs})}(h)
=
c_{m,i,i'}^{(\mathsf{obs},\mathsf{lvl})}(h)
+
c_{m,i,i'}^{(\mathsf{obs},\mathsf{vtl})}(h).
\end{equation}
The corresponding altitude-selection score is
\begin{equation}
J_{m,i,i'}^{(\mathsf{alt})}(h)
=
\gamma_1 f^{(\mathsf{eco})}(h)
+
\gamma_2 c_{m,i,i'}^{(\mathsf{obs})}(h).
\end{equation}
If neither direct overflight nor horizontal detouring is feasible at altitude $h$, we set
$J_{m,i,i'}^{(\mathsf{alt})}(h)=+\infty$. The cruising altitude for the considered segment is then selected as
\begin{equation}
h^{*}
=
\arg\min_{h\in\mathcal{H}}
J_{m,i,i'}^{(\mathsf{alt})}(h).
\end{equation}
If all candidate altitudes have infinite costs, the corresponding UAV-task transition is regarded as infeasible.

\noindent
\textit{Strategy 2: Optimal data-collection altitude decision.}
After UAV $u_m$ arrives at $p_{i'}$, it must further determine an appropriate altitude for data collection. Unlike the cruising-altitude decision, the collection-altitude decision must explicitly account for the sensing quality
$q_m^{(\mathsf{uav},t)}$, because operating closer to the target may improve data quality while simultaneously increasing the ecological penalty and the cost of changing altitude from the selected cruising altitude $h^{*}$.

Let $h_{i'}^{(\mathsf{uav},\mathsf{min})}$ denote the minimum permissible collection altitude associated with $p_{i'}$. To limit the computational complexity, the decoder performs a one-dimensional grid search with step size $l^{(\mathsf{search})}$ over the candidate set
\begin{equation}
\mathcal{H}_{i'}^{(\mathsf{collect})}
=\bigl\{
h_{i'}^{(\mathsf{uav},\mathsf{min})}+k l^{(\mathsf{search})}
\;\bigm|\;
k\in\mathbb{Z}_{\geq 0},\;
h_{i'}^{(\mathsf{uav},\mathsf{min})}+k l^{(\mathsf{search})}
\leq\min\left(h^{(\mathsf{eco})},h^{(\mathsf{max})}\right)
\bigr\}.
\end{equation}
If
$h_{i'}^{(\mathsf{uav},\mathsf{min})}>h^{(\mathsf{eco})}$,
the minimum permissible altitude is retained as the only candidate, provided that it does not exceed $h^{(\mathsf{max})}$. For a candidate collection altitude $h$, let
$c_{m,i'}^{(\mathsf{chg})}(h\mid h^{*})$ denote the additional cost associated with changing from the cruising altitude $h^{*}$ to $h$, including the corresponding vertical-motion energy and time. The collection-altitude utility is
\begin{equation}
J_{m,i'}^{(\mathsf{collect})}(h)
=
q_m^{(\mathsf{uav},t)}(h)
-
f^{(\mathsf{eco})}(h)
-
c_{m,i'}^{(\mathsf{chg})}(h\mid h^{*}).
\end{equation}
The optimal collection altitude is therefore selected as
\begin{equation}
h^{(\mathsf{collect})*}
=
\arg\max_{h\in\mathcal{H}_{i'}^{(\mathsf{collect})}}
J_{m,i'}^{(\mathsf{collect})}(h).
\end{equation}
This grid-search procedure reduces the complexity of the continuous altitude-decision problem while explicitly balancing sensing quality, ecological impact, and altitude-adjustment cost. The pseudocode corresponding to Strategies 1 and 2 is presented in Alg.~\ref{alg:high} in \ref{pseudocodes}.

\noindent
\textit{Strategy 3: Adaptive discrete speed selection.}
After determining the cruising and data-collection altitudes, the decoder selects the flight speed toward the target task. For computational tractability, the candidate horizontal speeds are discretized as
\begin{equation}
\mathcal{V}
=
\left\{
v^{(\mathsf{slow})},
v^{(\mathsf{mid})},
v^{(\mathsf{fast})}
\right\}.
\end{equation}
Candidate speeds that violate the UAV's instantaneous speed or acceleration constraints are removed before evaluation, yielding the feasible speed set
$\mathcal{V}_{m,i,i'}^{(\mathsf{feas})}\subseteq\mathcal{V}$.

For the flight segment from $p_i$ to $p_{i'}$, let $d_{i,i'}$ denote the horizontal route distance after accounting for the selected obstacle-avoidance strategy, and let
$T_{m,i,i'}^{(\mathsf{vtl})}$ denote the time required for the vertical movements determined by Strategies 1 and 2. If UAV $u_m$ becomes available at timeslot $t_m$, its predicted arrival timeslot under speed $v$ is
\begin{equation}
\mathbbm{t}_m^{(\mathsf{arr},v)}
=
t_m
+
\left\lceil
\frac{
d_{i,i'}/v
+
T_{m,i,i'}^{(\mathsf{vtl})}
}{
\Delta t
}
\right\rceil.
\end{equation}
Moreover, $\mathbbm{t}_{i'}^{(\mathsf{start})}$ and
$\mathbbm{t}_{i'}^{(\mathsf{end})}$ denote the starting timeslot and deadline of the task associated with $p_{i'}$, respectively. For a Type-$\mathsf{I}$ task, these values correspond to the beginning and end of the mission horizon, whereas for a predicted Type-$\mathsf{II}$ task, they correspond to its predicted active time window.

The decoder permits a UAV to arrive after the task starting timeslot, provided that a positive service interval remains before the deadline. The resulting reduction in the UAV's individual data contribution may subsequently be compensated through cooperative sensing by multiple UAVs. In contrast, arriving at or after the task deadline leaves no available sensing time and is therefore infeasible. Accordingly, the time-violation cost is defined as
\begin{equation}
C^{(\mathsf{time})}_{m,i,i'}(v)
=
\begin{cases}
0,
&
\mathbbm{t}_m^{(\mathsf{arr},v)}
\leq
\mathbbm{t}_{i'}^{(\mathsf{start})},
\\
\omega
\left(
\mathbbm{t}_m^{(\mathsf{arr},v)}
-
\mathbbm{t}_{i'}^{(\mathsf{start})}
\right),
&
\mathbbm{t}_{i'}^{(\mathsf{start})}
<
\mathbbm{t}_m^{(\mathsf{arr},v)}
<
\mathbbm{t}_{i'}^{(\mathsf{end})},
\\
+\infty,
&
\mathbbm{t}_m^{(\mathsf{arr},v)}
\geq
\mathbbm{t}_{i'}^{(\mathsf{end})},
\end{cases}
\end{equation}
where $\omega$ is a nonnegative coefficient that converts delayed arrival into the same cost scale as the energy-related term.

Let $E_{m,i,i'}^{(\mathsf{vtl})}$ denote the vertical-motion energy associated with the selected cruising and collection altitudes. The energy cost under candidate speed $v$ includes the horizontal flight energy, vertical-motion energy, and the hovering energy caused by early arrival:
\begin{equation}
\begin{split}
C_{m,i,i'}^{(\mathsf{energy})}(v)
&= e_m^{(\mathsf{mov},v)} \frac{d_{i,i'}}{v} + E_{m,i,i'}^{(\mathsf{vtl})} \\
&\quad + e_m^{(\mathsf{hov})} \max\left( \mathbbm{t}_{i'}^{(\mathsf{start})} - \mathbbm{t}_m^{(\mathsf{arr},v)}, 0 \right).
\end{split}
\end{equation}
Here, $e_m^{(\mathsf{mov},v)}$ denotes the movement-energy rate of UAV $u_m$ under speed $v$, while $e_m^{(\mathsf{hov})}$ denotes its hovering-energy consumption per timeslot. The speed-selection cost is 
\begin{equation}
J_{m,i,i'}(v)
=
C_{m,i,i'}^{(\mathsf{energy})}(v)
+
C_{m,i,i'}^{(\mathsf{time})}(v).
\end{equation}
The decoder traverses all feasible candidate speeds and selects
\begin{equation}
v^{*}
=
\arg\min_{v\in\mathcal{V}_{m,i,i'}^{(\mathsf{feas})}}
J_{m,i,i'}(v).
\end{equation}
When sufficient time remains before the task deadline, the decoder selects the speed that minimizes the combined movement and hovering energy. As the available service interval becomes shorter, a higher speed may be preferred to reduce the delayed-arrival cost. If
$\mathcal{V}_{m,i,i'}^{(\mathsf{feas})}=\emptyset$
or all candidate speeds produce infinite costs, the corresponding UAV-task transition is regarded as infeasible.

After completing the above three strategies, the decoder evaluates the feasibility and incremental execution cost of assigning the current task to each UAV. Specifically, when processing a task in the priority sequence
$\boldsymbol{\pi}_{o}^{(g)}$, the decoder reads the current fleet state
\begin{equation}
\label{WT}
\mathcal{W}(t)
=
\left\{
(\mathbf{l}_m,t_m,E_m)
\mid
u_m\in\mathcal{U}
\right\},
\end{equation}
where $\mathbf{l}_m$, $t_m$, and $E_m$ denote the current location, earliest available service timeslot, and remaining energy of UAV $u_m$, respectively. For each UAV, the decoder applies Strategies 1--3 to determine the cruising altitude, collection altitude, obstacle-avoidance route, and flight speed. A UAV is included in the feasible candidate set only if it can reach the target before the task deadline, satisfy the minimum-altitude and obstacle-avoidance constraints, respect the UAV collision-risk constraint, complete the required vertical and horizontal movements within its kinematic limits, and retain sufficient energy to execute the task and subsequently reach its designated destination. For each feasible UAV, the decoder further estimates the maximum amount of data that can be collected during the remaining portion of the task's active time window. The corresponding incremental execution cost includes the movement and hovering energy, altitude-adjustment cost, and ecological penalty accumulated along the generated trajectory. UAVs with finite incremental costs are ranked in ascending order to form a candidate list. The decoder first attempts to satisfy the task's data requirement using a single UAV. If no individual UAV can provide the required data volume, additional UAVs are sequentially selected from the candidate list, subject to the collision-risk constraint, until their aggregate data contribution satisfies the task requirement. The task assignment is committed only when the selected UAV or cooperative UAV group can collectively complete the task. Otherwise, the task is skipped and the fleet state remains unchanged, thereby preventing infeasible partial assignments. Once a task is successfully assigned, the locations, available timeslots, remaining energy budgets, and planned paths of the selected UAVs are updated before the decoder proceeds to the next task in
$\boldsymbol{\pi}_{o}^{(g)}$. Under a fixed tie-breaking rule, this decoding process deterministically maps each priority-encoded individual to a feasible set of task assignments and UAV trajectories. After all tasks in the priority sequence have been examined, the decoded solution is evaluated using the net reward in (\ref{eq:reward_net}). The resulting value defines the fitness of the corresponding individual and is subsequently used by the evolutionary update mechanism in SubM B.3.


\noindent
$\bullet$ \textbf{SubM B.3: Adaptive evolution integrated with Lévy flight.}

\noindent
\textit{Strategy 1: Lévy-flight-assisted adaptive evolutionary update.}
After each individual is hierarchically decoded and evaluated according to the net reward in (\ref{eq:reward_net}), $\rm ES^2A$ performs evolutionary updates in the task-priority space. Recall that the decoder in SubM B.2 depends primarily on the relative ordering of the priority weights rather than on their absolute magnitudes. Therefore, multiplying all entries of a priority vector by the same positive scalar does not alter its induced task-priority sequence and, consequently, cannot generate a new decoded solution. To ensure effective exploration of different task orderings, all evolutionary operators below apply dimension-wise and nonuniform perturbations to the priority vectors.

At the beginning of generation $g$, all individuals\footnote{Each individual is represented by a unique task-priority vector. Therefore, the terms ``individual'' and ``priority vector'' are used interchangeably when describing the evolutionary operators. Nevertheless, multiple distinct priority vectors may induce the same task-priority sequence and decoded solution.} are ranked in descending order according to their fitness values
$F(\mathbf{x}_{o}^{(g)})$ obtained through the decoding and evaluation procedure in SubM B.2. Let
$\mathbf{x}^{(\mathsf{best},g)}$ denote the historical best priority vector identified up to generation $g$. According to the predefined discoverer ratio
$\rho^{(\mathsf{f})}\in(0,1)$, the top
$\lfloor\rho^{(\mathsf{f})}N^{(\mathsf{p})}\rfloor$
individuals are designated as discoverers, while the remaining individuals are treated as joiners. We let $D=N^{(\mathsf{task})}$ denote the dimension of each priority vector. Since the priority weights must remain in $[0,1]$, we define the element-wise projection operator $\Pi_{[0,1]^D}(\mathbf{y})=
\left[
\min\{1,\max\{0,y_1\}\},
\ldots,
\min\{1,\max\{0,y_D\}\}
\right]^{\top}$,
which is applied after every evolutionary update. Moreover, the generation-dependent attenuation coefficient is defined as $a_g=1-\frac{g}{G^{(\mathsf{max})}}$, where $G^{(\mathsf{max})}$ denotes the maximum number of generations. As the evolutionary process proceeds, $a_g$ gradually decreases, enabling broader exploration during the early generations and more localized refinement during the later generations.

\noindent
\emph{(i) Discoverer update.}
Discoverers correspond to relatively high-fitness individuals and are responsible for exploiting promising task-priority structures while guiding the population toward high-quality regions. For each discoverer, a scalar
$r_o\sim{\rm U}(0,1)$, a dimension-wise random vector
$\mathbf{r}_o\sim{\rm U}(0,1)^D$, and a Gaussian perturbation vector
$\mathbf{z}_o\sim\mathcal{N}(\mathbf{0},\mathbf{I}_D)$
are independently generated. The discoverer is updated according to
\begin{equation}
\widetilde{\mathbf{x}}_{o}^{(g+1)}
=\begin{cases}
\Pi_{[0,1]^D}\left[
\mathbf{x}_{o}^{(g)}+a_g\mathbf{r}_o \odot (\mathbf{x}^{(\mathsf{best},g)}-\mathbf{x}_{o}^{(g)})+\sigma a_g\mathbf{z}_o
\right], & r_o<S^{(\mathsf{st})}, \\[6pt]
\Pi_{[0,1]^D}\left[
\mathbf{x}_{o}^{(g)}+\varepsilon a_g\mathbf{z}_o
\right], & r_o\geq S^{(\mathsf{st})},
\end{cases}
\end{equation}
where $S^{(\mathsf{st})}\in(0,1)$ denotes the warning threshold,
$\sigma$ controls the local perturbation around the best priority structure, and $\varepsilon$ controls the exploratory Gaussian perturbation. When
$r_o<S^{(\mathsf{st})}$, the discoverer performs a best-guided local search while retaining a small stochastic perturbation to avoid becoming trapped on a fitness plateau. When
$r_o\geq S^{(\mathsf{st})}$, a relatively less directed perturbation is applied to explore alternative task-priority structures. Unlike uniform scaling, the dimension-wise operations in both cases may alter the relative ordering of priority weights and therefore generate different task-consideration sequences for decoding.

\noindent
\emph{(ii) Joiner update.}
Joiners perform complementary search around the promising regions identified by discoverers. After the population is sorted, the joiners are further divided into two subgroups according to adaptive rank threshold $\tau^{(\mathsf{f})}=\left\lfloor\frac{\left(1+\rho^{(\mathsf{f})}\right)N^{(\mathsf{p})}}{2}\right\rfloor$.
The relatively high-fitness joiners refine solutions near the historical best, whereas the lower-fitness joiners conduct long-range exploration through Lévy-flight perturbations. For notational compactness, define
$N^{(\mathsf{f})}
\triangleq
\lfloor
\rho^{(\mathsf{f})}N^{(\mathsf{p})}
\rfloor$
and
$\mathbf{d}_{o}^{(g)}
\triangleq
\mathbf{x}_{o}^{(g)}
-
\mathbf{x}^{(\mathsf{best},g)}$. Their update rule is
\begin{equation}
\label{eq:update}
\widetilde{\mathbf{x}}_{o}^{(g+1)}
=\begin{cases}
\Pi_{[0,1]^D}\left[
\mathbf{x}^{(\mathsf{best},g)}+a_g\left(\boldsymbol{\eta}_o-\tfrac{1}{2}\mathbf{1}\right)\odot\mathbf{d}_{o}^{(g)}+\sigma a_g\mathbf{z}_o
\right], & \text{if } N^{(\mathsf{f})}<o\leq\tau^{(\mathsf{f})}, \\[6pt]
\Pi_{[0,1]^D}\left[
\mathbf{x}^{(\mathsf{best},g)}+\zeta a_g\mathbf{L}_{o}^{(g)}(D)
\right], & \text{if } o>\tau^{(\mathsf{f})}.
\end{cases}
\end{equation}
where
$\tau^{(\mathsf{f})}
=
\left\lfloor
(1+\rho^{(\mathsf{f})})N^{(\mathsf{p})}/2
\right\rfloor$
is the adaptive rank threshold used to divide the joiners into local-refinement and global-exploration subgroups.
Moreover,
$\boldsymbol{\eta}_o\sim{\rm U}(0,1)^D$
is a dimension-wise random vector,
$\mathbf{1}$ is the $D$-dimensional all-one vector,
$\mathbf{z}_o\sim\mathcal{N}(\mathbf{0},\mathbf{I}_D)$
is a standard Gaussian perturbation vector,
and $\odot$ denotes the element-wise product.
The coefficient $\zeta$ controls the step size of the Lévy-flight update, while
$\mathbf{L}_{o}^{(g)}(D)$
denotes a $D$-dimensional Lévy-flight random vector generated using the Mantegna algorithm, whose components follow a heavy-tailed Lévy distribution \cite{2025complex}.


For the first subgroup, the signed difference
$\mathbf{x}_{o}^{(g)}-\mathbf{x}^{(\mathsf{best},g)}$
is retained so that the update reflects the direction of each priority dimension relative to the current best solution. The dimension-wise coefficient
$\boldsymbol{\eta}_o-\frac{1}{2}\mathbf{1}$
generates small perturbations on both sides of the best priority vector, thereby supporting local refinement without forcing all joiners to collapse onto an identical structure. For the second subgroup, the heavy-tailed Lévy perturbation occasionally produces relatively large changes across selected priority dimensions, enabling low-fitness individuals to explore task orderings beyond the current local region. Therefore, the two joiner updates jointly balance local exploitation and long-range exploration in the priority space. The basic discoverer--joiner update is summarized in Alg.~\ref{alg:levy_basic_evolution_short} in \ref{pseudocodes}.

\noindent
\textit{Strategy 2: Danger-aware diversification and elitist preservation.}
Although the discoverer--joiner mechanism provides both local and global search capabilities, the indirect priority representation and hierarchical decoder may induce a piecewise-constant and multimodal fitness landscape. In particular, multiple priority vectors may yield the same task-priority sequence or be mapped by the decoder to the same feasible assignment and path set. Consequently, the population may become concentrated around a limited number of equivalent priority structures, while the historical best fitness may remain unchanged over multiple generations.

To detect such convergence degradation, we jointly monitor the diversity of the induced task-priority sequences and the stagnation duration of the historical best. Let
$\boldsymbol{\pi}_{o}^{(g)}$ denote the task-priority sequence induced by
$\mathbf{x}_{o}^{(g)}$, as defined in SubM B.1, and let
$\boldsymbol{\pi}^{(\mathsf{best},g)}$ denote the sequence induced by
$\mathbf{x}^{(\mathsf{best},g)}$. The population diversity at generation $g$ is measured as
\begin{equation}
\mathcal{D}^{(g)}
=
\frac{1}{N^{(\mathsf{p})}}
\sum_{o=1}^{N^{(\mathsf{p})}}
d_{\mathsf{K}}
\left(
\boldsymbol{\pi}_{o}^{(g)},
\boldsymbol{\pi}^{(\mathsf{best},g)}
\right),
\end{equation}
where
$d_{\mathsf{K}}(\cdot,\cdot)\in[0,1]$
denotes the normalized Kendall distance between two task-priority sequences. A smaller
$\mathcal{D}^{(g)}$
indicates that a larger portion of the population has converged to task orderings similar to the current best. Meanwhile, let $s^{(g)}$ denote the number of consecutive generations without a meaningful improvement in the historical best fitness. It is updated by
\begin{equation}
\hspace{-.35cm}s^{(g)}
=
\begin{cases}
0,
&
F\left(
\mathbf{x}^{(\mathsf{best},g)}
\right)
>
F\left(
\mathbf{x}^{(\mathsf{best},g-1)}
\right)
+
\epsilon_F,
\\
s^{(g-1)}+1,
&
\text{otherwise},
\end{cases}
\end{equation}
where $\epsilon_F>0$ is a numerical tolerance used to distinguish an effective fitness improvement from numerical fluctuations. The danger-aware mechanism is activated when
\begin{equation}
\mathcal{D}^{(g)}
\leq
\mathcal{D}^{(\mathsf{min})}
\quad
\text{or}
\quad
s^{(g)}
\geq
S^{(\mathsf{stall})},
\end{equation}
where
$\mathcal{D}^{(\mathsf{min})}$
is the minimum acceptable population-diversity threshold and
$S^{(\mathsf{stall})}$
is the maximum tolerated number of stagnant generations.

Once the mechanism is activated, $N^{(\mathsf{d})}$ non-elite individuals are selected to form the danger-aware set
$\mathcal{K}_g$. The historical best individual is excluded from
$\mathcal{K}_g$ to prevent the loss of the best solution found so far. For each selected individual
$o\in\mathcal{K}_g$, the basic-update result
$\widetilde{\mathbf{x}}_{o}^{(g+1)}$
is further adjusted according to
\begin{equation}
\label{eq:danger-aware-update}
\overline{\mathbf{x}}_{o}^{(g+1)}
=\begin{cases}
\Pi_{[0,1]^D}\left[
\widetilde{\mathbf{x}}_{o}^{(g+1)}+\boldsymbol{\epsilon}_o\odot\left(\mathbf{x}^{(\mathsf{best},g)}-\widetilde{\mathbf{x}}_{o}^{(g+1)}\right)
\right], & \text{if}~ F\left(\mathbf{x}_{o}^{(g)}\right)<F\left(\mathbf{x}^{(\mathsf{best},g)}\right)-\epsilon_F, \\[6pt]
\Pi_{[0,1]^D}\left[
\widetilde{\mathbf{x}}_{o}^{(g+1)}+\delta a_g\boldsymbol{\xi}_o
\right], & \text{if}~ \left| F\left(\mathbf{x}_{o}^{(g)}\right)-F\left(\mathbf{x}^{(\mathsf{best},g)}\right)\right|\leq\epsilon_F,
\end{cases}
\end{equation}
where
$\boldsymbol{\epsilon}_o\sim{\rm U}(0,1)^D$,
$\boldsymbol{\xi}_o\sim\mathcal{N}(\mathbf{0},\mathbf{I}_D)$,
and $\delta$ is the perturbation-amplitude coefficient. For individuals whose fitness is clearly lower than the historical best, the first update performs a dimension-wise interpolation toward the best priority vector, thereby strengthening exploitation while preserving part of the individual's original structure. For individuals located on the same or a numerically indistinguishable fitness plateau as the historical best, the second update introduces a controlled perturbation to generate alternative priority orderings and reduce population duplication. If the danger-aware condition is not activated, or if
$o\notin\mathcal{K}_g$, we set
\begin{equation}
\overline{\mathbf{x}}_{o}^{(g+1)}
=
\widetilde{\mathbf{x}}_{o}^{(g+1)}.
\end{equation}

After the danger-aware adjustment, every updated priority vector
$\overline{\mathbf{x}}_{o}^{(g+1)}$
is decoded using the hierarchical decoder in SubM B.2, and its fitness is recalculated according to (\ref{eq:reward_net}). The historical best is then updated by
\begin{equation}
\mathbf{x}^{(\mathsf{best},g+1)}
=
\arg\max_{
\mathbf{x}
\in
\left\{
\mathbf{x}^{(\mathsf{best},g)}
\right\}
\cup
\left\{
\overline{\mathbf{x}}_{1}^{(g+1)},
\ldots,
\overline{\mathbf{x}}_{N^{(\mathsf{p})}}^{(g+1)}
\right\}
}
F(\mathbf{x}).
\end{equation}
To enforce elitist preservation, if
$\mathbf{x}^{(\mathsf{best},g)}$
is not retained among the updated population, it replaces the lowest-fitness updated individual before the next generation begins. This ensures that the best decoded solution found during the evolutionary process is never lost. By jointly employing rank-changing dimension-wise updates, Lévy-flight exploration, task-order-based diversity detection, danger-aware perturbation, and elitist preservation, $\rm ES^2A$ maintains effective exploration of alternative task-priority structures while progressively refining high-quality offline planning solutions. The corresponding danger-aware and fitness-update procedure is summarized in Alg.~\ref{alg:danger_awareness_update_short} in \ref{pseudocodes}.


\subsubsection{Module C. Greedy-empowered timely adjustment when Type-$\mathsf{III}$ PoIs have needs (Stage 2)} 
Since Type-$\mathsf{III}$ PoIs emerge without prior knowledge of their locations or activation times, Module C performs online adaptation based on the offline task assignments and UAV paths generated by Module B. Instead of re-triggering the global evolutionary optimization, Module C preserves the main structure of each UAV's remaining offline plan and performs a local path insertion when a Type-$\mathsf{III}$ task appears. Specifically, the module first identifies UAVs that can execute the emergent task without invalidating their remaining offline missions. It then constructs a limited number of cooperative UAV groups through a greedy procedure and evaluates them using a cooperative net-benefit metric. In this manner, Module C balances timely responses to emergent sensing demands with low online computational overhead and limited disruption to the pre-planned trajectories.

\noindent
$\bullet$~\textbf{SubM C.1: Construction of the task-feasible set (Alg.~\ref{alg:online_replanning_stage2},  \ref{pseudocodes}).}
Suppose that a Type-$\mathsf{III}$ PoI
$p_i^{(\mathsf{III},t)}
\in
\mathcal{P}^{(\mathsf{III},t)}$
emerges at timeslot $t$, with active time window
$[t,\mathbbm{t}_i^{(-,t)}]$
and data requirement
$d_i^{(\mathsf{III},t)}$.
Based on the current fleet state $\mathcal{W}(t)$ defined in (\ref{WT}), Module C examines whether each UAV can locally insert the emergent task into its remaining offline plan.

For UAV $u_m\in\mathcal{U}$, the insertion is first attempted at the earliest feasible position after its currently executed task. The UAV is redirected from its current state to the emergent PoI and subsequently reconnects to its remaining pre-planned path. The flight altitude, obstacle-avoidance route, and speed are determined using the trajectory-generation mechanisms introduced in SubM B.2. Let
$\tau_{m,i}^{(\mathsf{fly})}$
denote the resulting flight time from the current location of UAV $u_m$ to the emergent PoI. Its earliest arrival timeslot is
\begin{equation}
\mathbbm{t}_{m,i}^{(\mathsf{arr})}
=
t_m
+
\tau_{m,i}^{(\mathsf{fly})},
\end{equation}
where $t_m$ denotes the earliest timeslot at which UAV $u_m$ becomes available according to its current state in $\mathcal{W}(t)$.

Since the emergent task cannot be served before its activation, the earliest data-collection start time is
\begin{equation}
\mathbbm{t}_{m,i}^{(\mathsf{start})}
=
\max
\left\{
\mathbbm{t}_{m,i}^{(\mathsf{arr})},
t
\right\}.
\end{equation}
UAV $u_m$ is temporally feasible only if
$\mathbbm{t}_{m,i}^{(\mathsf{start})}
\leq
\mathbbm{t}_i^{(-,t)}$.
Because the active window includes both boundary timeslots, the effective collection duration is
\begin{equation}
\Delta\mathbbm{t}_{m,i}^{(\mathsf{eff})}
=
\left(
\mathbbm{t}_i^{(-,t)}
-
\mathbbm{t}_{m,i}^{(\mathsf{start})}
+
1
\right)
\Delta t.
\end{equation}
Accordingly, the maximum amount of data that UAV $u_m$ can contribute is estimated as
\begin{equation}
d_{m,i}^{(\mathsf{avail})}
=
d_m^{(\mathsf{uav})}
\Delta\mathbbm{t}_{m,i}^{(\mathsf{eff})},
\label{eq:online_available_data}
\end{equation}
where $d_m^{(\mathsf{uav})}$ denotes the data-collection capability of UAV $u_m$ per unit time. The task-feasibility concepts used in Module C are defined as follows.

\begin{itemize}
    \item \textit{Candidate feasible UAV set:}
    Let $e_{m,i}^{(\mathsf{req})}$ be the energy required for $u_m$ to reach/serve the emergent task, reconnect to its remaining offline path, complete the retained offline tasks, and finally reach its destination, and
    $\chi_{m,i}^{(\mathsf{rem})}\in\{0,1\}$
    be whether the remaining offline plan remains feasible after the local insertion, including its task time-window and kinematic constraints. The candidate feasible UAV set is $\mathcal{U}_i^{(\mathsf{cand})}=
    \left\{
    u_m\in\mathcal{U}
    \,\middle|\,
    \mathbbm{t}_{m,i}^{(\mathsf{start})}
    \leq
    \mathbbm{t}_i^{(-,t)},
    \;
    E_m
    \geq
    e_{m,i}^{(\mathsf{req})},
    \;
    \chi_{m,i}^{(\mathsf{rem})}=1
    \right\}$, with $E_m$ being the remaining energy of UAV $u_m$. The altitude, obstacle-avoidance, and collision-free trajectory requirements are checked when estimating
    $\tau_{m,i}^{(\mathsf{fly})}$,
    $e_{m,i}^{(\mathsf{req})}$, and
    $\chi_{m,i}^{(\mathsf{rem})}$.

    \item \textit{Feasible cooperative group:}
    A UAV group $G_i\subseteq\mathcal{U}_i^{(\mathsf{cand})}$ is admissible only if its cooperative operation satisfies the collision-risk constraint $\theta_i(G_i)\leq
    \theta^{(\mathsf{max})}$, where $\theta_i(G_i)$ denotes the UAV collision risk associated with the simultaneous execution of task
    $p_i^{(\mathsf{III},t)}$
    by the UAVs in $G_i$. Moreover, the group must collectively satisfy the data requirement: $\sum_{u_m\in G_i}
    d_{m,i}^{(\mathsf{avail})}
    \geq
    d_i^{(\mathsf{III},t)}$. A group satisfying both conditions is referred to as a \textit{feasible task-completion solution (FTCS)}.

    \item \textit{Greedy construction of candidate cooperative groups:}
    Exhaustively enumerating all subsets of
    $\mathcal{U}_i^{(\mathsf{cand})}$
    would incur exponential complexity. We therefore construct only a limited number of promising FTCSs through a multi-start greedy procedure. For each candidate UAV, define its preliminary contribution-to-cost score as $\psi_{m,i}=\frac{
    d_{m,i}^{(\mathsf{avail})}
    }{
    \widehat{C}_{m,i}^{(\mathsf{ins})}
    +
    \epsilon_C
    }$, where $\widehat{C}_{m,i}^{(\mathsf{ins})}$ is the estimated local-insertion cost, including additional flight, hovering, path-disruption, and ecological costs, and $\epsilon_C>0$ prevents division by zero.

    The candidate UAVs are first ranked in descending order of
    $\psi_{m,i}$. Each UAV is then used once as an initial seed. Starting from the seed, the remaining UAVs are examined according to the ranked list and greedily added if their inclusion preserves the collision-risk constraint. The augmentation terminates once the aggregate available data satisfies
    $d_i^{(\mathsf{III},t)}$.
    If the data requirement cannot be satisfied, the corresponding seed is discarded. After removing duplicate groups, all successfully constructed FTCSs form
    \begin{equation}
    \mathcal{G}_i
    =
    \left\{
    G_i^{(1)},
    G_i^{(2)},
    \ldots,
    G_i^{(g)},
    \ldots,
    G_i^{(|\mathcal{G}_i|)}
    \right\}.
    \end{equation}
\end{itemize}


\noindent
$\bullet$~\textbf{SubM C.2: Net-benefit evaluation and greedy assignment for candidate cooperative groups.}
After constructing $\mathcal{G}_i$, Module C evaluates whether serving the emergent task provides sufficient utility to justify the additional flight and disruption imposed on the offline plan. Let
$r_i^{(\mathsf{III},t)}$
denote the reward obtained upon successful completion of the task associated with
$p_i^{(\mathsf{III},t)}$.
For a candidate cooperative group
$G_i^{(g)}\in\mathcal{G}_i$,
the task reward is evenly attributed to its participating UAVs for bookkeeping purposes:
\begin{equation}
\widetilde{r}_{i}^{(\mathsf{III},t,g)}
=
\frac{
r_i^{(\mathsf{III},t)}
}{
|G_i^{(g)}|
}.
\end{equation}
For UAV $u_m\in G_i^{(g)}$, its individual incremental net benefit is defined as
\begin{equation}
\Delta J_{m,i}^{(\mathsf{III},t,g)}
=
\widetilde{r}_{i}^{(\mathsf{III},t,g)}
-
\underbrace{
\left(
C_{m,i}^{(\mathsf{fly})}
+
C_{m,i}^{(\mathsf{hov})}
+
C_{m,i}^{(\mathsf{dist})}
\right)
}_{\text{local-insertion cost}}
-
\underbrace{
C_{m,i}^{(\mathsf{eco})}
}_{\text{ecological cost}},
\label{eq:single_uav_profit}
\end{equation}
where
$C_{m,i}^{(\mathsf{fly})}$
denotes the additional flight cost incurred by redirecting UAV $u_m$ to the emergent PoI,
$C_{m,i}^{(\mathsf{hov})}$
denotes the hovering cost caused by early arrival or asynchronous cooperative execution, and
$C_{m,i}^{(\mathsf{dist})}$
quantifies the disruption introduced by deviating from and reconnecting to the original offline path. Moreover,
$C_{m,i}^{(\mathsf{eco})}$
denotes the change in altitude-dependent ecological penalty caused by the locally adjusted trajectory. Consistent with SubM B.2, these quantities are normalized to a common utility scale before aggregation.

The cooperative net benefit of group $G_i^{(g)}$ is
\begin{equation}
\begin{aligned}
\Delta J_i^{(\mathsf{III},t)}
\left(
G_i^{(g)}
\right)=
\sum_{u_m\in G_i^{(g)}}
\Delta J_{m,i}^{(\mathsf{III},t,g)}
\end{aligned}
\label{eq:team_profit}
\end{equation}
The second equality shows that equal reward sharing does not change the system-level reward; it only distributes the fixed task reward among participating UAVs. Different cooperative groups nevertheless yield different net benefits because of their distinct fleet sizes, current locations, available times, remaining energy, and levels of disruption to the offline plan.

The candidate cooperative group with the maximum net benefit is selected as
\begin{equation}
G_i^{(*)}
=
\arg\max_{G_i^{(g)}\in\mathcal{G}_i}
\Delta J_i^{(\mathsf{III},t)}
\left(
G_i^{(g)}
\right).
\label{eq:best_team}
\end{equation}
If
$\mathcal{G}_i=\emptyset$,
the task cannot be completed under the current fleet state and is therefore rejected. Otherwise, the local re-planning decision is accepted only if
\begin{equation}
\Delta J_i^{(\mathsf{III},t)}
\left(
G_i^{(*)}
\right)
>
0.
\end{equation}
In this case, the Type-$\mathsf{III}$ task is inserted at the earliest feasible position of each selected UAV's remaining route, and the corresponding UAV states and trajectories are updated. If
$\Delta J_i^{(\mathsf{III},t)}
(G_i^{(*)})\leq 0$,
the task is rejected even though it is physically feasible, because its reward is insufficient to compensate for the additional insertion and ecological costs.

Therefore, Module C responds to Type-$\mathsf{III}$ PoIs through a constrained local insertion rather than global trajectory re-optimization. By combining feasibility screening, polynomial-complexity greedy group construction, and exact net-benefit evaluation over the generated candidates, it provides timely online adaptation while preserving the feasibility and overall coherence of the offline plan. The corresponding procedure is summarized in Alg.~\ref{alg:profit_greedy_online} in \ref{pseudocodes}.

\begingroup
\raggedbottom
\section{Evaluations}
\label{experiments}

We conduct extensive experiments on both real-world and simulated datasets to validate the effectiveness of FORTUNE, as well as its scalability across varying problem sizes.

\subsection{Evaluation Metrics and Benchmark Methods}
\label{metrics and methods}
Since FORTUNE offers a unique aspect on the integration of offline pre-planning and online adjustment for heterogeneous PoIs. To better evaluate the performance, we involve the following metrics from two perspectives:


\noindent
$\bullet$~\textit{Metrics related to offline planning stage.} \textit{(i)} We first calculate \textit{offline fitness function (OffF2)} associated with Types-$\mathsf{I}$ and -$\mathsf{II}$ PoIs, as expressed by $\mathbb{R}^{(\mathsf{I,II},\mathsf{net})}=\sum_{i=1}^{|\mathbb{P}^{(\mathsf{I})}|} r_i^{(\mathsf{I})}
+ \sum_{i=1}^{|\mathbb{P}^{(\mathsf{II})}|} \sum_{a=1}^{A_i} r_i^{(\mathsf{II},a)}-\mathbbm{f}$, where $\mathbbm{f}$ denotes the negative impact only related to these two PoI types. \textit{(ii)} Then, we consider \textit{offline task completion rate (OffTCR)}, which depends on the ratio of the number of tasks successfully completed by UAVs during pre-planning to the total number of tasks (Types-$\mathsf{I}$ and -$\mathsf{II}$). These two metrics aim to validate the effectiveness of our Transformer-based forecasting and $\rm ES^2A$, making sure that the offline planning stage can get a good beginning. 

\noindent
$\bullet$~\textit{Metrics related to online adjustment stage.} \textit{(i)} We first discuss \textit{first-timeslot Type-$\mathsf{III}$ task completion ratio (F3TSR)}, reflecting the ratio between Type-$\mathsf{III}$ tasks successfully completed and the total generated Type-$\mathsf{III}$ tasks in the first timeslot (see the rationale on choosing F3TSR in \ref{F3TSR}). \textit{(ii)} Then, we calculate \textit{practical net reward (PrcNR)} cross the whole process (offline+online), given by (\ref{eq:reward_net}). These two metrics provide complementary perspectives for evaluating the effectiveness of online adjustment. While F3TSR characterizes the capability of responding to newly emerging online tasks, PrcNR reflects the overall outcome achieved through the joint offline planning and online adjustment process. Therefore, the combination of them enables a comprehensive evaluation of both instantaneous online responsiveness and end-to-end operational performance.

We next introduce different views on benchmark designs.

\noindent
$\bullet$~\textit{Pre-planning-related methods.} \textit{(i)} Pre-$\rm{CS^2A}$: Leveraging Transformer-based predictions, conventional SSA is subsequently adopted to perform solution-space exploration. \textit{(ii)} Pre-$\rm {GA}$: Following Transformer-based prediction, a genetic algorithm (GA) is employed to search for high-quality scheduling solutions. \textit{(iii)} Pre-$\rm {PSO}$: Following Transformer-based prediction, a particle swarm optimization (PSO) algorithm is employed to search for high-quality UAV paths. \textit{(iv)} Pre-$\rm {DisG}$: After obtaining the Transformer-based predictions, Pre-$\rm {DisG}$ assigns each activated task to the nearest available UAV according to spatial proximity, enabling immediate task service upon task arrival.

\noindent
$\bullet$~\textit{Purely online method.} To outline the advantages of PoI prediction, We involve a purely online method PureOn-DistG, rendering both Type-$\mathsf{I}$ and Type-$\mathsf{II}$ PoIs unknown to UAVs. In this case, task assignment follows a distance-greedy policy that prioritizes the nearest available PoI.

\noindent
$\bullet$~\textit{Online adjustment methods.} We further design representative online benchmarks for handling emergent Type-$\mathsf{III}$ PoIs. It is worth emphasizing that, unlike conventional settings, our UAVs have already pre-generated proper paths to service the former two types of PoIs. Consequently, when a Type-III PoI unexpectedly appears, UAVs should respond without compromising the feasibility of their pre-planned tasks. This significantly restricts the set of admissible online actions and leaves limited room for large-scale trajectory re-optimization. Under such real-time operational constraints, lightweight assignment policies constitute practical and widely adopted online response mechanisms. Accordingly, we consider the following representative baselines. \textit{(i)} On-DistG: Upon the arrival, each UAV independently evaluates whether the additional active task can be completed while still retaining sufficient energy to finish its offline pre-planned mission. If feasible, the UAV selects the nearest Type-$\mathsf{III}$ PoI for data collection under constraint (C2). \textit{(ii)} On-RewdG: When Type-$\mathsf{III}$ PoIs emerge, each UAV first verifies whether executing an additional sensing task preserves the feasibility of its offline pre-planned mission. Among all feasible Type-$\mathsf{III}$ PoIs, a UAV preferentially selects the one with the highest reward under (C2). If excessive UAV aggregation violates this constraint, the remaining UAVs are sequentially reassigned to the next highest-reward feasible PoIs. UAVs unable to satisfy the energy requirement remain idle during the current timeslot. \textit{(iii)} On-RandA: Once a Type-$\mathsf{III}$ appears, a UAV is randomly selected from the feasible candidate set ($U_i^{\mathsf{cand}}$), consisting of UAVs that can complete the task while maintaining sufficient energy for their pre-planned missions. The selected UAV performs the sensing task, whereas other UAVs continue executing original plans.

\subsection{Real-World UTD19 Dataset-driven Experiments}
This section uses the real-world UTD19 dataset\footnote{https://utd19.ethz.ch/}, which contains traffic-flow measurements collected from multiple cities worldwide between 2017 and 2019. Traffic data from Basel, Switzerland, on April 17, May 17, and June 17, 2019, are adopted for evaluation. As shown in Fig.~\ref{fig:map} as an example, selected sensing locations are categorized into Type-$\mathsf{I}$ (blue) and Type-$\mathsf{II}$ (orange) POIs. The predicted time windows of Type-$\mathsf{II}$ POIs are incorporated into the offline planning stage. To capture unforeseen sensing requests, additional locations are designated as Type-$\mathsf{III}$ POIs (red), with randomly generated occurrence times and service windows for online replanning.

\begin{figure}[!htbp] 
\centering
\subfloat[][]{\includegraphics[width=0.2\linewidth]{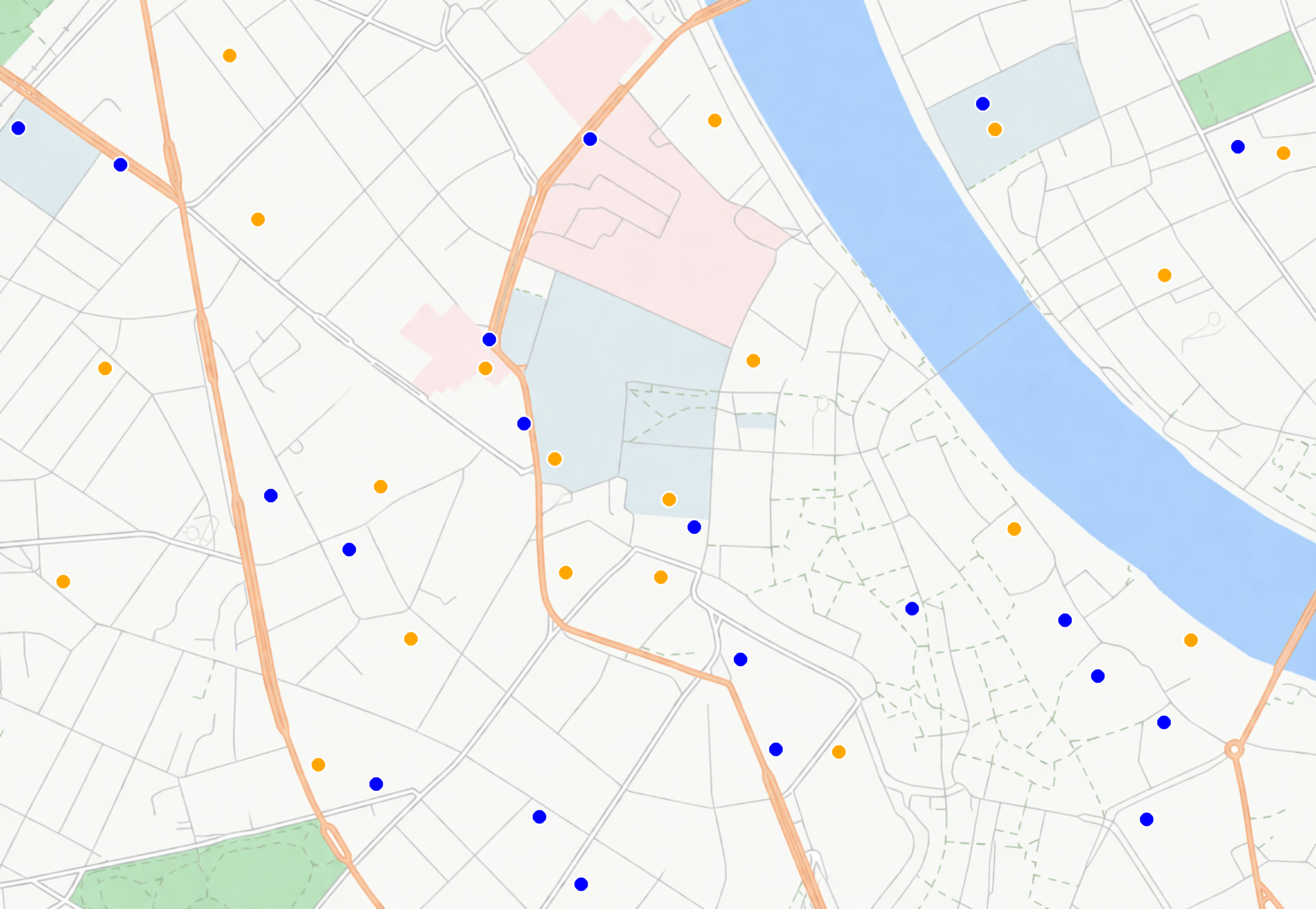}}%
~~~~~~~
\subfloat[][]{\includegraphics[width=0.2\linewidth]{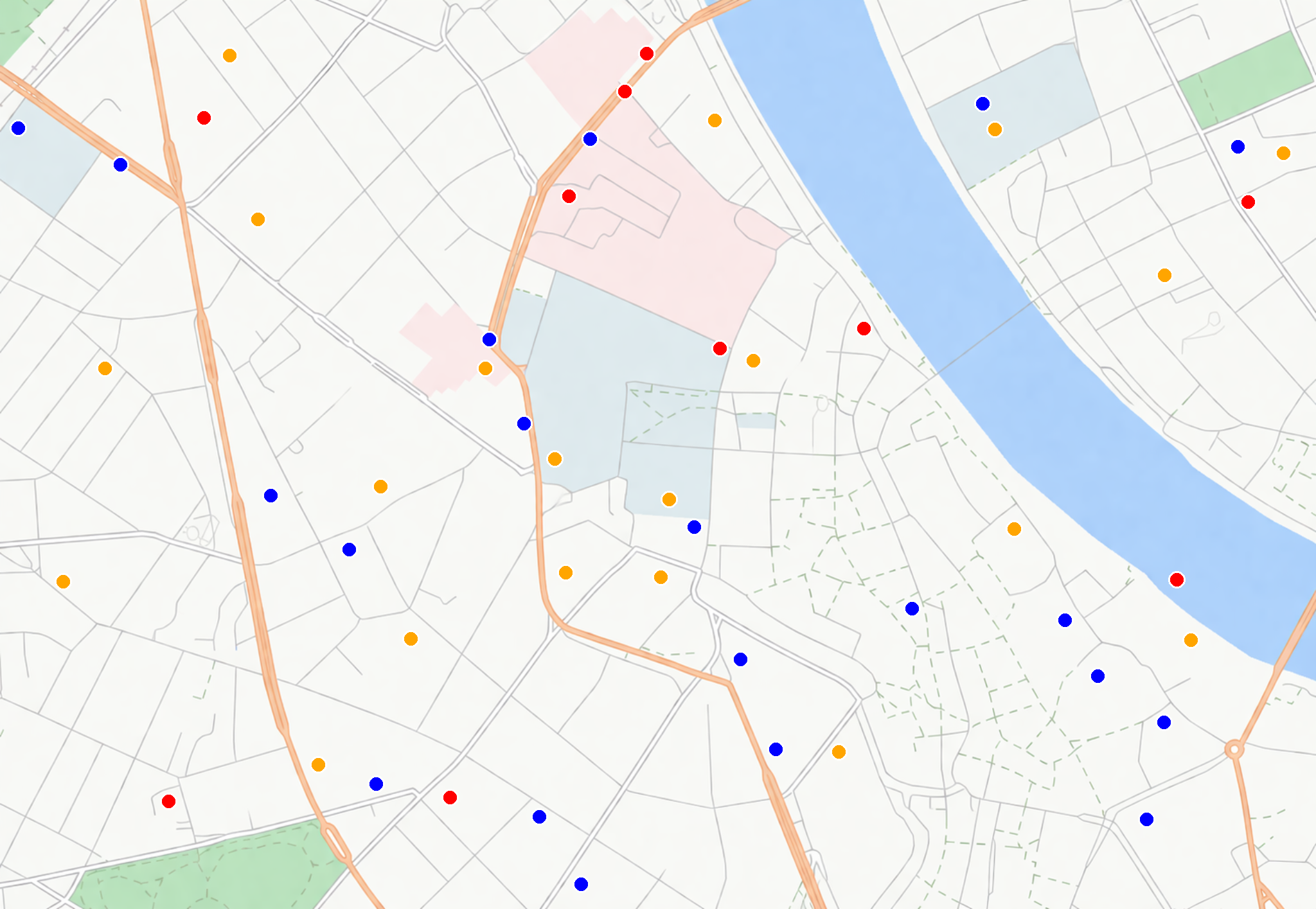}}
\caption{Schematic maps of PoIs related to offline pre-planning (a) and online replanning (b).}
\label{fig:map}
\end{figure}

\noindent
$\bullet$~\textit{Performance on PoI prediction.}
First, we conduct a comparative study of time-series forecasting models. Specifically, we compare the self-attention-based Transformer model (as our predictor) with the conventional long short-term memory (LSTM) network. The data is split into training, validation, and test sets at a ratio of 7:2:1, and the models are configured as detailed in Table \ref{transformerinfo} of \ref{parameter}. We adopt the mean squared error (MSE) as the loss function. Across all Type-$\mathsf{II}$ PoIs, the Transformer achieves an average MSE of 0.026, substantially lower than the 0.042 obtained by LSTM. Figure \ref{fig:Prediction} presents the predictions for three randomly selected PoIs.

\begin{figure*}[!htbp]
\centering
\subfloat[][]{\includegraphics[width=0.3\linewidth]{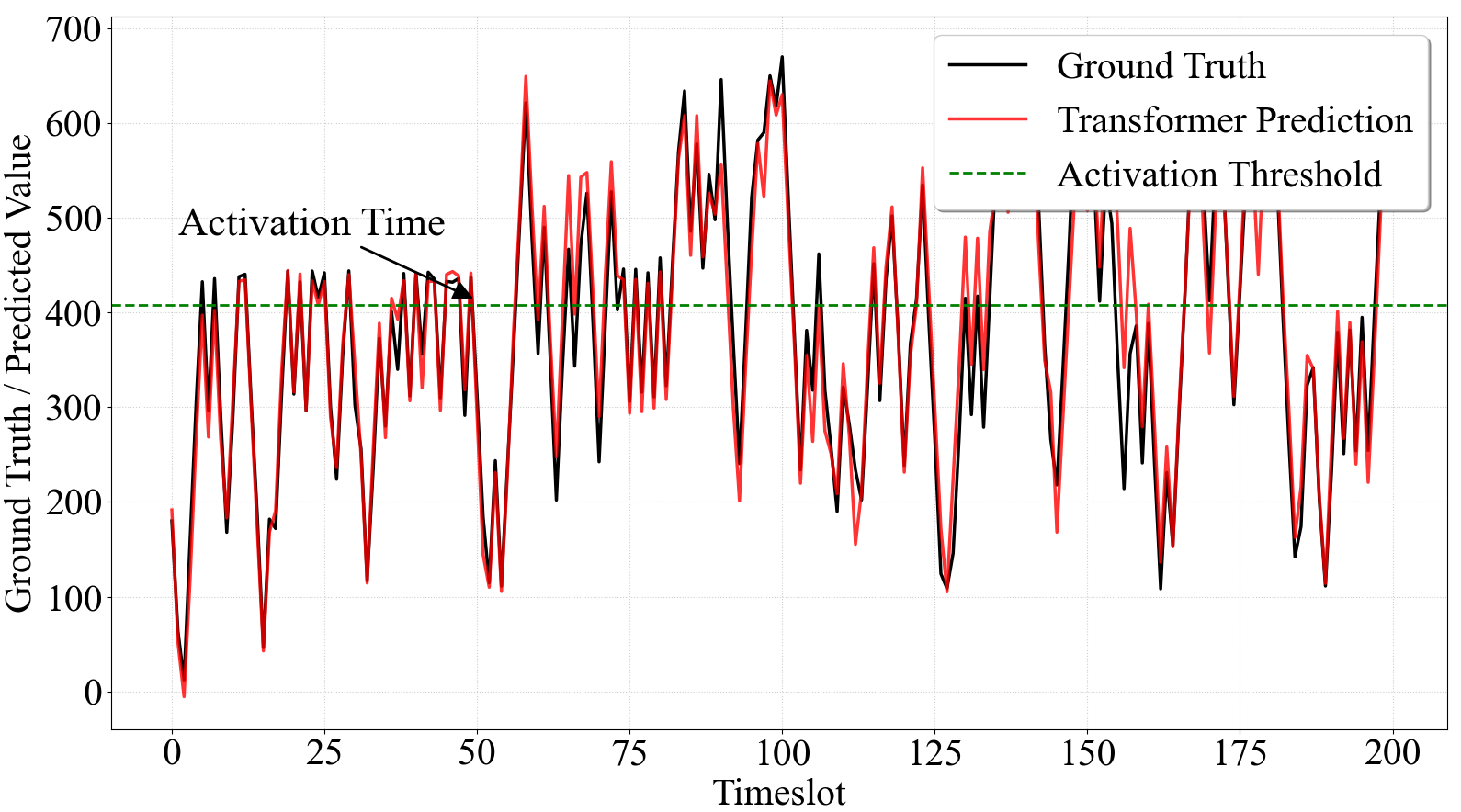}}
\subfloat[][]{\includegraphics[width=0.3\linewidth]{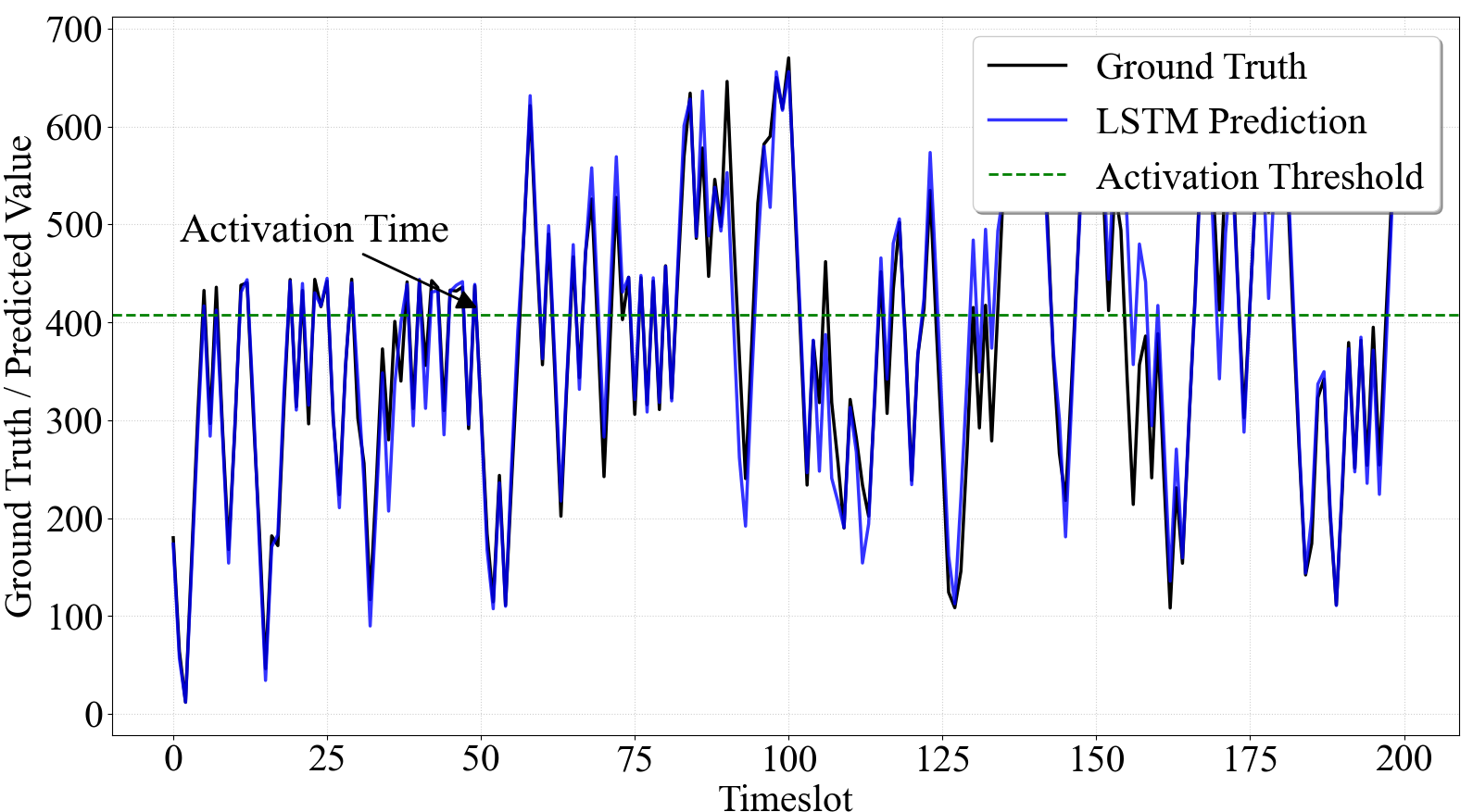}}
\subfloat[][]{\includegraphics[width=0.3\linewidth]{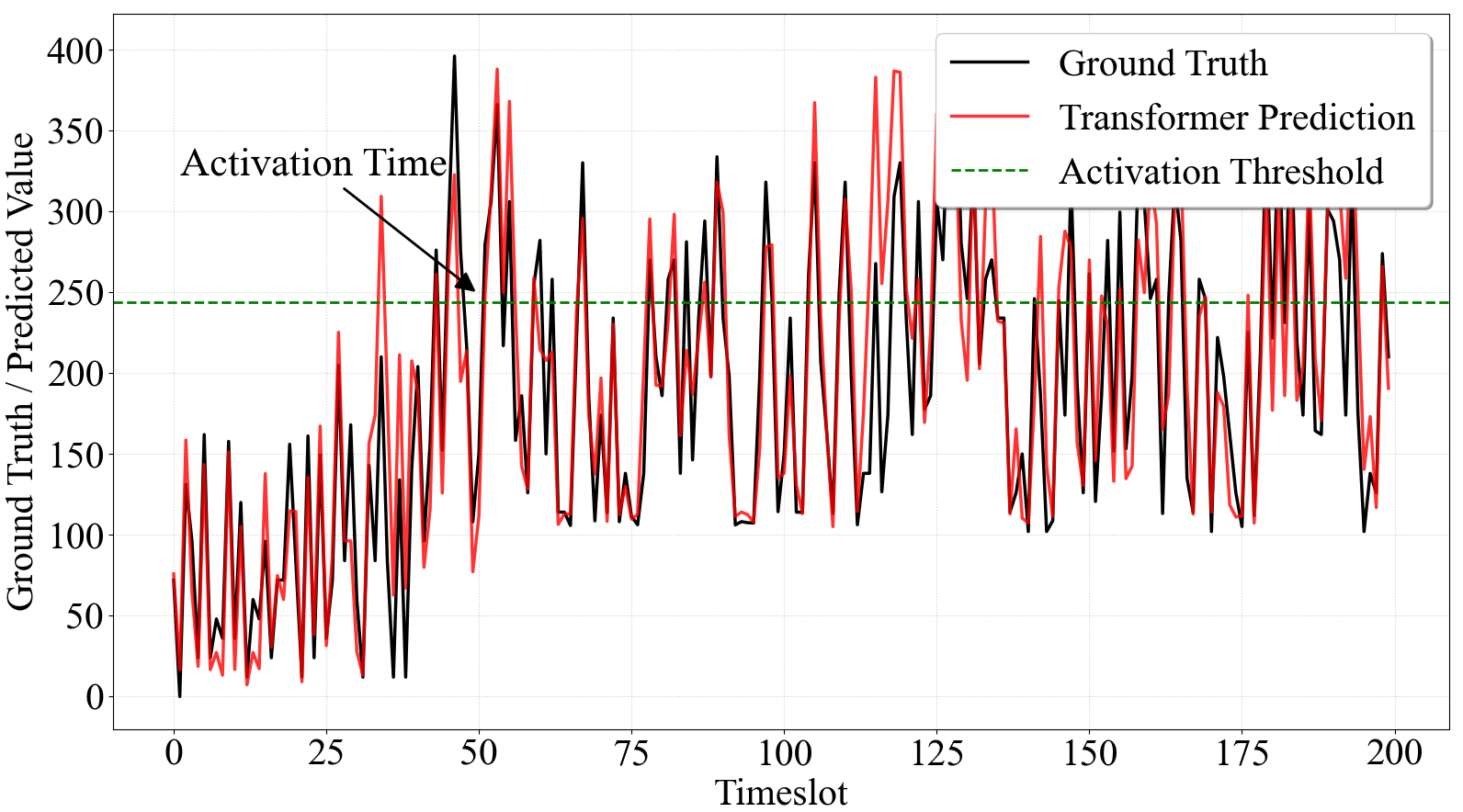}}\\
\subfloat[][]{\includegraphics[width=0.3\linewidth]{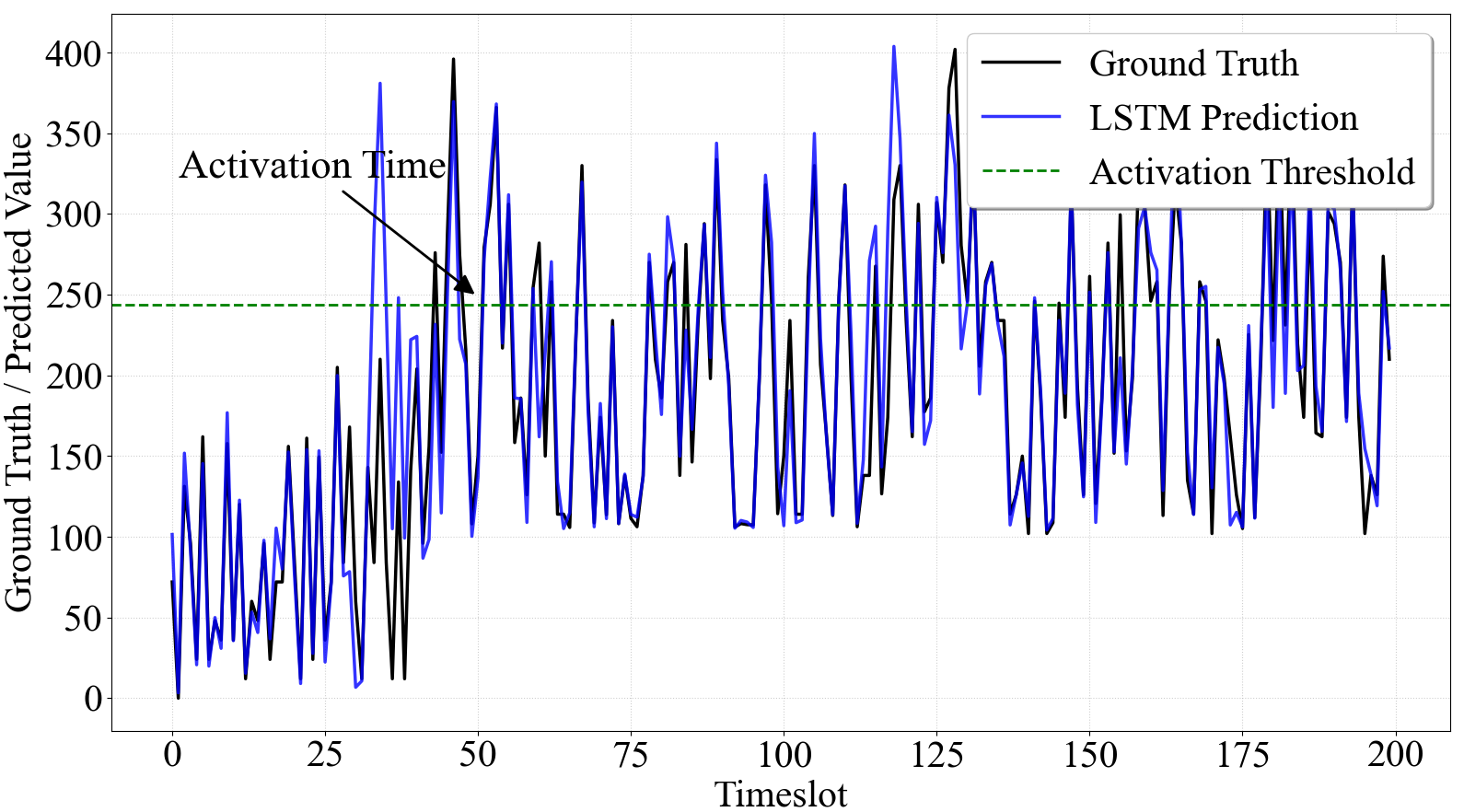}}
\subfloat[][]{\includegraphics[width=0.3\linewidth]{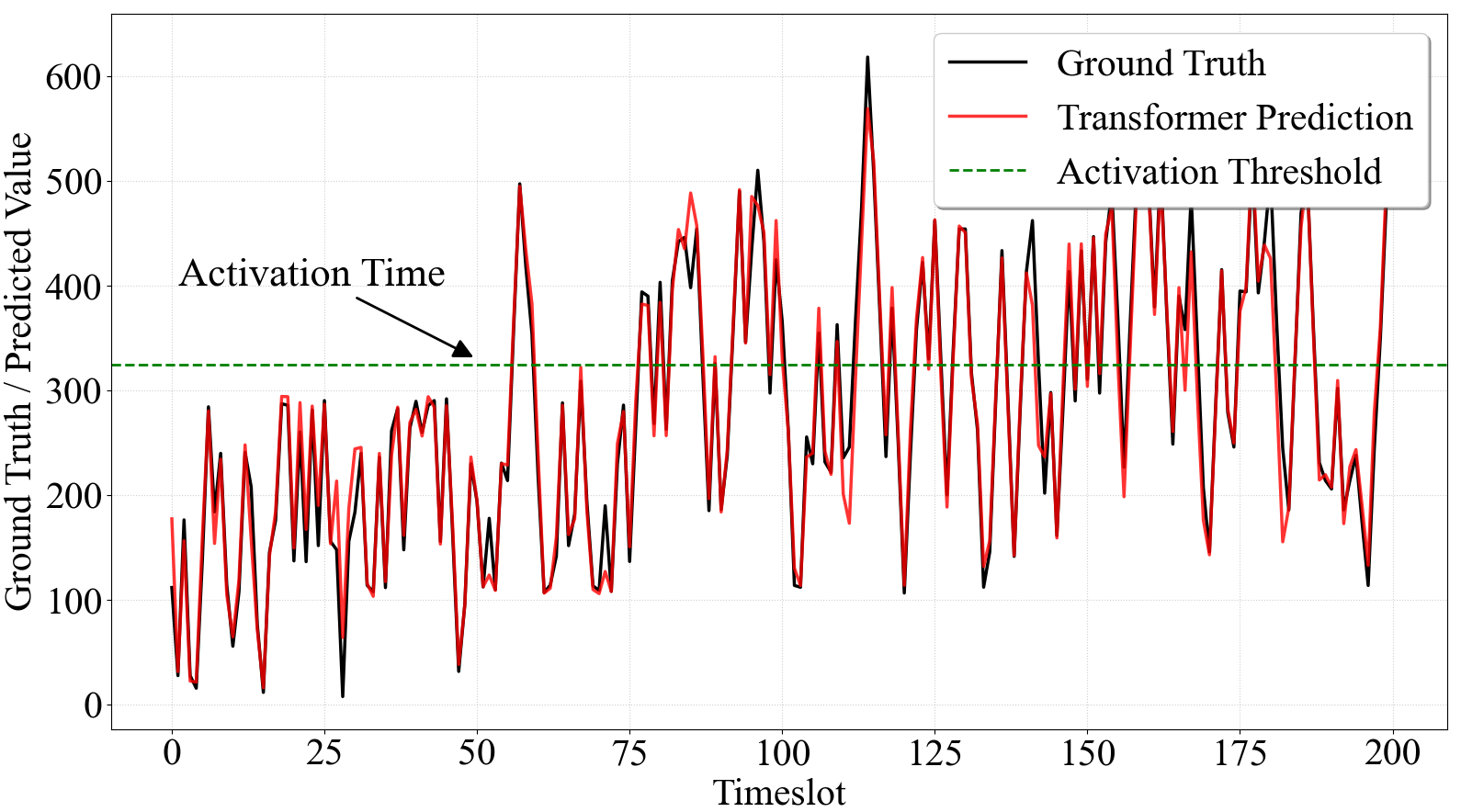}}
\subfloat[][]{\includegraphics[width=0.3\linewidth]{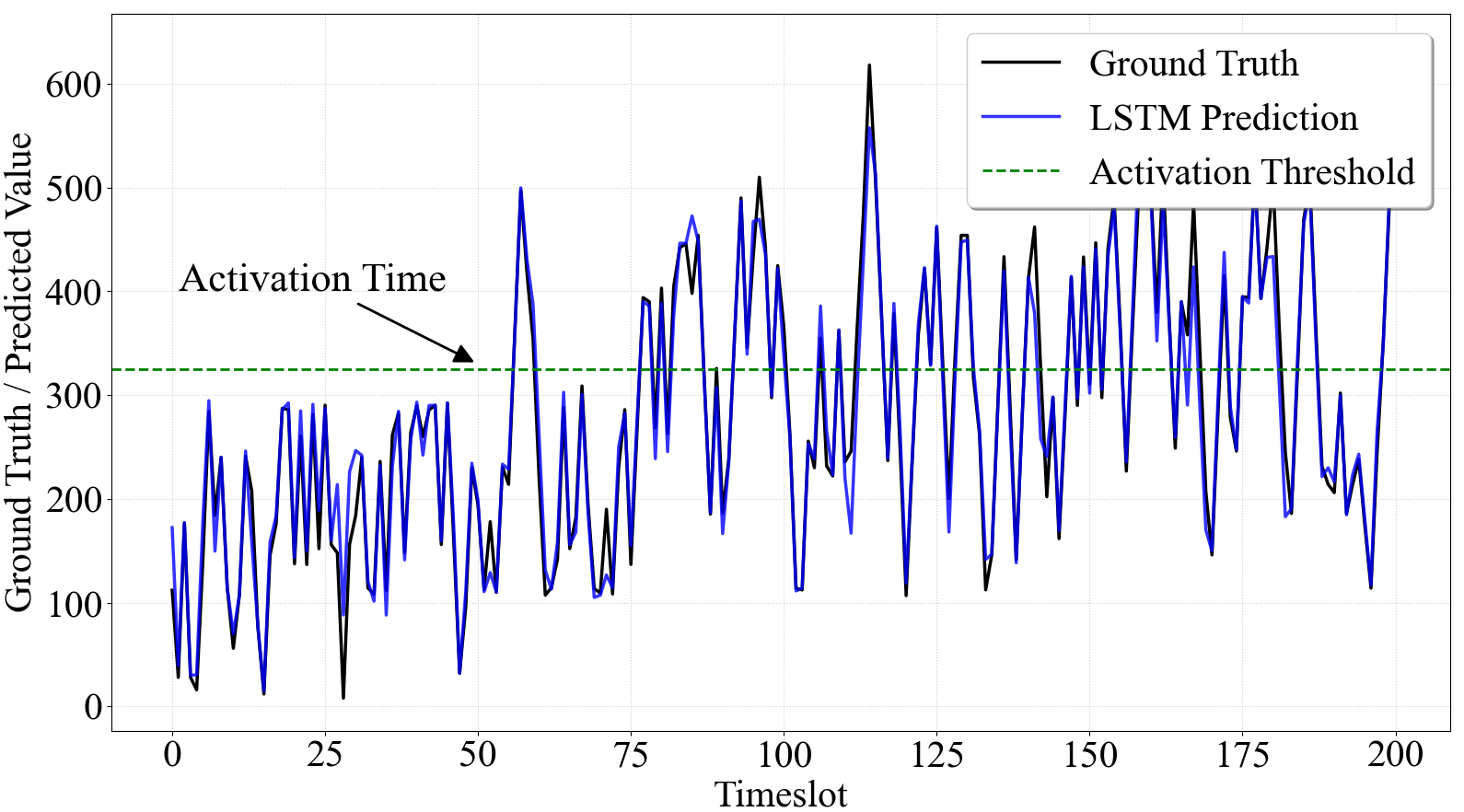}}
\caption{Prediction on type-$\mathsf{II}$ PoIs: three examples.}
\label{fig:Prediction}
\end{figure*}

Both models capture the overall temporal dynamics of traffic flow well; however, we adopt the Transformer for subsequent forecasting due to its superior accuracy and stronger capability of modeling long-range temporal dependencies. This advantage is particularly important in our setting, where the activation of a Type-$\mathsf{II}$ PoI is determined by whether its predicted traffic flow exceeds the threshold $o^{(*)}$. More accurate forecasts thus enable more reliable identification of PoI activation windows and reduce error propagation in subsequent UAV scheduling and sensing decisions. Accordingly, a Type-$\mathsf{II}$ PoI is considered active when its predicted traffic flow exceeds $o^{(*)}$. Since most PoIs exhibit sustained high traffic demand during the daytime, we set the common take-off time of the UAV swarm to the 60-th time slot. This choice aligns UAV deployment with the predominant activation period of Type-$\mathsf{II}$ PoIs, thereby improving task coverage under the energy-efficiency constraint.

\noindent
$\bullet$~\textit{Comparison with benchmark methods.}
The predicted PoI activation windows obtained from the Transformer are subsequently used to configure the offline pre-planning experiments. Parameters of scenario and our $\rm ES^2A$ are summarized in Tables \ref{tab:offline_params} and \ref{tab:es2a_params} (see \ref{parameter}), respectively.

\begin{figure*}[!htbp] 
\centering
\subfloat[][]{\includegraphics[width=0.25\linewidth]{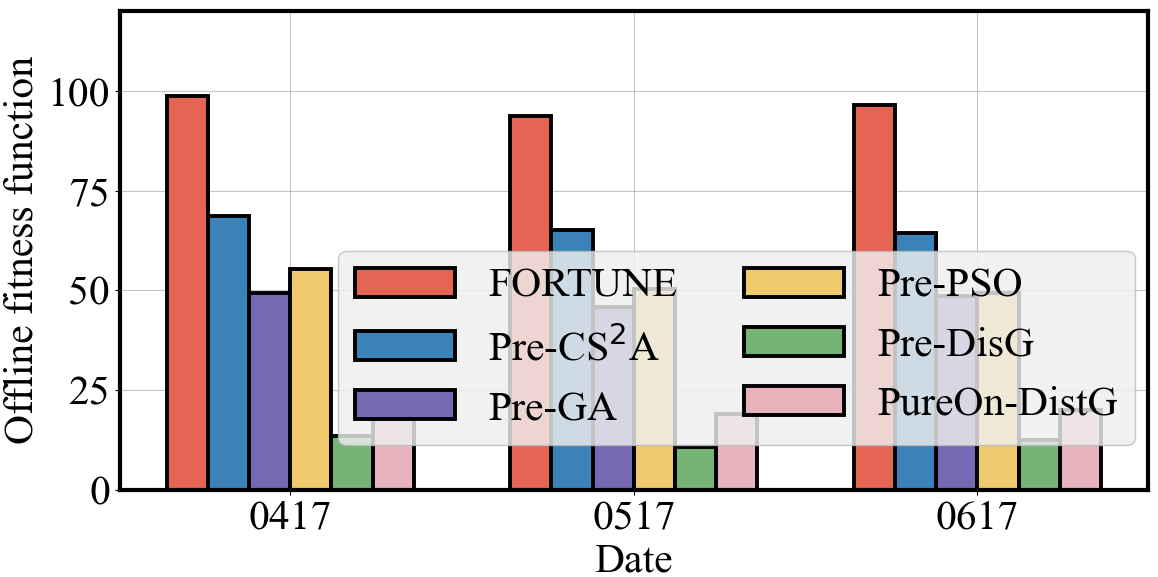}}%
\subfloat[][]{\includegraphics[width=0.255\linewidth]{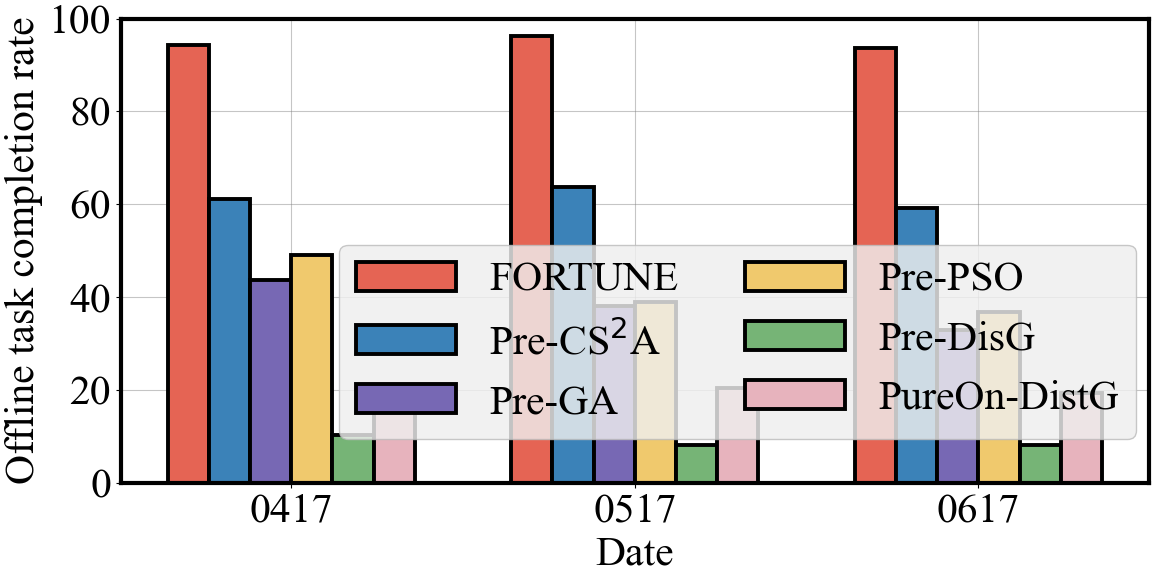}}
\subfloat[][]{\includegraphics[width=0.245\linewidth]{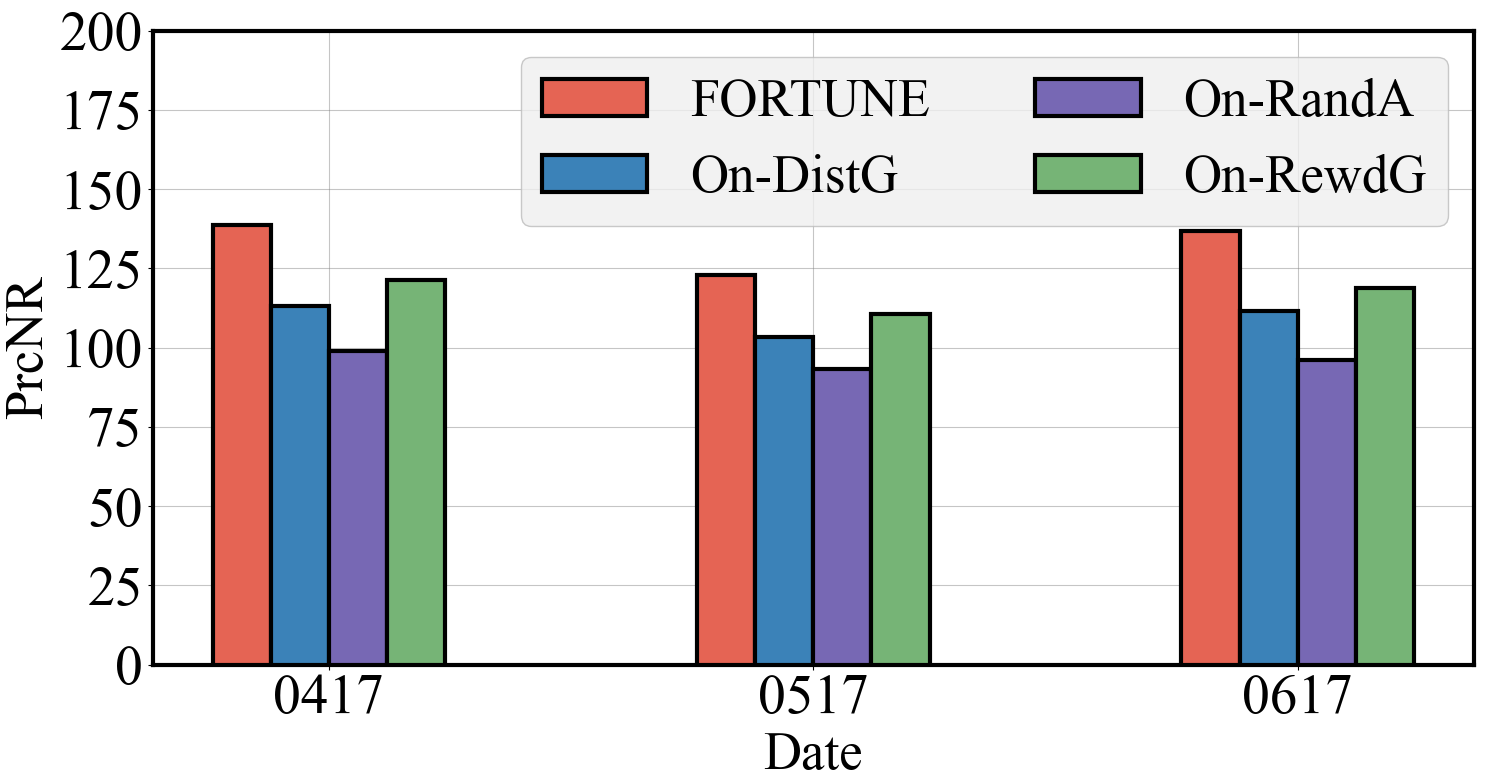}}%
\subfloat[][]{\includegraphics[width=0.245\linewidth]{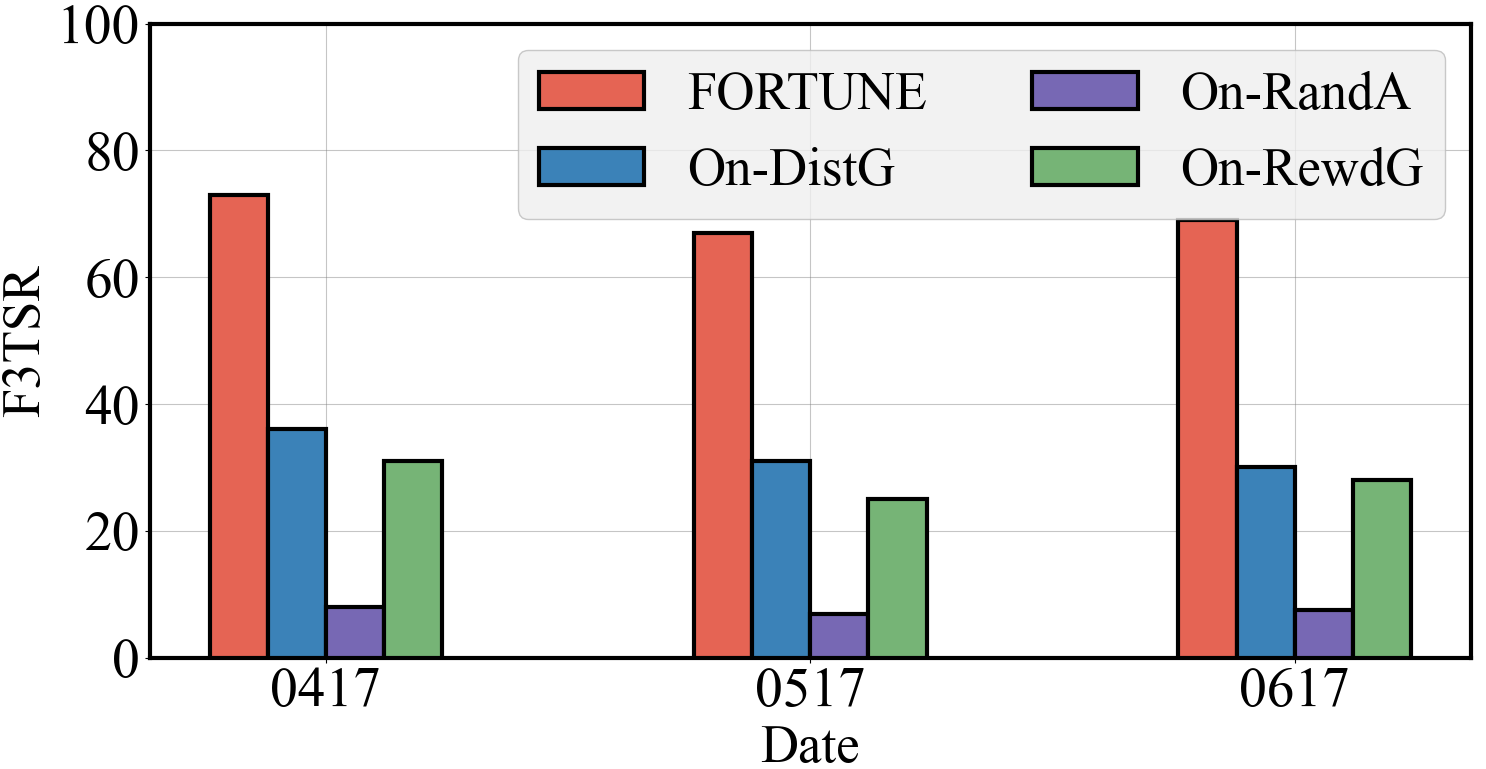}}
\caption{Performance on real-world dataset.}
\label{fig:realdata}
\end{figure*}

Figs. \ref{fig:realdata}(a)-(b) present the offline pre-planning results on UTD-19 dataset across different experimental days. Our $\rm ES^2A$ consistently achieves the OffF2 and OffTCR, with the latter approaching 95\% across all dates. These gains stem from three complementary enhancements. First, Tent-chaotic initialization promotes a more uniform distribution of priority sequences, substantially improving population diversity at the early search stage. Second, a hierarchical heuristic integrating joint 3D obstacle avoidance, adaptive altitude control, optimal sensing-altitude selection, and discrete velocity optimization effectively guides UAVs to complete data collection within the active windows of Type-$\mathsf{II}$ PoIs while satisfying tightly coupled 3D physical constraints, thereby reducing unnecessary hovering and energy expenditure. Finally, the combination of Lévy-flight-based long-range exploration and event-driven risk-aware perturbation significantly strengthens the optimizer's ability to escape local optima, yielding superior global search performance. Figs. \ref{fig:realdata}(c)-(d) further compare the online response strategies for Type-$\mathsf{III}$ PoIs based on the pre-planned offline paths. Our FORTUNE consistently outperforms all benchmark methods in both PrcNR (see (\ref{eq:reward_net})) and F3TSR. This advantage arises from its collaborative response mechanism, which overcomes the limitation of single-UAV dispatch by dynamically forming a cooperative UAV group $G_i$ to satisfy high-volume sensing demands. Moreover, the collaborative net-reward evaluation jointly accounts for mission rewards and response costs, including trajectory deviation and ecological penalties, enabling the selection of globally optimal response strategies. Compared with On-RewdG, On-DistG achieves a higher F3TSR by assigning the nearest available UAV, thereby shortening transit time and extending the effective sensing duration for task execution. Conversely, On-RewdG prioritizes high-reward PoIs, sacrificing partial task coverage in exchange for larger per-task returns, which compensates for unfinished tasks and consequently yields a higher PrcNR than On-DistG.

\FloatBarrier
\subsection{Experiments on Numerical Dataset}
\label{realworld}
To further assess the scalability of FORTUNE, we construct diverse scenario settings of increasing complexity. The duration of each time window is uniformly sampled from 5 to 10 timeslots. Unless otherwise specified, remaining parameters follow tables in \ref{parameter}. 

\noindent
\textit{Scenario\#1:} 20 UAVs operate over a ($3\mathrm{km}\times3\mathrm{km}$) area containing 10 Type-$\mathsf{I}$ PoIs and 10 Type-$\mathsf{II}$ PoIs.

\noindent
\textit{Scenario\#2:} 30 UAVs operate over a ($4\mathrm{km}\times4\mathrm{km}$) area containing 15 Type-$\mathsf{I}$ PoIs and 15 Type-$\mathsf{II}$ PoIs.

\noindent
\textit{Scenario\#3:} 40 UAVs operate over a ($5\mathrm{km}\times5\mathrm{km}$) area containing 20 Type-$\mathsf{I}$ PoIs and 20 Type-$\mathsf{II}$ PoIs.

\begin{figure}[!htbp] 
\centering
\subfloat[][]{\includegraphics[width=0.25\linewidth]{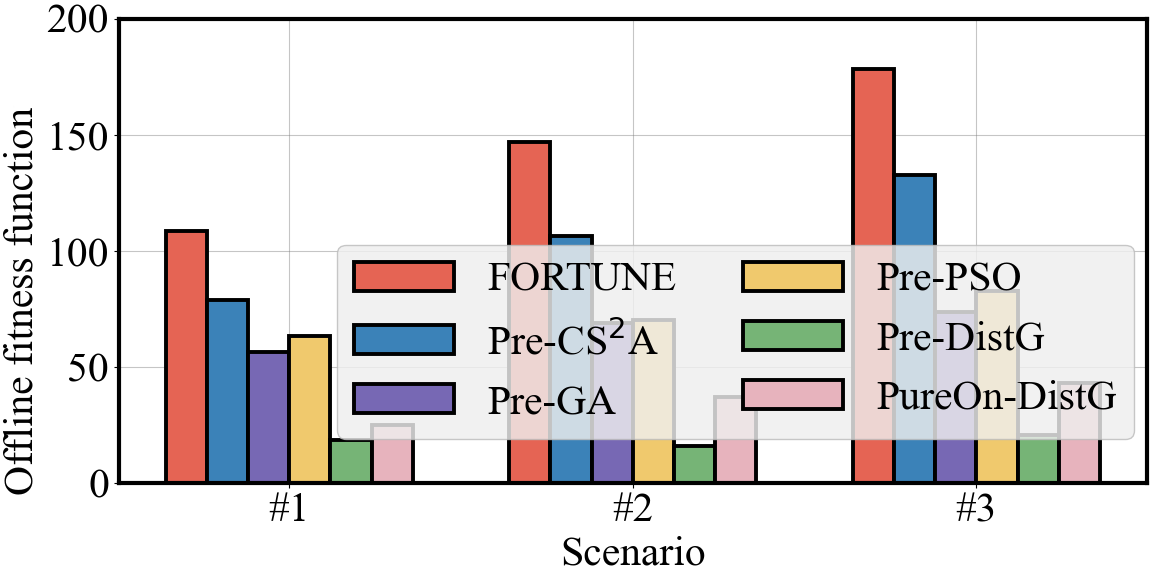}}%
\subfloat[][]{\includegraphics[width=0.25\linewidth]{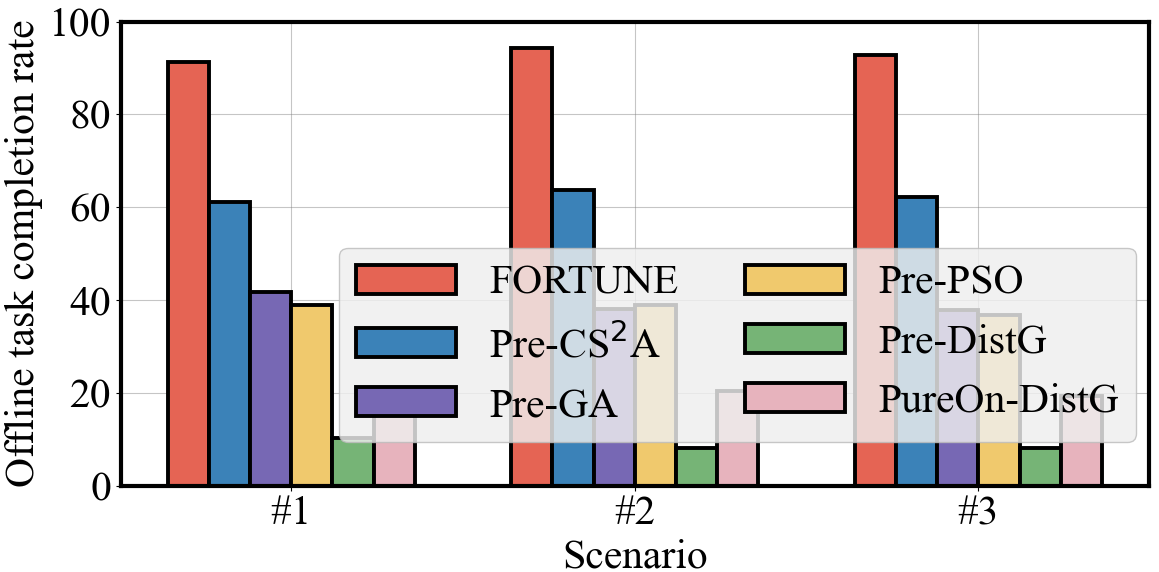}}
\subfloat[][]{\includegraphics[width=0.22\linewidth]{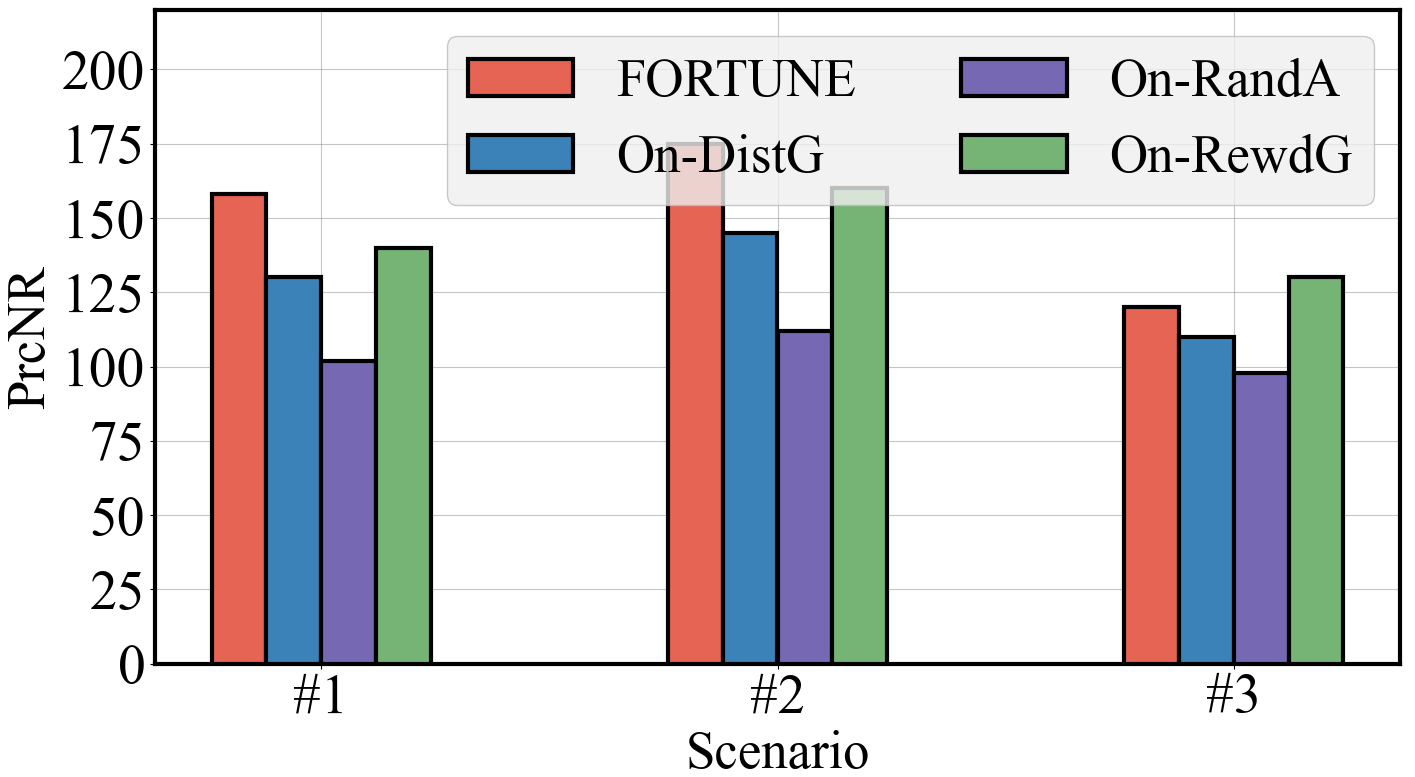}}%
\subfloat[][]{\includegraphics[width=0.24\linewidth]{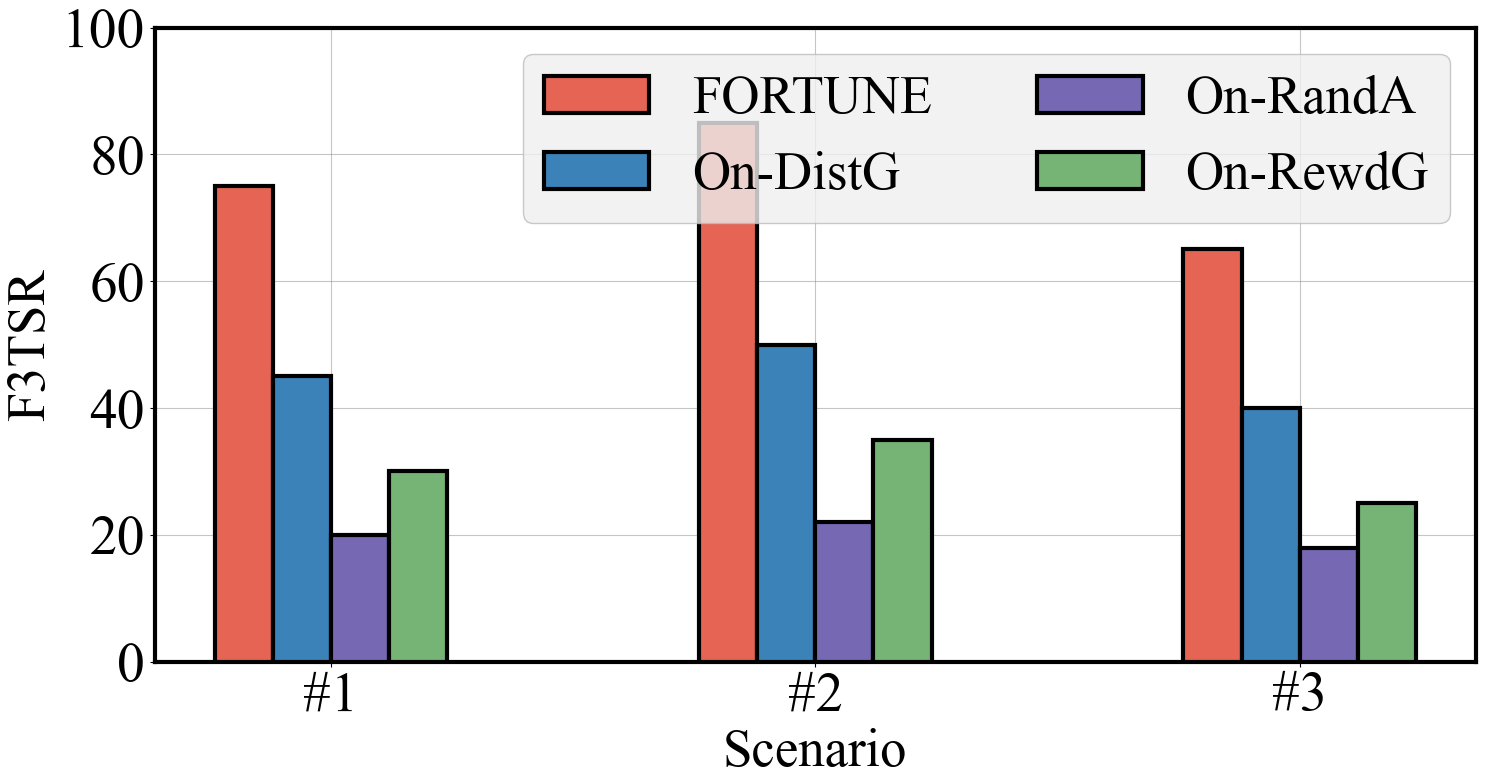}}
\caption{Performance on numerical data settings.}
\label{fig:numericaldata}
\end{figure}

Figs. \ref{fig:numericaldata}(a)-(b) compare OffF2 and OffTCR under varying problem scales. Consistent with the real-world experiments, our FORTUNE consistently achieves the best performance across all settings. This superiority stems from the synergistic integration of multiple algorithmic enhancements, which simultaneously improve population initialization quality and global search capability. More importantly, the results validate the robustness and scalability of FORTUNE, confirming its effectiveness for large-scale offline cooperative trajectory pre-planning in dynamic PoI-driven UAV collaboration. Moreover, Figs. \ref{fig:numericaldata}(c)-(d) exhibit performance trends consistent with those in Fig. \ref{fig:realdata} across all three scenarios, further validating the robustness of FORTUNE in accommodating stochastic Type-$\mathsf{III}$ PoIs through efficient online collaborative response. Furthermore, we visually illustrate our FORTUNE with examples across different timeslots, shown in Fig.~\ref{fig:3D}.

\begin{figure}[H]
\centering
\subfloat[][]{\includegraphics[width=0.16\linewidth]{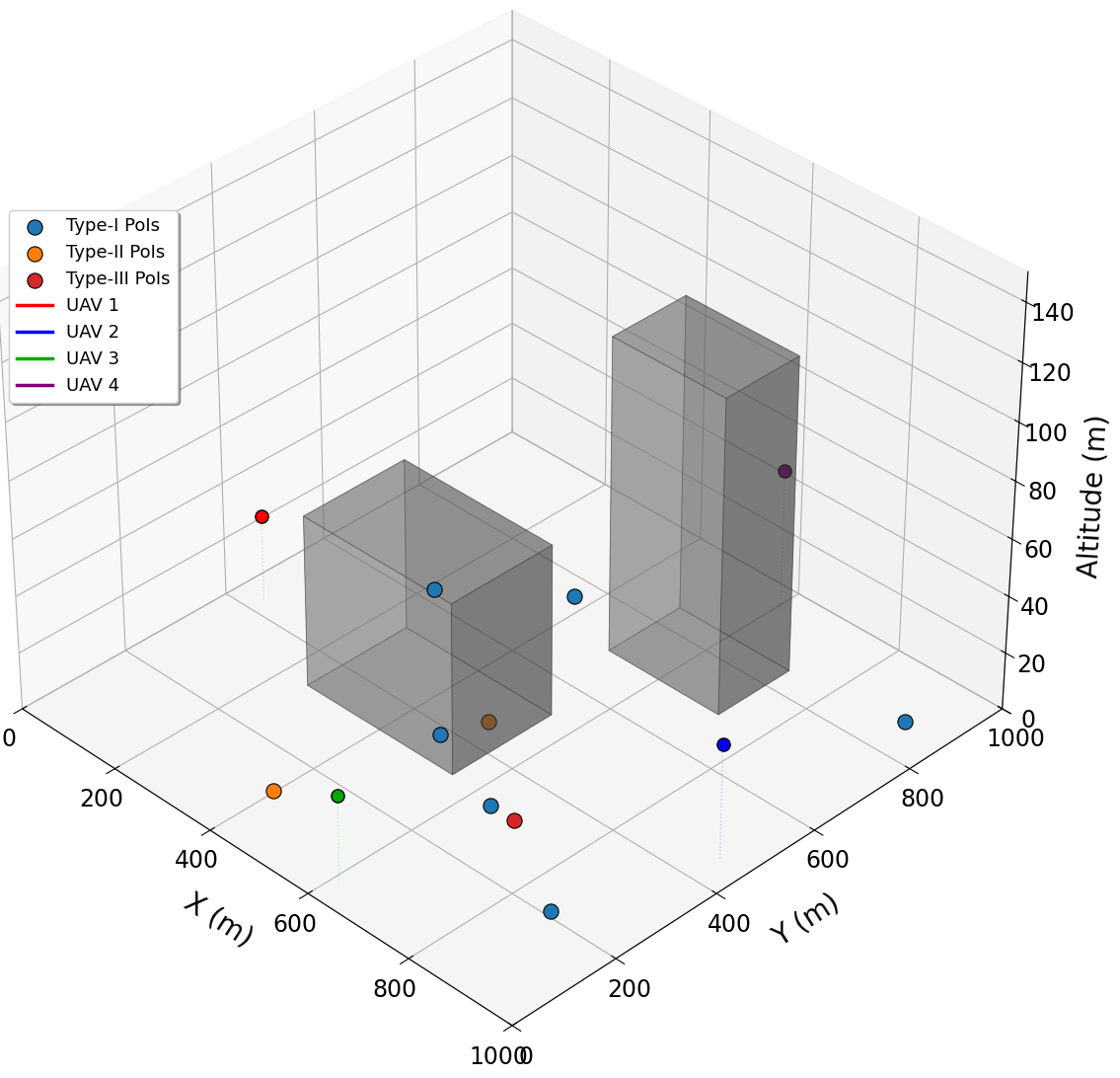}}
\subfloat[][]{\includegraphics[width=0.16\linewidth]{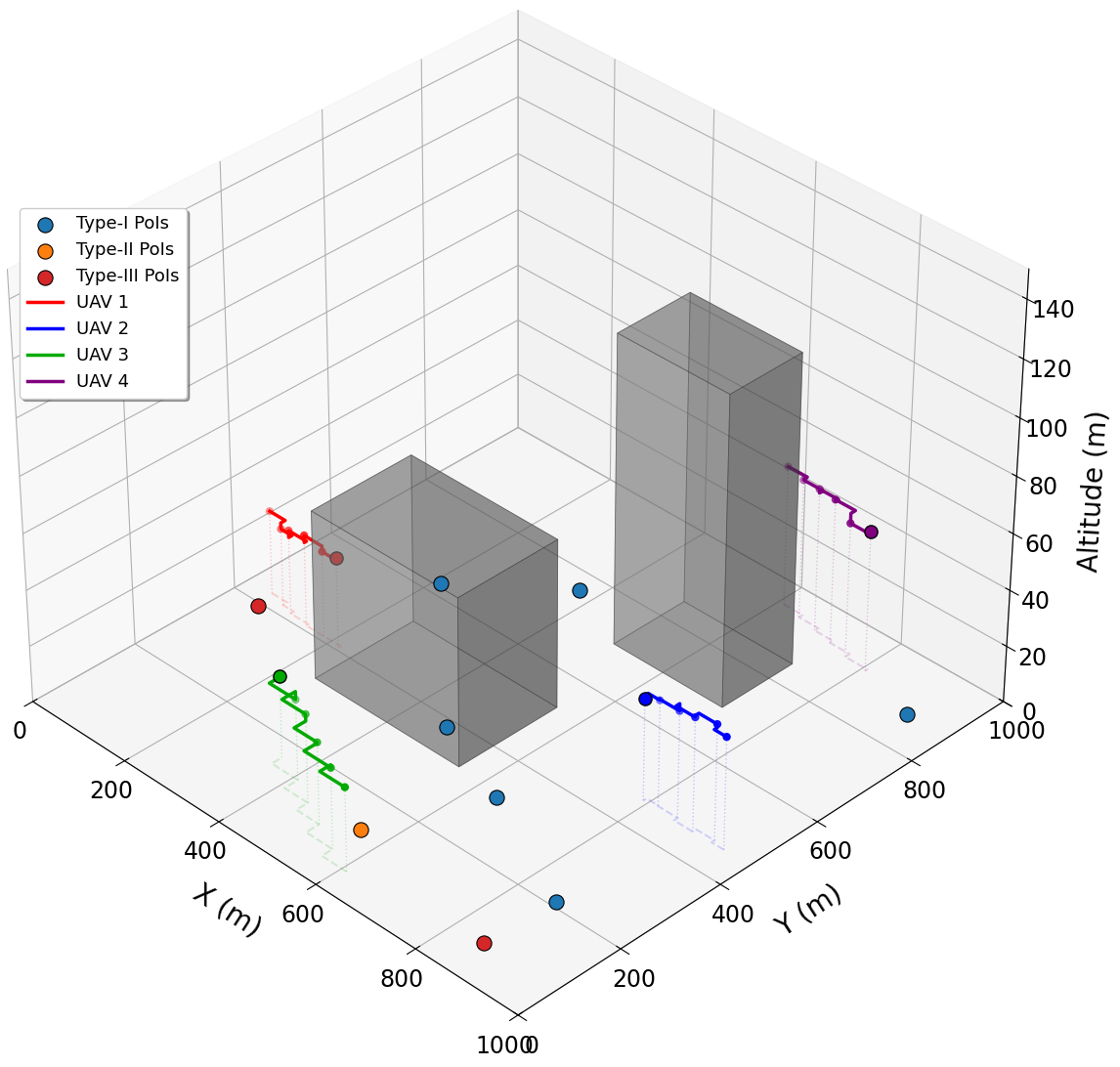}}
\subfloat[][]{\includegraphics[width=0.16\linewidth]{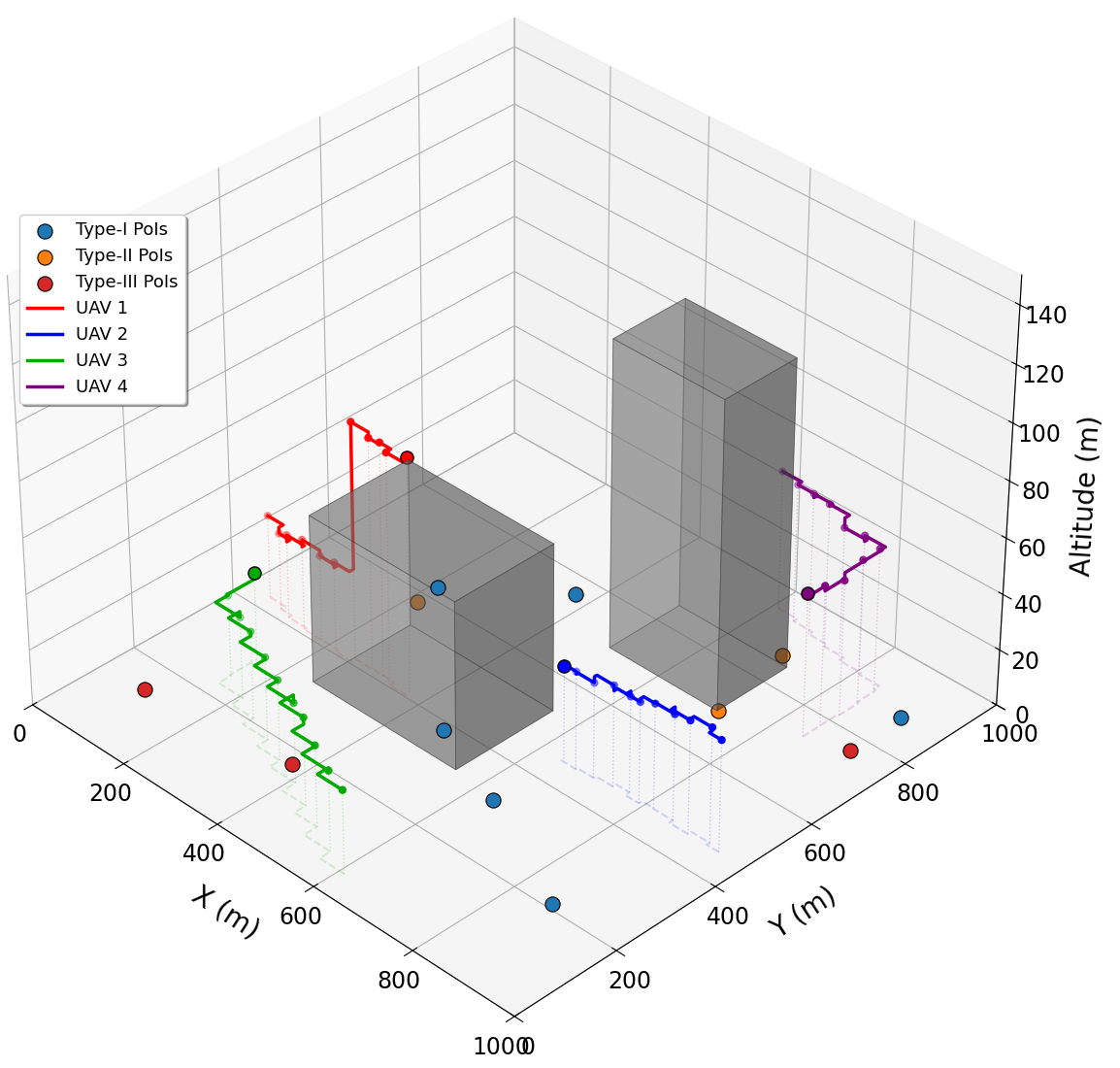}}
\subfloat[][]{\includegraphics[width=0.16\linewidth]{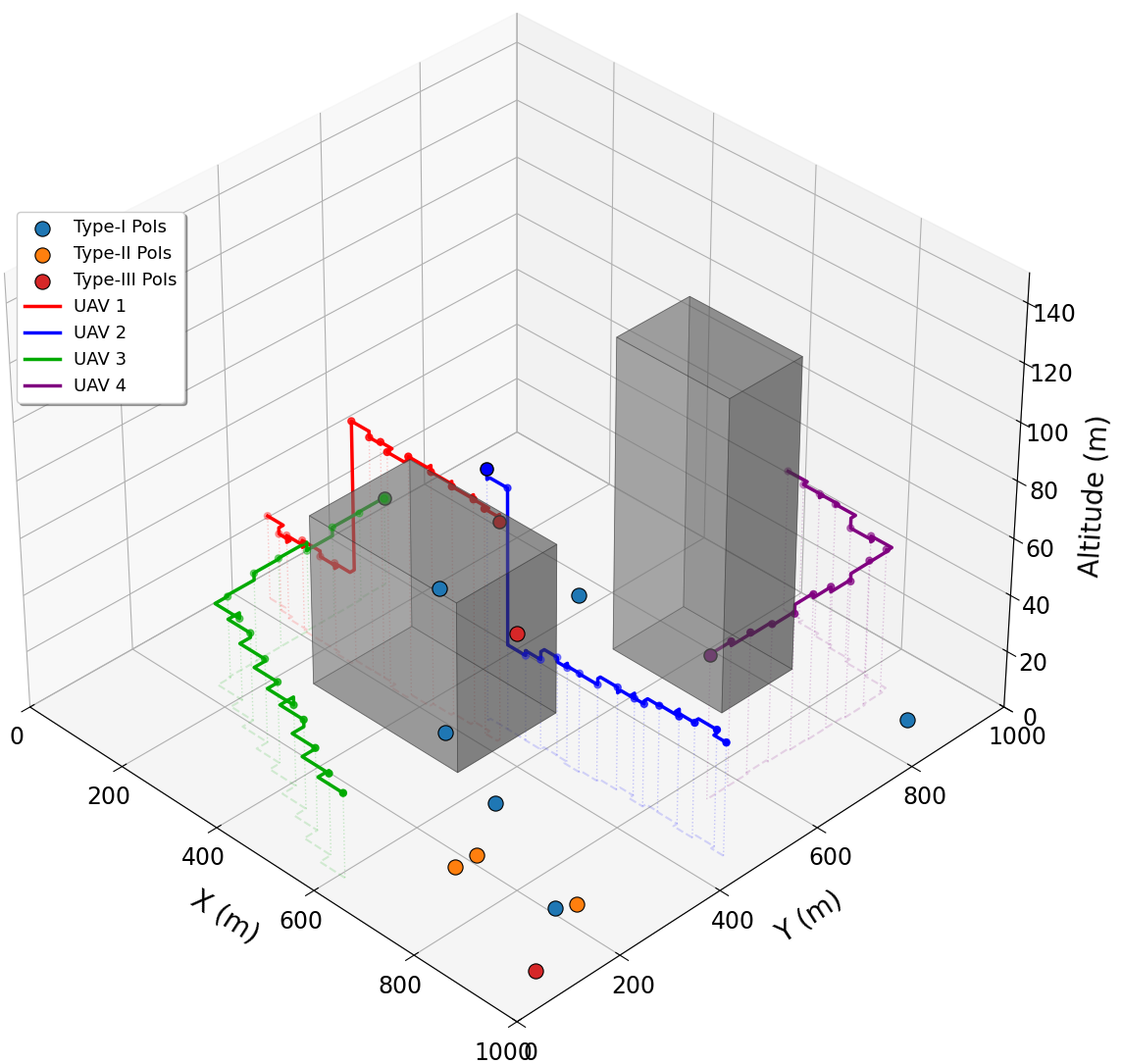}}
\subfloat[][]{\includegraphics[width=0.16\linewidth]{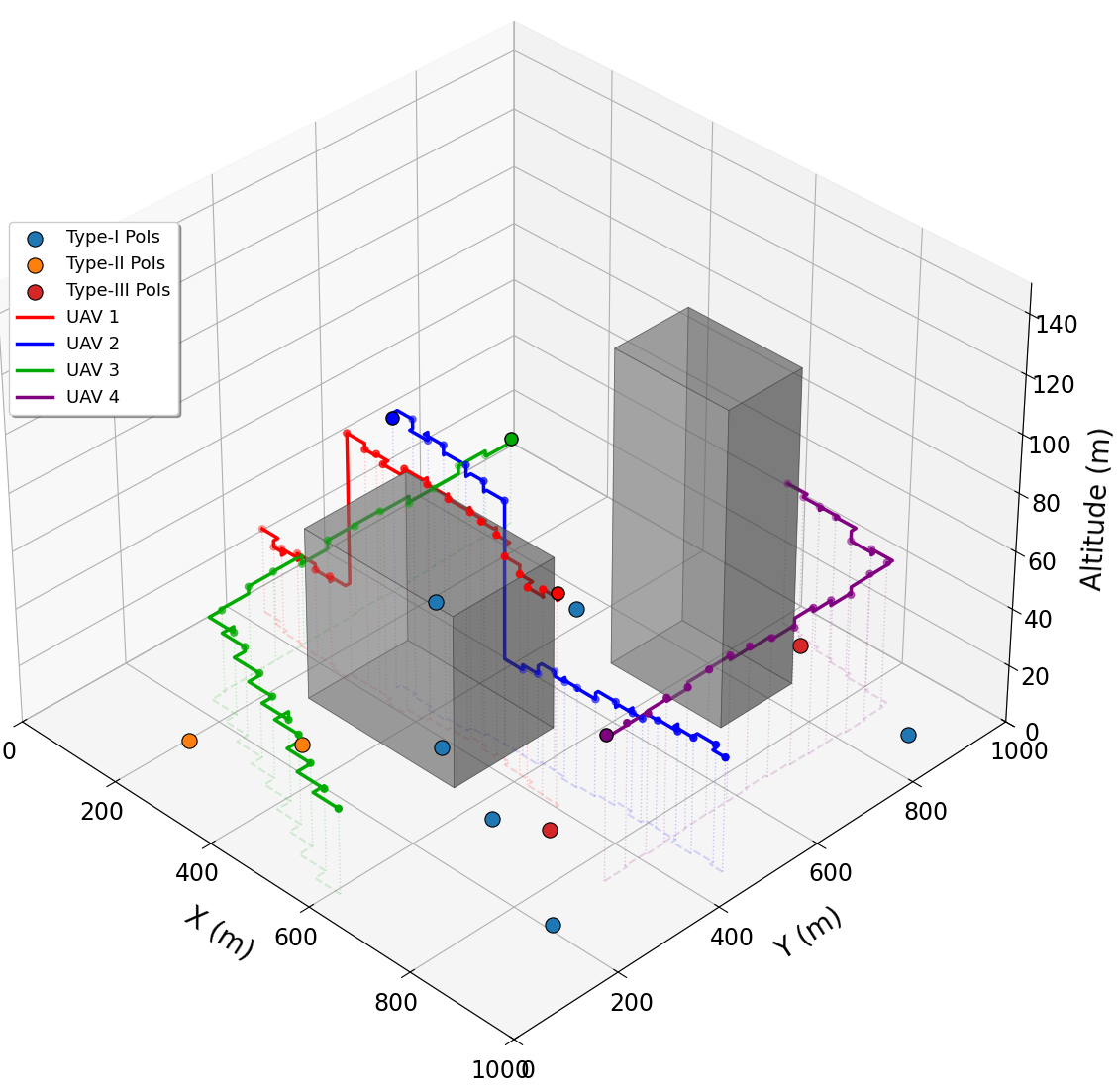}}
\subfloat[][]{\includegraphics[width=0.16\linewidth]{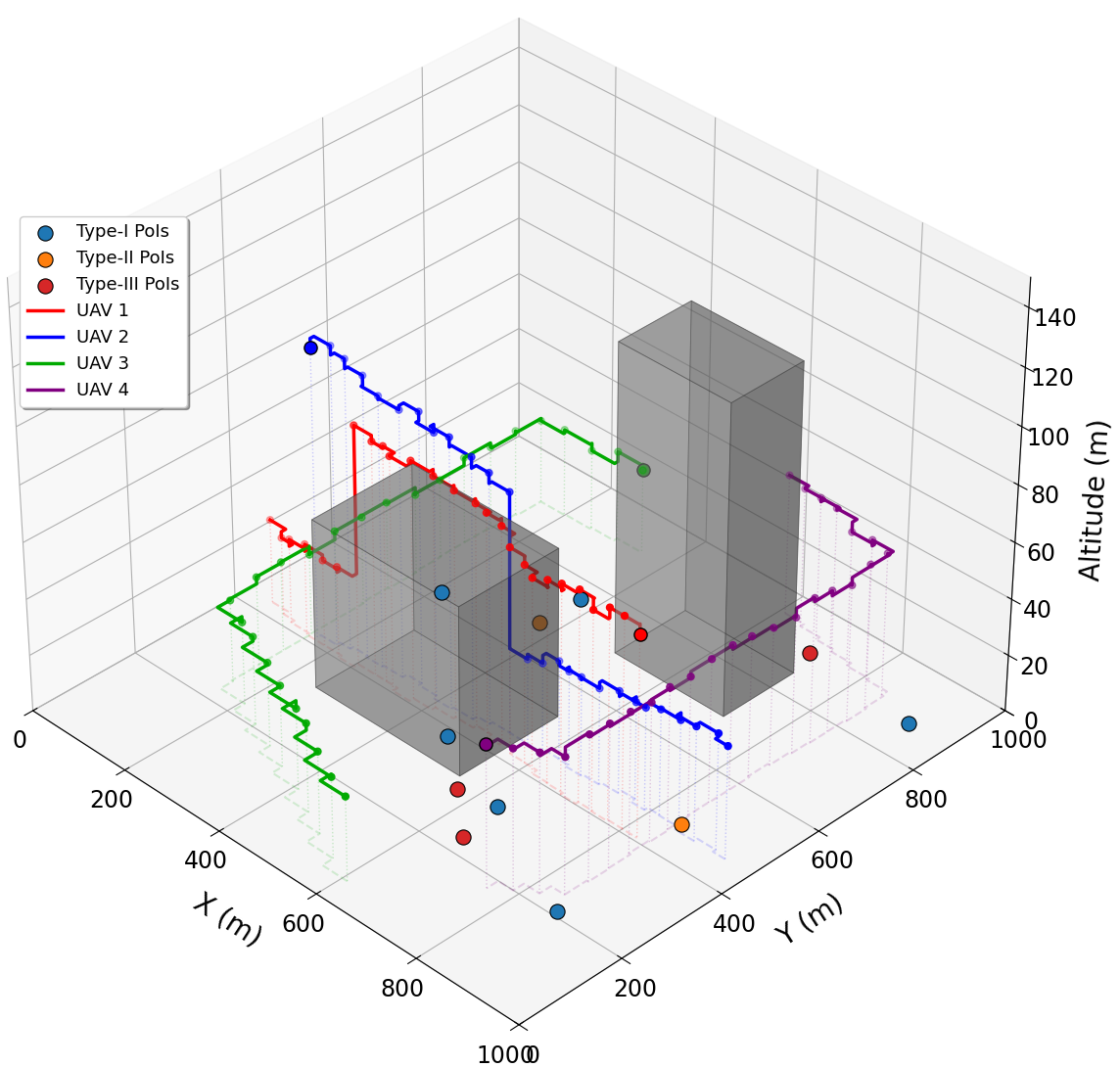}}
\caption{Examples of 3D UAV paths: randomly selected four UAVs as examples. }
\label{fig:3D}
\end{figure}

\endgroup
\section{Conclusion and Future Work}
This paper presented FORTUNE, a hierarchical offline-online framework for 3D low-altitude collaborative sensing under heterogeneous spatio-temporal uncertainty. To the best of our knowledge, this is the first work to unify persistent, predictable, and emergent PoIs within a single planning framework while explicitly incorporating altitude-dependent societal and environmental impacts into UAV decision-making. Extensive experimental results demonstrated the effectiveness, adaptability, and scalability.
Future research may advance embodied aerial intelligence through self-organizing collaborative autonomy, enabling UAV swarms to dynamically coordinate sensing responsibilities in response to evolving spatio-temporal demands, while achieving LA4Good.

\FloatBarrier
\bibliographystyle{splncs04}
\bibliography{referencev1}

\clearpage
  
\begin{appendix}
\renewcommand{\thealgocf}{\Alph{section}\arabic{algocf}}
\renewcommand{\theHequation}{appendix.\Alph{section}.\arabic{equation}}
\renewcommand{\theHfigure}{appendix.\Alph{section}.\arabic{figure}}
\renewcommand{\theHtable}{appendix.\Alph{section}.\arabic{table}}

\section{Literature Review}
\label{literature}

\noindent
This section reviews related literature in detail, further highlight our key differences and novelties.

\noindent
$\bullet$~\textit{View 1. Static/known PoIs vs. uncertain PoIs over spatio-temporal evolution.} Most existing studies regarding UAV-assisted data collection focus on known or fixed PoIs, which are typically modeled with predetermined parameters such as quantity, location, and active time windows. For instance, \textit{He} et al. \cite{2026TCCNRLstatic} implemented a two-timescale trajectory planning scheme within a leader-follower control framework for known hotspot areas. \textit{Guo} et al. \cite{2026TMCfixedregions} investigated hierarchical incentive mechanisms for UAV-assisted crowdsensing while optimizing data freshness within a fixed set of subregions. Similarly, \textit{Yuan} et al. \cite{2026TCOMRLknown} integrated transfer learning with reinforcement learning (RL) to solve 3D trajectory planning problems for known ground user hotspots. \textit{Ye} et al. \cite{2025TMCAoI} explored air-ground cooperative crowdsensing by optimizing the AoI metric, where vehicles dispatch UAVs from multiple stops to collect data from specific PoIs, with multi-agent curriculum learning (MACL). Furthermore, \textit{Wang} et al. \cite{2025TONAoIMARL} focused on urban data collection to maintain AoI within specified thresholds across diverse but known PoIs. \textit{Hou} et al. \cite{2025TVTtwoStagestatic} proposed a hierarchical optimization framework for multi-UAV collaboration to jointly address trajectory planning, energy management, and load balancing, where each sensing task is tied to a specific target in a designated area. \textit{Li} et al. \cite{2025TASEDPoI} studied air-ground collaborative sensing where UAVs are guided along informative trajectories to reach target positions rapidly. Although these works contribute significantly to UAV path planning and data freshness (e.g., via AoI), they largely treat PoIs as static entities with a priori known locations. This assumption often falls short of reality, as real-world networks exhibit complex spatio-temporal dynamics. While \textit{Li} et al. \cite{UAVbackground5} explored multi-task-oriented UAV crowdsensing with diverse AoI requirements and introduced a ``valid task handling index'', their work approaches the problem from a fundamentally different perspective with us. Unlike these studies, we concern  dynamic and uncertain PoIs, where the varying activation windows of them lead to sensing tasks with disparate timing and locations.

\noindent
$\bullet$~\textit{View 2. Way to optimize UAV paths.}
Path planning involves complex spatio-temporal coupling where current decisions propagate through the temporal horizon, leading to exponential complexity. Given the computational intractability of conventional integer programming in high-dimensional multi-agent systems, research has shifted toward \textit{RL-based approaches}, \textit{heuristic-driven methods}, and \textit{hybrid ideas that integrate different methodologies}. The rapid advancement of artificial intelligence (AI) has enabled
an effective integration of deep neural networks with RL, which are widely adopted path planning problems\cite{2025TMCAoI, 2025TONAoIMARL,2025TVTtwoStagestatic,2026TCCNRLstatic,2026TCOMRLknown,2024TVTDeng,2023TMCRL,2025TITSstaticPoI}. For example, \textit{Yu} et al. \cite{2025TITSstaticPoI} studied multi-UAV 3D trajectory planning strategy using multi-agent deep reinforcement learning (MADRL), by employing multi-agent proximal policy optimization (MAPPO). \textit{Deng} et al. \cite{2024TVTDeng} proposed a decentralized MADRL framework with multi-level state representations,
facilitating personalized training and better exploitation of environmental information. \textit{Gao} et al. \cite{2023TMCRL} applied multi-agent deep deterministic policy gradient (MADDPG) to jointly optimize task allocation and trajectory planning. Despite the efficacy in navigating dynamic environments, its practical deployment remains constrained by suboptimal sample efficiency, stochastic training instability, and fragile scalability. Existing literature predominantly focuses on restricted coordination scenarios, typically involving fewer than 10 agents (usually around 4), whereas scaling to denser swarms triggers an exponential explosion of the joint state-action space. This frequently precipitates gradient divergence and computational collapse during policy optimization. To circumvent the above bottlenecks, a large body of research has pivoted toward heuristic-based frameworks. Although rooted in classical optimization principles, they exhibit superior architectural scalability, computational parsimony, and structural robustness, rendering them more viable for high-density, large-scale systems. \textit{Wang} et al. \cite{2024IOT} presented a multi-vehicle path planning method based on multi-group cooperative particle swarm optimization (PSO), handling diverse intersections and traffic conditions. \textit{He} et al. \cite{2025ASCheuristic} investigated adaptive heuristic with collaborative search to for multi-UAV inspection path planning. \textit{Liu} et al. \cite{2024TII} considered UAV energy, mapping efficiency, and target structure coverage while respecting hardware constraints, with two-stage heuristic designed for trajectory planning. While heuristic methods enhance computational tractability in large-scale problems, their efficacy often hinges on idealized assumptions. Specifically, most existing models enforce a singleton visit constraint and assume stationary PoI distributions with deterministic activation windows. To tackle complex problems such as highly coupled and non-convex mixed-integer optimizations, research focus begins turning to hybrid solution design with multiple stages. For example, \textit{Yuan} et al.\cite{2025TCOMthreeStage} divide the problem into three stages, namely, clustering, trajectory planning, and resource allocation with algorithms designed respectively. \textit{Pan} et al.\cite{2026TCCNheuristicRL} combined genetic algorithm (GA) and MADRL, to achieve the distributed multi-UAV 3D trajectory planning collaborative optimization. \textit{Guo} et al.\cite{2026TCCNmultistage} decomposed the main problem into tow sub-ones, while designing zero-forcing algorithm and successive convex approximation, respectively, to support UAV-assisted sensing and communication.
Although these hybrid methodologies empowered solution design, dynamic and uncertain tasks (i.e., uncertain PoIs) are somehow overlooked. From a novel perspective, we propose a multi-module framework integrating prediction-enhanced PoI active time window estimation, UAV path pre-planning, and greedy online adjustment, enabling an effective solution to the complex problem.

\noindent
$\bullet$~\textit{View 3. Obstacle avoidance vs. socially compliant navigation.}
Existing UAV trajectory planning methods primarily formulate surrounding environments as physical obstacles, focusing on collision avoidance, energy efficiency, and mission completion while paying limited attention to human-centered social considerations \cite{2025TITSstaticPoI,2025TVTheuristic,2026TCCNheuristicRL}. However, as UAVs are increasingly deployed in low-altitude urban environments, socially compliant flight behavior has become equally important. Unlike conventional obstacle avoidance, UAVs inevitably share airspace with nearby pedestrians and residents. Flying at low altitudes or in close proximity to people may violate personal space, reduce perceived safety, and induce psychological discomfort, while rotor noise further degrades human comfort and environmental acceptability. These human-aware considerations have been extensively investigated in socially aware robot navigation and human-robot interaction, where factors such as proxemics, perceived safety, and social norms are explicitly incorporated into motion planning\cite{2025SANG,2025ACMsurvey}. Nevertheless, such socially compliant principles remain largely overlooked in existing UAV path planning frameworks. To bridge this gap, we explicitly incorporate a human-aware social cost into trajectory optimization, enabling UAVs to adapt their flight altitude and path according to socially compliant factors, thereby jointly optimizing sensing performance and social acceptability in accordance with LA4Good paradigm.

\section{Pseudocodes related to $ES^2A$}
\label{pseudocodes}
\noindent
In the following, we show detailed algorithmic tables corresponding to our $ES^2A$ design.

\begin{algorithm2e}[H]
    \caption{Altitude optimization for path pre-planning}
    \label{alg:high}
    \scriptsize
    \SetInd{0.25em}{0.7em}
    \SetAlgoNlRelativeSize{-1}
    \setlength{\interspacetitleruled}{2pt}
    \setlength{\algotitleheightrule}{0.4pt}

    \KwIn{UAV $u_m$; start point $p_i=(x_i,y_i,h_i)$; target point
    $p_{i'}=(x_{i'},y_{i'},h_{i'})$; obstacle set
    $\mathbbm{b}_{i,i'}^{(\mathsf{path})}$; maximum obstacle height
    $h_{i,i'}^{(\mathsf{obs})}$; altitude bounds
    $[h^{(\mathsf{min})},h^{(\mathsf{max})}]$; minimum collection
    altitude $h_{i'}^{(\mathsf{uav},\mathsf{min})}$; safety margin
    $\delta^{(\mathsf{safe})}$; thresholds
    $\varkappa^{(\mathsf{eco})}$ and $\varkappa^{(\mathsf{mid})}$;
    weighting factors $\gamma_1,\gamma_2$; search step
    $l^{(\mathsf{search})}$}

    \KwOut{Selected cruising altitude
    $h_m^{(\mathsf{uav},t)*}$; selected collection altitude
    $h^{(\mathsf{collect})*}$; selected horizontal route
    $\mathcal{R}_{i,i'}^{*}$ and its distance $d_{i,i'}$;
    vertical-motion time $T_{m,i,i'}^{(\mathsf{vtl})}$ and energy
    $E_{m,i,i'}^{(\mathsf{vtl})}$}

    \tcp{Joint obstacle-avoidance and cruising-altitude decision}

    $h^{(\mathsf{ct})}
    \leftarrow
    \max\{h_i,h_{i'}\}$\;

    Obtain $h^{(\mathsf{eco})}$ from
    $f^{(\mathsf{eco})}
    \!\left(h^{(\mathsf{eco})}\right)
    =
    \varkappa^{(\mathsf{eco})}$\;

    Obtain $h^{(\mathsf{mid})}$ from
    $f^{(\mathsf{eco})}
    \!\left(h^{(\mathsf{mid})}\right)
    =
    \varkappa^{(\mathsf{mid})}$\;

    $\mathcal{H}
    \leftarrow
    \left\{
    h^{(\mathsf{ct})},
    h^{(\mathsf{eco})},
    h^{(\mathsf{mid})}
    \right\}$\;

    \If{$\mathbbm{b}_{i,i'}^{(\mathsf{path})}\neq\emptyset$}{
        $h^{(\mathsf{limit})}
        \leftarrow
        h_{i,i'}^{(\mathsf{obs})}
        +
        \delta^{(\mathsf{safe})}$\;

        $\mathcal{H}
        \leftarrow
        \mathcal{H}
        \cup
        \left\{
        h^{(\mathsf{limit})}
        \right\}$\;
    }

    $\mathcal{H}
    \leftarrow
    \operatorname{unique}
    \left(
    \mathcal{H}
    \cap
    [h^{(\mathsf{min})},h^{(\mathsf{max})}]
    \right)$\;

    \ForEach{$h\in\mathcal{H}$}{
        \If{the direct horizontal route is collision-free at altitude $h$}{
            Construct the straight route
            $\mathcal{R}_{i,i'}(h)$\;

            $c_{m,i,i'}^{(\mathsf{obs},\mathsf{lvl})}(h)
            \leftarrow 0$\;
        }
        \Else{
            Attempt to construct a feasible Manhattan-detour route
            $\mathcal{R}_{i,i'}(h)$\;

            \eIf{a feasible detour exists}{
                Calculate the detour distance
                $D^{(\mathsf{mht})}(h)$\;

                Calculate
                $c_{m,i,i'}^{(\mathsf{obs},\mathsf{lvl})}(h)$\;
            }{
                $J_{m,i,i'}^{(\mathsf{alt})}(h)
                \leftarrow +\infty$\;
            }
        }

        \If{$J_{m,i,i'}^{(\mathsf{alt})}(h)\neq+\infty$}{
            Calculate
            $c_{m,i,i'}^{(\mathsf{obs},\mathsf{vtl})}(h)$\;

            $c_{m,i,i'}^{(\mathsf{obs})}(h)
            \leftarrow
            c_{m,i,i'}^{(\mathsf{obs},\mathsf{lvl})}(h)
            +
            c_{m,i,i'}^{(\mathsf{obs},\mathsf{vtl})}(h)$\;

            $J_{m,i,i'}^{(\mathsf{alt})}(h)
            \leftarrow
            \gamma_1 f^{(\mathsf{eco})}(h)
            +
            \gamma_2
            c_{m,i,i'}^{(\mathsf{obs})}(h)$\;
        }
    }

    \If{$J_{m,i,i'}^{(\mathsf{alt})}(h)=+\infty$ for all
    $h\in\mathcal{H}$}{
        \KwRet{infeasible UAV--task transition}\;
    }

    $h_m^{(\mathsf{uav},t)*}
    \leftarrow
    \arg\min_{h\in\mathcal{H}}
    J_{m,i,i'}^{(\mathsf{alt})}(h)$\;

    $\mathcal{R}_{i,i'}^{*}
    \leftarrow
    \mathcal{R}_{i,i'}
    \!\left(h_m^{(\mathsf{uav},t)*}\right)$\;

    $d_{i,i'}
    \leftarrow
    \operatorname{length}
    \left(
    \mathcal{R}_{i,i'}^{*}
    \right)$\;

    \tcp{Grid search for the data-collection altitude}

    \If{$h_{i'}^{(\mathsf{uav},\mathsf{min})}
    >
    h^{(\mathsf{max})}$}{
        \KwRet{infeasible UAV--task transition}\;
    }

    \eIf{$h_{i'}^{(\mathsf{uav},\mathsf{min})}
    >
    h^{(\mathsf{eco})}$}{
        $\mathcal{H}_{i'}^{(\mathsf{collect})}
        \leftarrow
        \left\{
        h_{i'}^{(\mathsf{uav},\mathsf{min})}
        \right\}$\;
    }{
        $\mathcal{H}_{i'}^{(\mathsf{collect})}
        \leftarrow
        \left\{
        h_{i'}^{(\mathsf{uav},\mathsf{min})}
        +
        k l^{(\mathsf{search})}
        \,\middle|\,
        k\in\mathbb{Z}_{\geq 0},
        \;
        h_{i'}^{(\mathsf{uav},\mathsf{min})}
        +
        k l^{(\mathsf{search})}
        \leq
        \min
        \left\{
        h^{(\mathsf{eco})},
        h^{(\mathsf{max})}
        \right\}
        \right\}$\;
    }

    \ForEach{$h\in\mathcal{H}_{i'}^{(\mathsf{collect})}$}{
        Calculate the data-collection quality
        $q_m^{(\mathsf{uav},t)}(h)$\;

        Calculate the altitude-adjustment cost
        $c_{m,i'}^{(\mathsf{chg})}
        \!\left(
        h\mid h_m^{(\mathsf{uav},t)*}
        \right)$\;

        $J_{m,i'}^{(\mathsf{collect})}(h)
        \leftarrow
        q_m^{(\mathsf{uav},t)}(h)
        -
        f^{(\mathsf{eco})}(h)
        -
        c_{m,i'}^{(\mathsf{chg})}
        \!\left(
        h\mid h_m^{(\mathsf{uav},t)*}
        \right)$\;
    }

    $h^{(\mathsf{collect})*}
    \leftarrow
    \arg\max_{
    h\in\mathcal{H}_{i'}^{(\mathsf{collect})}
    }
    J_{m,i'}^{(\mathsf{collect})}(h)$\;

    Calculate
    $T_{m,i,i'}^{(\mathsf{vtl})}$
    and
    $E_{m,i,i'}^{(\mathsf{vtl})}$
    induced by
    $h_m^{(\mathsf{uav},t)*}$
    and
    $h^{(\mathsf{collect})*}$\;
\end{algorithm2e}

\begin{algorithm2e}[!htbp]
	\caption{Adaptive discrete speed selection}
	\label{alg:speed_selection}
	\footnotesize

	\KwIn{UAV $u_m$; horizontal route distance $d_{i,i'}$;
	earliest available timeslot $t_m$; task starting timeslot
	$\mathbbm{t}_{i'}^{(\mathsf{start})}$ and deadline
	$\mathbbm{t}_{i'}^{(\mathsf{end})}$; vertical-motion time
	$T_{m,i,i'}^{(\mathsf{vtl})}$ and energy
	$E_{m,i,i'}^{(\mathsf{vtl})}$; candidate horizontal-speed set
	$\mathcal{V}$; timeslot duration $\Delta t$; delay coefficient
	$\omega$; movement- and hovering-energy rates
	$e_m^{(\mathsf{mov},v)}$ and $e_m^{(\mathsf{hov})}$;
	current UAV motion state and applicable kinematic constraints}

	\KwOut{Selected speed $v^*$; predicted arrival timeslot
	$\mathbbm{t}_m^{(\mathsf{arr},*)}$; corresponding energy cost
	$C_{m,i,i'}^{(\mathsf{energy})*}$ and speed-selection cost
	$J_{m,i,i'}^*$}

	\tcp{Construct the kinematically feasible speed set}

	$\mathcal{V}_{m,i,i'}^{(\mathsf{feas})}
	\leftarrow
	\left\{
	v\in\mathcal{V}
	\,\middle|\,
	v\ \text{satisfies the current speed and acceleration constraints}
	\right\}$\;

	\If{$\mathcal{V}_{m,i,i'}^{(\mathsf{feas})}=\emptyset$}{
		\Return{infeasible UAV--task transition}\;
	}

	$J_{m,i,i'}^{(\mathsf{min})}\leftarrow+\infty$\;
	$v^*\leftarrow\varnothing$\;
	$\mathbbm{t}_m^{(\mathsf{arr},*)}\leftarrow+\infty$\;

	\ForEach{$v\in\mathcal{V}_{m,i,i'}^{(\mathsf{feas})}$ in a fixed order}{
		Calculate the horizontal flight time:~$T_{m,i,i'}^{(\mathsf{hor})}(v)\leftarrow\frac{d_{i,i'}}{v}$;

		Predict the arrival timeslot:~$\mathbbm{t}_m^{(\mathsf{arr},v)}\leftarrow t_m+\left\lceil\frac{T_{m,i,i'}^{(\mathsf{hor})}(v)
		+T_{m,i,i'}^{(\mathsf{vtl})}}{\Delta t}\right\rceil$;

		\eIf{
			$\mathbbm{t}_m^{(\mathsf{arr},v)}
			\geq
			\mathbbm{t}_{i'}^{(\mathsf{end})}$
		}{
			$C_{m,i,i'}^{(\mathsf{time})}(v)
			\leftarrow+\infty$\;
		}{
			\eIf{
				$\mathbbm{t}_m^{(\mathsf{arr},v)}
				\leq
				\mathbbm{t}_{i'}^{(\mathsf{start})}$
			}{
				$C_{m,i,i'}^{(\mathsf{time})}(v)
				\leftarrow 0$\;
			}{
				$C_{m,i,i'}^{(\mathsf{time})}(v)
				\leftarrow
				\omega
				\left(
				\mathbbm{t}_m^{(\mathsf{arr},v)}
				-
				\mathbbm{t}_{i'}^{(\mathsf{start})}
				\right)$\;
			}
		}

		Calculate the energy cost:~$
		\begin{aligned}
		C_{m,i,i'}^{(\mathsf{energy})}(v)
		\leftarrow\;&
		e_m^{(\mathsf{mov},v)}
		T_{m,i,i'}^{(\mathsf{hor})}(v)
		+
		E_{m,i,i'}^{(\mathsf{vtl})}
		+
		e_m^{(\mathsf{hov})}
		\max
		\left\{
		\mathbbm{t}_{i'}^{(\mathsf{start})}
		-
		\mathbbm{t}_m^{(\mathsf{arr},v)},
		0
		\right\}
		\end{aligned}$;    

		$J_{m,i,i'}(v)
		\leftarrow
		C_{m,i,i'}^{(\mathsf{energy})}(v)
		+
		C_{m,i,i'}^{(\mathsf{time})}(v)$\;

		\If{$J_{m,i,i'}(v)<J_{m,i,i'}^{(\mathsf{min})}$}{
			$J_{m,i,i'}^{(\mathsf{min})}
			\leftarrow
			J_{m,i,i'}(v)$\;

			$v^*\leftarrow v$\;

			$\mathbbm{t}_m^{(\mathsf{arr},*)}
			\leftarrow
			 \mathbbm{t}_m^{(\mathsf{arr},v)}$\;

			$C_{m,i,i'}^{(\mathsf{energy})*}
			\leftarrow
			 C_{m,i,i'}^{(\mathsf{energy})}(v)$\;
		}
	}

	\If{$v^*=\varnothing$}{
		\Return{infeasible UAV--task transition}\;
	}

	$J_{m,i,i'}^*
	\leftarrow
	J_{m,i,i'}^{(\mathsf{min})}$\;

	\Return{$v^*$,
	$\mathbbm{t}_m^{(\mathsf{arr},*)}$,
	$C_{m,i,i'}^{(\mathsf{energy})*}$,
	and $J_{m,i,i'}^*$}\;
\end{algorithm2e}
	
	
		
		
		
		


\begin{algorithm2e}[!htbp]
    \caption{Basic evolutionary update integrated with L\'evy flight}
    \label{alg:levy_basic_evolution_short}
    \small

    \KwIn{Current population
    $\mathcal{X}^{(g)}
    =
    \{\mathbf{x}_o^{(g)}\}_{o=1}^{N^{(\mathsf{p})}}$;
    fitness values
    $\{F(\mathbf{x}_o^{(g)})\}_{o=1}^{N^{(\mathsf{p})}}$;
    historical best priority vector
    $\mathbf{x}^{(\mathsf{best},g)}$;
    current generation index $g$;
    maximum number of generations $G^{(\mathsf{max})}$;
    discoverer ratio $\rho^{(\mathsf{f})}$;
    warning threshold $S^{(\mathsf{st})}$;
    local perturbation coefficient $\sigma$;
    exploratory perturbation coefficient $\varepsilon$;
    L\'evy-flight step-size coefficient $\zeta$}

    \KwOut{Provisionally updated population
    $\widetilde{\mathcal{X}}^{(g+1)}
    =
    \{\widetilde{\mathbf{x}}_o^{(g+1)}\}_{o=1}^{N^{(\mathsf{p})}}$}

    $D\leftarrow N^{(\mathsf{task})}$\;

    $N^{(\mathsf{f})}
    \leftarrow
    \left\lfloor
    \rho^{(\mathsf{f})}N^{(\mathsf{p})}
    \right\rfloor$\;

    $\tau^{(\mathsf{f})}
    \leftarrow
    \left\lfloor
    \dfrac{
    \left(1+\rho^{(\mathsf{f})}\right)
    N^{(\mathsf{p})}
    }{2}
    \right\rfloor$\;

    $a_g
    \leftarrow
    1-\dfrac{g}{G^{(\mathsf{max})}}$\;

    Sort $\mathcal{X}^{(g)}$ in descending order of fitness,
    using a fixed deterministic tie-breaking rule, and reindex
    the individuals according to their ranks\;

    \tcp{Discoverer update}

    \For{$o\leftarrow 1$ \KwTo $N^{(\mathsf{f})}$}{
        Generate
        $r_o\sim\mathcal{U}(0,1)$,
        $\mathbf{r}_o\sim\mathcal{U}(0,1)^D$, and
        $\mathbf{z}_o\sim\mathcal{N}(\mathbf{0},\mathbf{I}_D)$\;

        \eIf{$r_o<S^{(\mathsf{st})}$}{
            $\widetilde{\mathbf{x}}_o^{(g+1)}
            \leftarrow
            \Pi_{[0,1]^D}
            \left[
            \mathbf{x}_o^{(g)}
            +
            a_g\mathbf{r}_o
            \odot
            \left(
            \mathbf{x}^{(\mathsf{best},g)}
            -
            \mathbf{x}_o^{(g)}
            \right)
            +
            \sigma a_g\mathbf{z}_o
            \right]$\;
        }{
            $\widetilde{\mathbf{x}}_o^{(g+1)}
            \leftarrow
            \Pi_{[0,1]^D}
            \left[
            \mathbf{x}_o^{(g)}
            +
            \varepsilon a_g\mathbf{z}_o
            \right]$\;
        }
    }

    \tcp{Joiner update}

    \For{$o\leftarrow N^{(\mathsf{f})}+1$
    \KwTo $N^{(\mathsf{p})}$}{

        $\mathbf{d}_o^{(g)}
        \leftarrow
        \mathbf{x}_o^{(g)}
        -
        \mathbf{x}^{(\mathsf{best},g)}$\;

        \eIf{$o\leq\tau^{(\mathsf{f})}$}{
            Generate
            $\boldsymbol{\eta}_o
            \sim
            \mathcal{U}(0,1)^D$
            and
            $\mathbf{z}_o
            \sim
            \mathcal{N}(\mathbf{0},\mathbf{I}_D)$\;

            $\widetilde{\mathbf{x}}_o^{(g+1)}
            \leftarrow
            \Pi_{[0,1]^D}
            \left[
            \mathbf{x}^{(\mathsf{best},g)}
            +
            a_g
            \left(
            \boldsymbol{\eta}_o
            -
            \dfrac{1}{2}\mathbf{1}
            \right)
            \odot
            \mathbf{d}_o^{(g)}
            +
            \sigma a_g\mathbf{z}_o
            \right]$\;
        }{
            Generate a $D$-dimensional L\'evy-flight vector
            $\mathbf{L}_o^{(g)}(D)$ using the Mantegna algorithm\;

            $\widetilde{\mathbf{x}}_o^{(g+1)}
            \leftarrow
            \Pi_{[0,1]^D}
            \left[
            \mathbf{x}^{(\mathsf{best},g)}
            +
            \zeta a_g
            \mathbf{L}_o^{(g)}(D)
            \right]$\;
        }
    }

    \Return{$\widetilde{\mathcal{X}}^{(g+1)}$}\;
\end{algorithm2e}

\begin{algorithm2e}[!htbp]
    \caption{Danger-aware diversification, fitness update, and elitist preservation}
    \label{alg:danger_awareness_update_short}
    \footnotesize

    \KwIn{Current population
    $\mathcal{X}^{(g)}
    =
    \{\mathbf{x}_o^{(g)}\}_{o=1}^{N^{(\mathsf{p})}}$
    and fitness values
    $\{F(\mathbf{x}_o^{(g)})\}_{o=1}^{N^{(\mathsf{p})}}$;
    provisionally updated population
    $\widetilde{\mathcal{X}}^{(g+1)}
    =
    \{\widetilde{\mathbf{x}}_o^{(g+1)}\}_{o=1}^{N^{(\mathsf{p})}}$;
    historical best priority vector
    $\mathbf{x}^{(\mathsf{best},g)}$;
    stagnation counter $s^{(g)}$;
    generation index $g$ and maximum number of generations
    $G^{(\mathsf{max})}$;
    number of danger-aware individuals $N^{(\mathsf{d})}$;
    diversity threshold $\mathcal{D}^{(\mathsf{min})}$;
    stagnation threshold $S^{(\mathsf{stall})}$;
    fitness tolerance $\epsilon_F$;
    perturbation coefficient $\delta$}

    \KwOut{Updated population
    $\mathcal{X}^{(g+1)}
    =
    \{\mathbf{x}_o^{(g+1)}\}_{o=1}^{N^{(\mathsf{p})}}$;
    updated fitness values;
    historical best priority vector
    $\mathbf{x}^{(\mathsf{best},g+1)}$;
    updated stagnation counter $s^{(g+1)}$}

    $D\leftarrow N^{(\mathsf{task})}$\;

    $a_g
    \leftarrow
    1-\dfrac{g}{G^{(\mathsf{max})}}$\;

    \tcp{Detect population convergence degradation}

    Construct the task-priority sequences
    $\{\boldsymbol{\pi}_o^{(g)}\}_{o=1}^{N^{(\mathsf{p})}}$
    and
    $\boldsymbol{\pi}^{(\mathsf{best},g)}$
    according to (\ref{eq:priority_order})\;

    $\mathcal{D}^{(g)}
    \leftarrow
    \dfrac{1}{N^{(\mathsf{p})}}
    \displaystyle\sum_{o=1}^{N^{(\mathsf{p})}}
    d_{\mathsf{K}}
    \left(
    \boldsymbol{\pi}_o^{(g)},
    \boldsymbol{\pi}^{(\mathsf{best},g)}
    \right)$\;

    \For{$o\leftarrow 1$ \KwTo $N^{(\mathsf{p})}$}{
        $\mathbf{x}_o^{(g+1)}
        \leftarrow
        \widetilde{\mathbf{x}}_o^{(g+1)}$\;
    }

    \If{
        $\mathcal{D}^{(g)}
        \leq
        \mathcal{D}^{(\mathsf{min})}$
        \textbf{or}
        $s^{(g)}
        \geq
        S^{(\mathsf{stall})}$
    }{
        Select
        $\min\{N^{(\mathsf{d})},N^{(\mathsf{p})}-1\}$
        non-elite individuals to form
        $\mathcal{K}_g$\;

        \tcp{Danger-aware adjustment}

        \ForEach{$o\in\mathcal{K}_g$}{
            \eIf{
                $F(\mathbf{x}_o^{(g)})
                <
                F(\mathbf{x}^{(\mathsf{best},g)})
                -
                \epsilon_F$
            }{
                Generate
                $\boldsymbol{\epsilon}_o
                \sim
                \mathcal{U}(0,1)^D$\;

                $\mathbf{x}_o^{(g+1)}
                \leftarrow
                \Pi_{[0,1]^D}
                \left[
                \widetilde{\mathbf{x}}_o^{(g+1)}
                +
                \boldsymbol{\epsilon}_o
                \odot
                \left(
                \mathbf{x}^{(\mathsf{best},g)}
                -
                \widetilde{\mathbf{x}}_o^{(g+1)}
                \right)
                \right]$\;
            }{
                Generate
                $\boldsymbol{\xi}_o
                \sim
                \mathcal{N}(\mathbf{0},\mathbf{I}_D)$\;

                $\mathbf{x}_o^{(g+1)}
                \leftarrow
                \Pi_{[0,1]^D}
                \left[
                \widetilde{\mathbf{x}}_o^{(g+1)}
                +
                \delta a_g\boldsymbol{\xi}_o
                \right]$\;
            }
        }
    }

    \tcp{Re-decoding and fitness evaluation}

    \For{$o\leftarrow 1$ \KwTo $N^{(\mathsf{p})}$}{
        Decode $\mathbf{x}_o^{(g+1)}$ using the hierarchical
        decoder in SubM B.2\;

        Calculate
        $F(\mathbf{x}_o^{(g+1)})$
        according to (\ref{eq:reward_net})\;
    }

    $\mathbf{x}^{(\mathsf{cand},g+1)}
    \leftarrow
    \arg\max_{
    \mathbf{x}\in\mathcal{X}^{(g+1)}
    }
    F(\mathbf{x})$\;

    \tcp{Historical-best and stagnation update}

    \eIf{
        $F(\mathbf{x}^{(\mathsf{cand},g+1)})
        >
        F(\mathbf{x}^{(\mathsf{best},g)})
        +
        \epsilon_F$
    }{
        $\mathbf{x}^{(\mathsf{best},g+1)}
        \leftarrow
        \mathbf{x}^{(\mathsf{cand},g+1)}$\;

        $s^{(g+1)}\leftarrow 0$\;
    }{
        $\mathbf{x}^{(\mathsf{best},g+1)}
        \leftarrow
        \mathbf{x}^{(\mathsf{best},g)}$\;

        $s^{(g+1)}
        \leftarrow
        s^{(g)}+1$\;
    }

    \tcp{Elitist preservation}

    \If{
        $\mathbf{x}^{(\mathsf{best},g+1)}
        \notin
        \mathcal{X}^{(g+1)}$
    }{
        $o^{(\mathsf{worst})}
        \leftarrow
        \arg\min_{
        1\leq o\leq N^{(\mathsf{p})}
        }
        F(\mathbf{x}_o^{(g+1)})$\;

        $\mathbf{x}_{o^{(\mathsf{worst})}}^{(g+1)}
        \leftarrow
        \mathbf{x}^{(\mathsf{best},g+1)}$\;

        $F(\mathbf{x}_{o^{(\mathsf{worst})}}^{(g+1)})
        \leftarrow
        F(\mathbf{x}^{(\mathsf{best},g+1)})$\;
    }

    Sort $\mathcal{X}^{(g+1)}$ in descending order of fitness
    using a fixed deterministic tie-breaking rule\;

    \Return{
    $\mathcal{X}^{(g+1)}$,
    $\{F(\mathbf{x}_o^{(g+1)})\}_{o=1}^{N^{(\mathsf{p})}}$,
    $\mathbf{x}^{(\mathsf{best},g+1)}$,
    and $s^{(g+1)}$
    }\;
\end{algorithm2e}

\begin{algorithm2e}[!htbp]
    \caption{Greedy construction of candidate cooperative groups for an emergent Type-$\mathsf{III}$ task}
    \label{alg:online_replanning_stage2}
    \footnotesize

    \KwIn{
        Emergent task $p_i^{(\mathsf{III},t)}$ appearing at timeslot $t$;
        current fleet state
        $\mathcal{W}(t)
        =
        \{(\mathbf{l}_m,t_m,E_m)\mid u_m\in\mathcal{U}\}$;
        UAV set $\mathcal{U}$;
        task data requirement $d_i^{(\mathsf{III},t)}$;
        active time window $[t,\mathbbm{t}_i^{(-,t)}]$;
        timeslot duration $\Delta t$;
        collision-risk threshold $\theta^{(\mathsf{max})}$;
        numerical constant $\epsilon_C>0$
    }

    \KwOut{
        Candidate cooperative set $\mathcal{G}_i$
    }

    $\mathcal{U}_i^{(\mathsf{cand})}
    \leftarrow
    \emptyset$\;

    \tcp{Construct the candidate feasible UAV set}

    \ForEach{$u_m\in\mathcal{U}$}{
        Use Algorithms~\ref{alg:high} and
        \ref{alg:speed_selection} to construct the earliest
        feasible local insertion from the current state of
        $u_m$ to $p_i^{(\mathsf{III},t)}$\;

        \If{no feasible insertion route exists}{
            \textbf{continue}\;
        }

        Obtain the flight time
        $\tau_{m,i}^{(\mathsf{fly})}$
        and calculate $\mathbbm{t}_{m,i}^{(\mathsf{arr})}\leftarrow t_m+\tau_{m,i}^{(\mathsf{fly})}$\;

        $\mathbbm{t}_{m,i}^{(\mathsf{start})}
        \leftarrow
        \max
        \left\{
        \mathbbm{t}_{m,i}^{(\mathsf{arr})},
        t
        \right\}$\;

        \If{
            $\mathbbm{t}_{m,i}^{(\mathsf{start})}
            >
            \mathbbm{t}_i^{(-,t)}$
        }{
            \textbf{continue}\;
        }

        $\Delta\mathbbm{t}_{m,i}^{(\mathsf{eff})}
        \leftarrow
        \left(
        \mathbbm{t}_i^{(-,t)}
        -
        \mathbbm{t}_{m,i}^{(\mathsf{start})}
        +
        1
        \right)
        \Delta t$\;

        $d_{m,i}^{(\mathsf{avail})}
        \leftarrow
        d_m^{(\mathsf{uav})}
        \Delta\mathbbm{t}_{m,i}^{(\mathsf{eff})}$\;

        Estimate the required energy
        $e_{m,i}^{(\mathsf{req})}$
        for reaching and serving the emergent task,
        reconnecting to the remaining offline path,
        completing all retained tasks, and reaching the
        designated destination\;

        Evaluate
        $\chi_{m,i}^{(\mathsf{rem})}\in\{0,1\}$,
        where $\chi_{m,i}^{(\mathsf{rem})}=1$ indicates that
        the remaining offline mission remains feasible after
        the local insertion\;

        Estimate the preliminary insertion cost
        $\widehat{C}_{m,i}^{(\mathsf{ins})}$\;

        \If{
            $E_m\geq e_{m,i}^{(\mathsf{req})}$
            \textbf{and}
            $\chi_{m,i}^{(\mathsf{rem})}=1$
        }{
            $\psi_{m,i}
            \leftarrow
            \dfrac{
            d_{m,i}^{(\mathsf{avail})}
            }{
            \widehat{C}_{m,i}^{(\mathsf{ins})}
            +
            \epsilon_C
            }$\;

            $\mathcal{U}_i^{(\mathsf{cand})}
            \leftarrow
            \mathcal{U}_i^{(\mathsf{cand})}
            \cup
            \{u_m\}$\;
        }
    }

    \If{$\mathcal{U}_i^{(\mathsf{cand})}=\emptyset$}{
        $\mathcal{G}_i\leftarrow\emptyset$\;
        \Return{$\mathcal{G}_i$}\;
    }

    Sort the UAVs in $\mathcal{U}_i^{(\mathsf{cand})}$
    in descending order of $\psi_{m,i}$ using a fixed
    deterministic tie-breaking rule, and denote the resulting
    ordered list by $\mathcal{O}_i$\;

    $\mathcal{G}_i\leftarrow\emptyset$\;

    \tcp{Multi-start greedy construction of cooperative groups}

    \ForEach{seed UAV $u_s\in\mathcal{O}_i$}{
        $G_i^{(s)}\leftarrow\{u_s\}$\;

        $D_i^{(s)}
        \leftarrow
        d_{s,i}^{(\mathsf{avail})}$\;

        \If{
            $\theta_i(G_i^{(s)})
            >
            \theta^{(\mathsf{max})}$
        }{
            \textbf{continue}\;
        }

        \ForEach{$u_m\in\mathcal{O}_i\setminus G_i^{(s)}$}{
            \If{
                $D_i^{(s)}
                \geq
                d_i^{(\mathsf{III},t)}$
            }{
                \textbf{break}\;
            }

            $\widehat{G}
            \leftarrow
            G_i^{(s)}\cup\{u_m\}$\;

            \If{
                $\theta_i(\widehat{G})
                \leq
                \theta^{(\mathsf{max})}$
            }{
                $G_i^{(s)}
                \leftarrow
                \widehat{G}$\;

                $D_i^{(s)}
                \leftarrow
                D_i^{(s)}
                +
                d_{m,i}^{(\mathsf{avail})}$\;
            }
        }

        \If{
            $D_i^{(s)}
            \geq
            d_i^{(\mathsf{III},t)}$
        }{
            $\mathcal{G}_i
            \leftarrow
            \mathcal{G}_i
            \cup
            \{G_i^{(s)}\}$\;
        }
    }

    Remove duplicate cooperative groups from
    $\mathcal{G}_i$\;

    \Return{$\mathcal{G}_i$}\;
\end{algorithm2e}

\begin{algorithm2e}[!htbp]
    \caption{Net-benefit evaluation and acceptance of greedily generated cooperative groups}
    \label{alg:profit_greedy_online}
    \footnotesize

    \KwIn{
        Emergent task $p_i^{(\mathsf{III},t)}$;
        candidate cooperative set $\mathcal{G}_i$ generated by
        Alg.~\ref{alg:online_replanning_stage2}, together with
        the corresponding feasible local-insertion plans;
        task reward $r_i^{(\mathsf{III},t)}$
    }

    \KwOut{
        Accepted cooperative group $G_i^{(*)}$;
        maximum candidate cooperative net benefit
        $\Delta J_i^{(\mathsf{max})}$
    }

    $\widehat{G}_i\leftarrow\emptyset$\;
    $\Delta J_i^{(\mathsf{max})}\leftarrow-\infty$\;

    \If{$\mathcal{G}_i=\emptyset$}{
        $G_i^{(*)}\leftarrow\emptyset$\;
        \Return{$G_i^{(*)},
        \Delta J_i^{(\mathsf{max})}$}\;
    }

    \tcp{Exact net-benefit evaluation over the generated candidates}

    \ForEach{$G_i^{(g)}\in\mathcal{G}_i$}{
        $\widetilde{r}_i^{(\mathsf{III},t,g)}
        \leftarrow
        \dfrac{
        r_i^{(\mathsf{III},t)}
        }{
        |G_i^{(g)}|
        }$\;

        $\Delta J_i^{(\mathsf{III},t)}
        \left(
        G_i^{(g)}
        \right)
        \leftarrow 0$\;

        Determine the group-specific local-insertion timing
        according to the feasible insertion plans associated
        with $G_i^{(g)}$\;

        \ForEach{$u_m\in G_i^{(g)}$}{
            Evaluate the normalized additional flight cost
            $C_{m,i}^{(\mathsf{fly})}$\;

            Evaluate the normalized hovering cost
            $C_{m,i}^{(\mathsf{hov})}$\;

            Evaluate the normalized path-disruption cost
            $C_{m,i}^{(\mathsf{dist})}$\;

            Evaluate the normalized incremental ecological cost
            $C_{m,i}^{(\mathsf{eco})}$\;

            $\Delta J_{m,i}^{(\mathsf{III},t,g)}
            \leftarrow
            \widetilde{r}_i^{(\mathsf{III},t,g)}
            -
            C_{m,i}^{(\mathsf{fly})}
            -
            C_{m,i}^{(\mathsf{hov})}
            -
            C_{m,i}^{(\mathsf{dist})}
            -
            C_{m,i}^{(\mathsf{eco})}$\;

            $\Delta J_i^{(\mathsf{III},t)}
            \left(
            G_i^{(g)}
            \right)
            \leftarrow
            \Delta J_i^{(\mathsf{III},t)}
            \left(
            G_i^{(g)}
            \right)
            +
            \Delta J_{m,i}^{(\mathsf{III},t,g)}$\;
        }

        \If{
            $\Delta J_i^{(\mathsf{III},t)}
            \left(
            G_i^{(g)}
            \right)
            >
            \Delta J_i^{(\mathsf{max})}$
        }{
            $\Delta J_i^{(\mathsf{max})}
            \leftarrow
            \Delta J_i^{(\mathsf{III},t)}
            \left(
            G_i^{(g)}
            \right)$\;

            $\widehat{G}_i
            \leftarrow
            G_i^{(g)}$\;
        }
        \ElseIf{
            $\Delta J_i^{(\mathsf{III},t)}
            \left(
            G_i^{(g)}
            \right)
            =
            \Delta J_i^{(\mathsf{max})}$
            under numerical precision
        }{
            Retain the preferred group between
            $G_i^{(g)}$ and $\widehat{G}_i$ according to a fixed
            deterministic tie-breaking rule\;
        }
    }

    \tcp{Accept or reject the local re-planning decision}

    \eIf{$\Delta J_i^{(\mathsf{max})}>0$}{
        $G_i^{(*)}\leftarrow\widehat{G}_i$\;
    }{
        $G_i^{(*)}\leftarrow\emptyset$\;
    }

    \Return{$G_i^{(*)},
    \Delta J_i^{(\mathsf{max})}$}\;
\end{algorithm2e}

\newpage
\clearpage
\section{Computational Complexity Analysis}
\label{complexity}
\noindent
We analyze the computational complexity of the three modules in FORTUNE. Let $N^{(\mathsf{enc})}$ denote the number of Transformer encoder layers, and let $N^{(\mathsf{tr})}$ and $N^{(\mathsf{ep})}$ denote the total number of training sequences and training epochs, respectively. For Module B, define
$N^{(\mathsf{alt})}=|\mathcal{H}|$,
$N^{(\mathsf{col})}
=
\max_{i'}|\mathcal{H}_{i'}^{(\mathsf{collect})}|$,
$N^{(\mathsf{spd})}=|\mathcal{V}|$, and
$B^{(\mathsf{seg})}
=
\max_{i,i'}
|\mathbbm{b}_{i,i'}^{(\mathsf{path})}|$.
Here, $N^{(\mathsf{alt})}\leq 4$ and
$N^{(\mathsf{spd})}=3$ under the adopted implementation.
We assume that the obstacle-intersection and detour checks for one candidate altitude require
$\mathcal{O}(B^{(\mathsf{seg})})$ operations.

\noindent
\textit{For Module A}, the self-attention operation of one Transformer encoder layer over an input sequence of length
$L^{(\mathsf{his})}$ has complexity
$\mathcal{O}((L^{(\mathsf{his})})^2d^{(\mathsf{model})})$.
The linear projections and position-wise feed-forward network require
$\mathcal{O}(L^{(\mathsf{his})}(d^{(\mathsf{model})})^2)$
operations when the feed-forward width is proportional to
$d^{(\mathsf{model})}$. Therefore, the offline training complexity is
\begin{equation}
\mathcal{C}_{\mathsf{A}}^{(\mathsf{train})}
=
\mathcal{O}
\left(
N^{(\mathsf{ep})}
N^{(\mathsf{tr})}
N^{(\mathsf{enc})}
\left[
(L^{(\mathsf{his})})^2d^{(\mathsf{model})}
+
L^{(\mathsf{his})}(d^{(\mathsf{model})})^2
\right]
\right).
\end{equation}
Before UAV take-off, inference is conducted for all Type-$\mathsf{II}$ PoIs. Including the linear prediction head and the threshold-based extraction of active windows, its complexity is
\begin{equation}
\begin{aligned}
\mathcal{C}_{\mathsf{A}}^{(\mathsf{inf})}
=
\mathcal{O}
\Big(
|\mathcal{P}^{(\mathsf{II})}|
\big[
N^{(\mathsf{enc})}
\big(
(L^{(\mathsf{his})})^2d^{(\mathsf{model})}
+
L^{(\mathsf{his})}(d^{(\mathsf{model})})^2
\big)
+
L^{(\mathsf{pred})}d^{(\mathsf{model})}
+
L^{(\mathsf{pred})}
\big]
\Big).
\end{aligned}
\end{equation}
Since model training is performed offline, \textbf{only
$\mathcal{C}_{\mathsf{A}}^{(\mathsf{inf})}$
contributes to the mission-preparation latency}.

\noindent
\textit{For Module B}, decoding one task for one UAV requires evaluating the candidate cruising altitudes, collection altitudes, and discrete speeds. The corresponding UAV--task transition complexity is
\begin{equation}
\mathcal{C}^{(\mathsf{trans})}
=
\mathcal{O}
\left(
N^{(\mathsf{alt})}B^{(\mathsf{seg})}
+
N^{(\mathsf{col})}
+
N^{(\mathsf{spd})}
\right).
\end{equation}
For one priority-encoded individual, sorting its
$N^{(\mathsf{task})}$ priority weights requires
$\mathcal{O}(N^{(\mathsf{task})}\log N^{(\mathsf{task})})$.
For each task, the decoder examines all $|\mathcal{U}|$ UAVs and ranks the feasible UAV candidates, leading to
\begin{equation}
\begin{aligned}
\mathcal{C}^{(\mathsf{dec})}
=
\mathcal{O}
\Big(
N^{(\mathsf{task})}\log N^{(\mathsf{task})}
+
N^{(\mathsf{task})}|\mathcal{U}|
\big[
\mathcal{C}^{(\mathsf{trans})}
+
\log|\mathcal{U}|
\big]
+
|\mathcal{U}||\mathcal{T}|
\Big),
\end{aligned}
\end{equation}
where the final term corresponds to a direct timeslot-wise evaluation of the decoded net reward; this term can be reduced if the reward and ecological costs are accumulated during decoding.

Tent-map initialization requires
$\mathcal{O}(N^{(\mathsf{p})}N^{(\mathsf{task})})$
operations. In each generation, the discoverer--joiner updates require
$\mathcal{O}(N^{(\mathsf{p})}N^{(\mathsf{task})})$,
while population sorting requires
$\mathcal{O}(N^{(\mathsf{p})}\log N^{(\mathsf{p})})$.
Using an inversion-count implementation, the Kendall-distance-based diversity evaluation requires
$\mathcal{O}(N^{(\mathsf{p})}N^{(\mathsf{task})}
\log N^{(\mathsf{task})})$.
The dominant operation is the re-decoding and fitness evaluation of all individuals. Consequently, the total complexity of Module B is
\begin{equation}
\begin{aligned}
\mathcal{C}_{\mathsf{B}}
=
\mathcal{O}
\Big(
N^{(\mathsf{p})}N^{(\mathsf{task})}
+
(G^{(\mathsf{max})}+1)
N^{(\mathsf{p})}
\mathcal{C}^{(\mathsf{dec})}
+
G^{(\mathsf{max})}
N^{(\mathsf{p})}
N^{(\mathsf{task})}
\log N^{(\mathsf{task})}
+
G^{(\mathsf{max})}
N^{(\mathsf{p})}
\log N^{(\mathsf{p})}
\Big).
\end{aligned}
\end{equation}
Since hierarchical decoding dominates the vector-level evolutionary operations, the above can be compactly expressed as
\begin{equation}
\mathcal{C}_{\mathsf{B}}
=
\mathcal{O}
\left(
G^{(\mathsf{max})}
N^{(\mathsf{p})}
N^{(\mathsf{task})}
\left[
\log N^{(\mathsf{task})}
+
|\mathcal{U}|
\left(
N^{(\mathsf{alt})}B^{(\mathsf{seg})}
+
N^{(\mathsf{col})}
+
N^{(\mathsf{spd})}
+
\log|\mathcal{U}|
\right)
\right]
\right),
\end{equation}
apart from the optional
$\mathcal{O}(G^{(\mathsf{max})}N^{(\mathsf{p})}
|\mathcal{U}||\mathcal{T}|)$
timeslot-wise fitness-evaluation term. Similar to Module A, Module B is executed offline; thus, its computational overhead is incurred before the online stage and has less impact on real-time performance.

\noindent
\textit{For Module C}, let
$M_i=|\mathcal{U}_i^{(\mathsf{cand})}|$
denote the number of candidate feasible UAVs for emergent task
$p_i^{(\mathsf{III},t)}$, and let
$N^{(\mathsf{rem})}$ denote the maximum number of remaining offline tasks examined when verifying whether a local insertion preserves the original mission. The feasibility screening of one UAV, including local route generation, altitude and speed selection, reconnection, and remaining-plan verification, has complexity
\begin{equation}
\mathcal{C}^{(\mathsf{ins})}
=
\mathcal{O}
\left(
N^{(\mathsf{alt})}B^{(\mathsf{seg})}
+
N^{(\mathsf{col})}
+
N^{(\mathsf{spd})}
+
N^{(\mathsf{rem})}
\right).
\end{equation}
Screening all UAVs therefore requires
$\mathcal{O}(|\mathcal{U}|\mathcal{C}^{(\mathsf{ins})})$,
and sorting the candidate UAVs requires
$\mathcal{O}(M_i\log M_i)$.
The multi-start greedy construction uses each candidate UAV once as a seed and scans at most $M_i$ UAVs for each seed, resulting in
$\mathcal{O}(M_i^2)$
incremental feasibility checks. Algorithm~\ref{alg:profit_greedy_online} evaluates at most $M_i$ generated groups, each containing at most $M_i$ UAVs, and therefore also requires
$\mathcal{O}(M_i^2)$.
Thus, the complexity of processing one Type-$\mathsf{III}$ task is
\begin{equation}
\mathcal{C}_{\mathsf{C},i}
=
\mathcal{O}
\left(
|\mathcal{U}|\mathcal{C}^{(\mathsf{ins})}
+
M_i\log M_i
+
M_i^2
\right).
\end{equation}
Since $M_i\leq|\mathcal{U}|$, a task-independent upper bound is
\begin{equation}
\mathcal{C}_{\mathsf{C},i}
=
\mathcal{O}
\left(
|\mathcal{U}|\mathcal{C}^{(\mathsf{ins})}
+
|\mathcal{U}|^2
\right).
\end{equation}
Across the whole mission, the online complexity is
\begin{equation}
\mathcal{C}_{\mathsf{C}}
=
\mathcal{O}
\left(
\sum_{t\in\mathcal{T}}
\sum_{i=1}^{|\mathcal{P}^{(\mathsf{III},t)}|}
\left[
|\mathcal{U}|\mathcal{C}^{(\mathsf{ins})}
+
M_i^2
\right]
\right).
\end{equation}
Therefore, Module C avoids the
$\mathcal{O}(2^{M_i})$
complexity of exhaustive cooperative-group enumeration and maintains polynomial online complexity, which supports real-time path refinement for emergent Type-$\mathsf{III}$ tasks.

\newpage
\clearpage
\section{Rationale on F3TSR}
\label{F3TSR}

\noindent
We specifically evaluate F3TSR because the performance on later online tasks is inherently coupled with historical task execution and accumulated resource consumption, making it difficult to isolate the effectiveness of the online planning strategy itself. In particular, UAVs operate under limited onboard energy budgets, and every online task completed in earlier timeslots consumes flight and sensing resources that cannot be replenished during the mission. Such cumulative energy expenditure progressively reduces the maneuverability and reachable sensing region of each UAV, thereby directly affecting its ability to respond to newly emerging Type-$\mathsf{III}$ PoIs in subsequent timeslots. Consequently, the completion rate of later Type-$\mathsf{III}$ tasks is influenced not only by the real-time decision-making capability of an algorithm, but also by the historical planning outcomes and resource allocation decisions accumulated over previous timeslots. Comparing algorithms using the overall Type-$\mathsf{III}$ completion ratio would therefore inevitably mix the effectiveness of online adaptation with the long-term impact of prior decisions, leading to a biased evaluation. To eliminate this temporal coupling effect and ensure a fair comparison, we evaluate only the Type-$\mathsf{III}$ tasks generated in the first timeslot, where all algorithms start from identical initial energy budgets, identical UAV states, and the same offline pre-planned trajectories. Under these controlled conditions, the observed performance differences can be attributed solely to the quality of the online adjustment strategy. Therefore, F3TSR serves as a representative and unbiased indicator for assessing the real-time responsiveness and online task-handling capability of different algorithms.

\section{Parameter Settings}
\label{parameter}
\noindent
We also show tables related to different parameter settings.
\begin{table}[htbp]
\centering
\footnotesize
\centering
        \caption{Hyperparameters for Transformer and LSTM}
        \label{transformerinfo}
        \begin{tabular}{lcc}
            \toprule
            \textbf{Parameter} & \textbf{Transformer} & \textbf{LSTM} \\
            \midrule
            Hidden Size        & 64   & 64   \\
            Layers             & 2    & 2    \\
            Heads              & 4    & N/A  \\
            Learning Rate      & 0.01 & 0.01 \\
            Batch Size         & 32   & 32   \\
            Epochs             & 300  & 300  \\
            Optimizer          & Adam & Adam \\
            \bottomrule
        \end{tabular}
\end{table}

\begin{table}[htbp]
\centering
\footnotesize
\centering
        \caption{Scenario parameter settings for offline pre-planning stage based on UTD19}
        \label{tab:offline_params}
        \begin{tabular}{lc}
            \toprule
            \textbf{Parameter} & \textbf{Value} \\
            \midrule
            Number of UAVs & 20 \\
            Map size/km & $4 \times 4$ \\
            Total time slots & 200 \\
            Total PoIs (Type $\mathsf{I}$/$\mathsf{II}$) & 10 / 10 \\
            Base reward per task & $[2, 5]$ \\
            Minimum flying altitude / m & 50 \\
            Maximum flying altitude / m & 150 \\
            Critical altitude/m & 100 \\
            Marginal reward coefficient for data quality & 0.5 \\
            Ecological penalty weight & 0.05 \\
            \bottomrule
        \end{tabular}
\end{table}

\begin{table}[htbp]
\centering
\footnotesize
\centering
        \caption{Core parameter settings of our $\rm ES^2A$}
        \label{tab:es2a_params}
        \begin{tabular}{lc}
            \toprule
            \textbf{Parameter}  & \textbf{Value} \\
            \midrule
            Population size  & 50 \\
            Maximum iterations & 300 \\
            Proportion of discoverers & 0.2 \\
            Alarm threshold & 0.7 \\
            L\'evy flight step scaling factor & 0.01 \\
            Number of danger-aware individuals & 12 \\
            Perturbation coefficient for best individual & 0.05 \\
            Task priority encoding range & $[0, 1]$ \\
            \bottomrule
        \end{tabular}
\end{table}

\end{appendix}

\end{document}